\documentclass[runningheads]{llncs}

\usepackage{eccv}

\usepackage{eccvabbrv}

\usepackage[utf8]{inputenc}
\DeclareUnicodeCharacter{2220}{\ensuremath{\angle}}   
\DeclareUnicodeCharacter{2212}{\ensuremath{-}}         
\DeclareUnicodeCharacter{00D7}{\ensuremath{\times}}    
\DeclareUnicodeCharacter{2192}{\ensuremath{\rightarrow}} 
\DeclareUnicodeCharacter{00B0}{\ensuremath{^\circ}}    

\usepackage{graphicx}
\usepackage{booktabs}       
\usepackage{multirow}       
\usepackage{colortbl}       
\usepackage{tcolorbox}      
\usepackage{float}          
\usepackage{placeins}       
\usepackage{enumitem}       
\usepackage{tabularx,array,multicol}

\definecolor{bestcolor}{RGB}{144,238,144}    
\definecolor{secondcolor}{RGB}{173,216,230}  
\definecolor{thirdcolor}{RGB}{255,255,204}   
\newcommand{\best}[1]{\cellcolor{bestcolor}\textbf{#1}}
\newcommand{\second}[1]{\cellcolor{secondcolor}#1}
\newcommand{\third}[1]{\cellcolor{thirdcolor}#1}

\usepackage[accsupp]{axessibility}
\usepackage{hyperref}

\usepackage{orcidlink}

\begin{document}

\newcommand{\papertitle}{Beyond Final Answers: CRYSTAL Benchmark for Transparent Multimodal Reasoning Evaluation}
\title{\papertitle}

\titlerunning{CRYSTAL: Transparent Multimodal Reasoning Evaluation}

\author{Wayner Barrios \and SouYoung Jin}

\authorrunning{W. Barrios and S. Jin}

\institute{Dartmouth College, Hanover, NH, USA\\
\email{\{wayner.j.barrios.quiroga.gr, souyoung.jin\}@dartmouth.edu}}

\maketitle

\begin{abstract}
Modern multimodal large language models (MLLMs) achieve impressive results on vision-language benchmarks, yet existing evaluations judge only final answers, making shortcuts indistinguishable from genuine understanding.
We introduce \textbf{CRYSTAL}~(\textit{\textbf{C}lear \textbf{R}easoning via \textbf{Y}ielded \textbf{S}teps, \textbf{T}raceability and \textbf{L}ogic}), a diagnostic benchmark with 6,372 instances that evaluates multimodal reasoning through verifiable intermediate steps. We propose two complementary metrics: \emph{Match~F1}, which scores step-level precision and recall via semantic similarity matching, and \emph{Ordered Match~F1}, which further penalizes disordered reasoning chains. References are constructed through a Delphi-inspired pipeline where four independent MLLMs generate trajectories, aggregated via semantic clustering and validated through human quality gates.
Evaluation of 20 MLLMs, including commercial frontier systems not used during benchmark construction, reveals systematic failures invisible to accuracy: universal cherry-picking (precision far exceeds recall), non-monotonic scaling trade-offs, and disordered reasoning, where even the strongest reasoners preserve under 60\% of matched steps in correct order.
Beyond evaluation, we propose the \textbf{Causal Process Reward (CPR)}, a multiplicative reward that couples answer correctness with step-level alignment, and \textbf{CPR-Curriculum}, which progressively increases reasoning difficulty during training. CPR-Curriculum achieves +32\% Match~F1 via GRPO where additive reward strategies fail, improving reasoning without manual step annotation.
\end{abstract}


\begin{figure}[!t]
\centering
\scriptsize
\setlength{\tabcolsep}{3pt}

\begin{minipage}[t]{0.28\linewidth}
\vspace{0pt}
\centering
\includegraphics[width=\textwidth]{images/teaser_sample0.jpg}\\[2pt]
{\scriptsize\textit{Input image}}
\end{minipage}%
\hfill%
\begin{minipage}[t]{0.70\linewidth}
\vspace{0pt}
\begin{tcolorbox}[
    colback=gray!5, colframe=gray!60, boxrule=0.6pt, arc=1.5mm,
    left=2pt, right=2pt, top=2pt, bottom=2pt
]
{\scriptsize\raggedright
\textbf{Q:} Which of the 3 objects is the smallest?\\[2pt]
A.~Right\quad B.~Left\quad C.~Middle\qquad \textbf{Ground-Truth: C}
}
\end{tcolorbox}
\vspace{0.06cm}
\begin{tcolorbox}[
    colback=gray!3, colframe=gray!50, boxrule=0.6pt, arc=1.5mm,
    left=2pt, right=2pt, top=1pt, bottom=1pt,
    title={\scriptsize\bfseries Previous Benchmarks \normalfont\textit{(Answer-Only)}},
    fonttitle=\scriptsize
]
\centering\scriptsize
Model Answer: \textbf{C} \;\textcolor{green!60!black}{$\checkmark$}\quad
\textbf{Accuracy: \textcolor{green!60!black}{100\%}} \;\textcolor{gray}{\textit{(reasoning invisible)}}
\end{tcolorbox}
\vspace{0.06cm}
\begin{tcolorbox}[
    colback=blue!3, colframe=blue!40, boxrule=0.8pt, arc=1.5mm,
    left=2pt, right=2pt, top=2pt, bottom=2pt,
    title={\scriptsize\bfseries CRYSTAL \normalfont\textit{(Answer + Step-Level Evaluation)}},
    fonttitle=\scriptsize
]
\centering\scriptsize
\begin{tabular}{@{}p{0.52\linewidth}c@{\hskip 6pt}l@{}}
\toprule
\textbf{Model's Predicted Steps} & \textbf{Match} & \\
\midrule
\textit{Three gaming consoles on desk} & \textcolor{green!60!black}{$\checkmark$} & \\
\textit{Middle is \textcolor{red}{\textbf{larger}} than others} & \textcolor{red}{$\times$} &
\smash{\raisebox{-6.0ex}{\tikz{
    \node[fill=red!15, draw=red!70, rounded corners=2pt,
          inner sep=2pt, font=\scriptsize, text width=2cm, align=center] {
        \textcolor{red}{\textbf{Contradiction!}} Claims \textit{larger} but answers \textit{smallest}
    };
}}} \\
\textit{Most powerful based on size} & \textcolor{red}{$\times$} & \\
\textbf{Final Answer: C} & \textcolor{green!60!black}{$\checkmark$} & \\
\bottomrule
\end{tabular}\\[3pt]
\textbf{Match F1: \textcolor{red}{0.15}} \quad \textbf{Accuracy: \textcolor{green!60!black}{100\%}}
\end{tcolorbox}
\end{minipage}

\vspace{-0.1cm}
\caption{%
\textbf{The lucky guess problem.}
LLaVA-v1.6-7B answers correctly (C) but contradicts itself by claiming the middle console is \emph{larger} while selecting it as smallest.
Previous benchmarks score 100\%; CRYSTAL compares the model's predicted steps against reference reasoning steps via Match~F1 (0.15), exposing flawed reasoning.
}
\label{fig:teaser}
\end{figure}

\section{Introduction}
\label{sec:intro}
Modern multimodal large language models (MLLMs) have achieved impressive performance on vision-language benchmarks by integrating pretrained visual encoders with large language models~\cite{lu2024mathvista,wang2024mathvision,realworldqa}. Datasets such as MathVista~\cite{lu2024mathvista} consolidate diverse mathematical reasoning tasks, while RealWorldQA~\cite{realworldqa} challenges models with spatial understanding in real-world images. Yet a critical limitation persists: \emph{these benchmarks judge performance by final answers alone}. Without observing intermediate steps, shortcuts become indistinguishable from genuine understanding~\cite{zhang2024mmcot}. Recent theoretical analysis shows that answer-centric evaluation structurally incentivizes hallucination by penalizing models that signal uncertainty~\cite{kalai2025languagemodelshallucinate}. This motivates evaluation frameworks that assess \emph{how} models reason, not merely whether answers are correct.

Consider the example in Figure~\ref{fig:teaser} from RealWorldQA, which asks ``\textit{Which of the 3 objects is the smallest?}'' given an image of three Xbox consoles. A model answers correctly (C: middle console), but its reasoning reveals a critical contradiction: it states the middle console is \emph{larger} than the others while claiming it is the smallest. Traditional benchmarks award full credit to this ``lucky guess'' despite fundamentally flawed reasoning (Match F1: 0.15). When reasoning trajectories remain unobserved, systematic errors in perception or logic go undetected, and models exploit evaluation shortcuts rather than develop genuine understanding. Recent work addresses this by separating rationale generation from answer inference~\cite{zhang2024mmcot,cheng2024comt}. However, elicited steps typically lack structured checkpoints for machine-verifiable evaluation~\cite{xu2025mpbench}, and rarely expose \emph{where} in the reasoning chain a failure originates~\cite{zhang2024mmcot}.

We introduce \textbf{CRYSTAL}~(\textit{\textbf{C}lear \textbf{R}easoning via \textbf{Y}ielded \textbf{S}teps, \textbf{T}raceability and \textbf{L}ogic}), a diagnostic benchmark with verifiable intermediate checkpoints.\footnote{Benchmark and data: \url{https://huggingface.co/datasets/waybarrios/CRYSTAL}; code: \url{https://github.com/SEE-AI-Lab/crystal}.} Each instance contains reasoning steps scored via two novel metrics: \emph{Match~F1}, which evaluates step-level precision and recall through semantic similarity matching, and \emph{Ordered Match~F1}, which further penalizes disordered reasoning chains. By exposing the intermediate steps, CRYSTAL makes visible \emph{where} a model's reasoning breaks down, whether at perception or inference, rather than collapsing every failure into a single binary outcome.

We construct references through a Delphi-inspired pipeline~\cite{nasa2021delphi}: four independent MLLMs from different families generate trajectories, which are aggregated via semantic clustering and validated by a fifth model plus human quality gates (Section~\ref{sec:pipeline}).

Beyond evaluation, CRYSTAL's step-level references enable a new training paradigm. Standard reinforcement learning rewards combine accuracy and reasoning as independent additive terms, allowing models to maximize accuracy through guessing while ignoring reasoning quality. We propose the \textbf{Causal Process Reward (CPR)}, which multiplicatively couples both objectives: the model receives full reward only when it answers correctly \textit{and} produces faithful reasoning steps. We further introduce \textbf{CPR-Curriculum}, which stabilizes learning by progressively increasing reasoning difficulty during training.

Our contributions: \textbf{(i)} CRYSTAL, a diagnostic benchmark with 6,372 instances containing verifiable intermediate reasoning steps for fine-grained evaluation; \textbf{(ii)} Match~F1 and Ordered Match~F1, two complementary metrics that measure step-level reasoning quality via semantic similarity matching and penalize disordered reasoning chains, respectively; \textbf{(iii)} CPR and CPR-Curriculum, multiplicative reward strategies that improve both accuracy and Match~F1 via GRPO~\cite{grpo} without manual step annotation; \textbf{(iv)} evaluation of 20 MLLMs revealing universal cherry-picking behavior, persisting even in commercial frontier systems not used during benchmark construction.

\section{Related Work}
\label{sec:related}

\textbf{Answer-Centric Benchmarks.} Recent evaluation suites remain largely answer-centric despite broader scope. \emph{MathVista} and \emph{MathVision} probe mathematical reasoning with 3{,}040 competition-level problems across 16 disciplines \cite{lu2024mathvista,wang2024mathvision}, \emph{MMMU} targets expert-level multi-discipline understanding \cite{yue2024mmmu}, while \emph{SUGARCREPE} and \emph{VisMin} isolate compositional failures through minimal-pair contrastive examples \cite{hsieh2023sugarcrepe,awal2024vismin}. \emph{MMEvalPro} calibrates multimodal benchmarks with triplet-based evaluation to reduce lucky guessing~\cite{huang2025mmevalpro}, and \emph{LIME} demonstrates that carefully curated subsets can match full-benchmark signal at lower cost~\cite{zhu2025lime}. Despite their breadth, these benchmarks score only final answers, limiting visibility into intermediate reasoning.

\noindent\textbf{Process Evaluation \& Decoupling.}
Complementary work evaluates reasoning processes through intermediate steps. \emph{Multimodal-CoT} separates rationale generation from answer inference to reduce hallucination \cite{zhang2024mmcot}, \emph{Visual CoT} and \emph{Visual Sketchpad} provide step-annotated rationales and diagrammatic chains of thought \cite{shao2024visualcot,hu2024visualsketchpad}, while \emph{MME-CoT} benchmarks CoT quality and robustness \cite{jiang2025mmecot}. \emph{CoMT} requires multimodal outputs (creation, deletion, update, selection) \cite{cheng2024comt}, and \emph{MINERVA} evaluates complex video reasoning with LLM judges for multi-step temporal inference~\cite{nagrani2025minerva}. Decoupling approaches like \emph{Prism} and \emph{ViperGPT} separate perception from symbolic reasoning \cite{qiao2024prism,suris2023vipergpt}, and \emph{MPBench} emphasizes trajectory-level scoring \cite{xu2025mpbench}. These motivate machine-checkable, step-level verification of reasoning chains.

\noindent\textbf{Evaluation Methodology \& Hallucination.}
Binary answer-centric evaluation structurally incentivizes hallucination by penalizing abstention over guessing, favoring confident incorrect predictions over calibrated uncertainty~\cite{kalai2025languagemodelshallucinate}. This misaligns evaluation with trustworthy deployment. Our work addresses this by evaluating reasoning processes: Match F1 rewards semantically justified steps while penalizing spurious reasoning (low precision) and incomplete coverage (low recall), aligning incentives with transparent, verifiable inference.

\noindent\textbf{Robustness \& Reliability.}
Trustworthy reasoning requires robustness and calibration. \emph{MultiTrust} covers truthfulness, safety, and fairness across 32 tasks \cite{zhang2024multitrust}. MLLMs remain vulnerable to adversarial attacks despite instruction tuning \cite{cui2024robustlmm}, while calibration methods like \emph{Self-Calibrated Tuning} and \emph{CaRot} improve OOD detection \cite{yu2024sct,oh2024carot}. Real-world benchmarks reveal persistent gaps in long-horizon multimodality \cite{zhang2025mmerealworld}, and compositional probes expose sensitivity to minimal changes \cite{hsieh2023sugarcrepe,awal2024vismin}. Our design emphasizes deterministic step graders and trajectory coherence at the process level.
\section{CRYSTAL Dataset \& Benchmark}
\label{sec:dataset}
We introduce \textbf{CRYSTAL} (\textit{\textbf{C}lear \textbf{R}easoning via \textbf{Y}ielded \textbf{S}teps, \textbf{T}raceability and \textbf{L}ogic}), a diagnostic benchmark that evaluates multimodal reasoning \emph{step by step}. Unlike traditional VQA benchmarks that only assess final answer correctness, CRYSTAL provides each question with a sequence of natural language reference reasoning steps that capture the intermediate inferences required to arrive at the correct answer. These references are generated through a multi-agent framework (Section~\ref{sec:pipeline}) that aggregates outputs from independent MLLMs via semantic clustering, ensuring diverse yet high-quality reasoning paths. CRYSTAL spans 6,372 questions covering visual perception, compositional reasoning, spatial relations, counting, and logical inference. Evaluation compares predicted steps against references using semantic similarity matching (Match F1) alongside answer accuracy, enabling fine-grained analysis of \emph{where} and \emph{why} models fail. Table~\ref{tab:key_statistics} summarizes key statistics.


\begin{table}[t]
\caption{\textbf{Key statistics of CRYSTAL.}}
\label{tab:key_statistics}
\centering
\footnotesize
\begin{minipage}[t]{0.38\linewidth}
\centering
\begin{tabular}{l r}
\toprule
\textbf{Statistic} & \textbf{Value} \\
\midrule
Total examples & 6,372 \\
Avg.\ steps & 11.6 \\
Step range & 3--42 \\
Easy / Med / Hard & 48.5 / 44.1 / 7.4\% \\
Annotation & Multi-agent + human \\
\bottomrule
\end{tabular}
\end{minipage}%
\hfill%
\begin{minipage}[t]{0.55\linewidth}
\centering
\begin{tabular}{l r}
\toprule
\textbf{Source} & \textbf{Count} \\
\midrule
MathVision~\cite{wang2024mathvision} & 3,039 (47.7\%) \\
ScienceQA-IMG~\cite{lu2022scienceqa} & 2,017 (31.7\%) \\
RealWorldQA~\cite{realworldqa} & 765 (12.0\%) \\
MMVP~\cite{tong2024mmvp} & 299 (4.7\%) \\
PlotQA~\cite{plotqa} & 252 (3.9\%) \\
\bottomrule
\end{tabular}
\end{minipage}
\end{table}


\begin{figure}[ht]
\centering
\setlength{\tabcolsep}{4pt}
\renewcommand{\arraystretch}{1.0}

\begin{tabular}{@{}ccc@{}}
\begin{minipage}[t]{0.32\textwidth}
  \centering
  {\scriptsize\bfseries RealWorldQA\par}
  \vspace{2pt}
  \fcolorbox{black!20}{white}{%
    \includegraphics[width=0.95\linewidth,height=2.5cm,keepaspectratio]{images/dataset_example_sample26.jpg}%
  }\\[2pt]
  {\scriptsize
    \colorbox{blue!6}{%
      \parbox{0.95\linewidth}{%
        \textbf{Q:} In which direction is the dog traveling? \\[2pt]
        \textbf{GT:} Left to Right%
      }%
    }%
  }\\[1pt]
  {\scriptsize
    \colorbox{yellow!8}{%
      \parbox{0.95\linewidth}{%
        \textbf{Reference Steps (6):}
        \begin{enumerate}[leftmargin=11pt,itemsep=1pt,parsep=0pt,topsep=2pt]
          \setlength{\parskip}{0pt}
          \item Person walking dog across street.
          \item Dog's head turned toward right.
          \item Body orientation indicates rightward movement.
          \item Motion blur suggests active movement.
          \item Leash extends from left to right.
          \item Dog traveling left to right.
        \end{enumerate}
      }%
    }%
  }
\end{minipage}
&
\begin{minipage}[t]{0.32\textwidth}
  \centering
  {\scriptsize\bfseries MMVP\par}
  \vspace{2pt}
  \fcolorbox{black!20}{white}{%
    \includegraphics[width=0.95\linewidth,height=2.5cm,keepaspectratio]{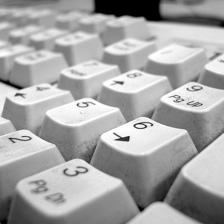}%
  }\\[2pt]
  {\scriptsize
    \colorbox{blue!6}{%
      \parbox{0.95\linewidth}{%
        \textbf{Q:} Can you see the key ``Z'' in the image? \quad \\[2pt]
        \textbf{GT:} No%
      }%
    }%
  }\\[1pt]
  {\scriptsize
    \colorbox{yellow!8}{%
      \parbox{0.95\linewidth}{%
        \textbf{Reference Steps (10):}
        \begin{enumerate}[leftmargin=11pt,itemsep=1pt,parsep=0pt,topsep=2pt]
          \setlength{\parskip}{0pt}
          \item The image shows a close-up of keyboard keys.
          \item Foreground keys include a key labeled `9' with `Pg Up' text.
          \item Another visible key is labeled `6' with a right arrow.
          \item Other keys in view carry navigation labels rather than letters.
        \end{enumerate}
        [\textit{6 more steps}]%
      }%
    }%
  }
\end{minipage}
&
\begin{minipage}[t]{0.32\textwidth}
  \centering
  {\scriptsize\bfseries ScienceQA\par}
  \vspace{2pt}
  \fcolorbox{black!20}{white}{%
    \includegraphics[width=0.95\linewidth,height=2.5cm,keepaspectratio]{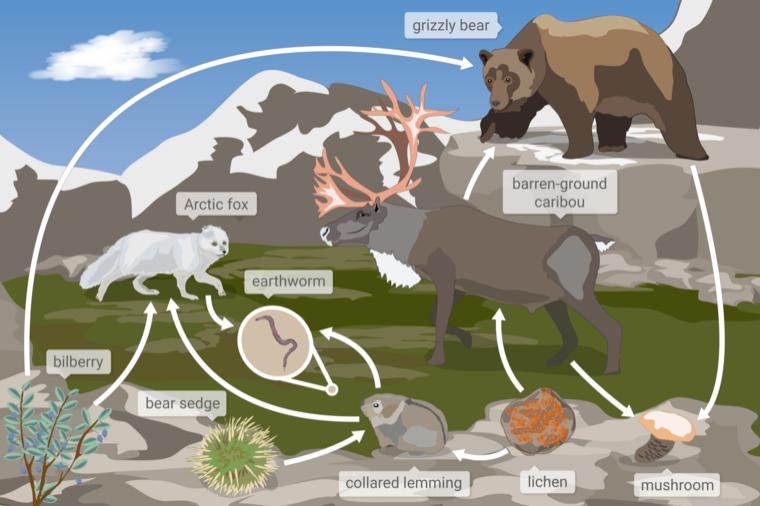}%
  }\\[2pt]
  {\scriptsize
    \colorbox{blue!6}{%
      \parbox{0.95\linewidth}{%
        \textbf{Q:} Which organism contains matter from lichen? \\[2pt]
        \textbf{GT:} Mushroom%
      }%
    }%
  }\\[1pt]
  {\scriptsize
    \colorbox{yellow!8}{%
      \parbox{0.95\linewidth}{%
        \textbf{Reference Steps (15):}
        \begin{enumerate}[leftmargin=11pt,itemsep=1pt,parsep=0pt,topsep=2pt]
          \setlength{\parskip}{0pt}
          \item Food web shows energy flow.
          \item Arrows = matter transfer direction.
          \item Lichen at bottom level.
          \item Arrow from lichen to mushroom.
          \item Mushroom consumes lichen matter.
          \item Mushroom contains lichen matter.
        \end{enumerate}
        [\textit{9 more steps}]%
      }%
    }%
  }
\end{minipage}
\end{tabular}

\caption{\textbf{CRYSTAL spans diverse multimodal reasoning scenarios.} Three examples: \textbf{(Left)} RealWorldQA (spatial understanding), \textbf{(Middle)} MMVP (fine-grained perception), \textbf{(Right)} ScienceQA (multi-hop logic). Parentheses give the reference step count per example.}
\label{fig:dataset_examples_horizontal}
\end{figure}


Figure~\ref{fig:dataset_examples_horizontal} illustrates representative examples from CRYSTAL spanning diverse reasoning scenarios. Each instance combines visual inputs with natural language questions and verifiable step-by-step reasoning paths, enabling fine-grained evaluation of where models succeed (correct perception, valid inference) versus where they fail (hallucinated objects, logical errors).

\subsection{Multi-Agent Reasoning Step Generation}
\label{sec:pipeline}

Each input consists of a question $Q$, an image $I$, and the correct answer $A$. Our goal is to produce an ordered sequence of minimal reasoning steps necessary to derive $A$ from $Q$ and $I$. The pipeline comprises four phases with two iterative quality gates inspired by the Delphi method~\cite{nasa2021delphi}.

\noindent\textbf{Phase 1: Independent generation.}
Four open-source MLLMs from different architecture families (Qwen2.5-VL-72B~\cite{qwen2vl2024}, InternVL3-76B~\cite{internvl3}, Gemma3-27B~\cite{gemma2024}, and Llama-4-Maverick~\cite{llama4}) independently generate candidate reasoning steps from the triplet $(Q, I, A)$, using distinct random seeds to reduce correlated errors~\cite{krogh1995neural,breiman1996bagging}.

\noindent\textbf{Phase 2: Semantic clustering and ordering.}
We embed all candidate steps using a sentence encoder $f(\cdot)$ and compute pairwise cosine similarity. Steps with similarity $\ge\tau$ form edges in an undirected graph; connected components define semantic clusters $\{\mathcal{C}_k\}$, each grouping paraphrastic formulations of the same reasoning step. For each cluster, we select the representative minimizing average within-cluster dissimilarity:
\begin{equation}
x_k^{\star} \;=\; \arg\min_{x\in \mathcal{C}_k}\; \frac{1}{|\mathcal{C}_k|}\sum_{y\in \mathcal{C}_k}\!\bigl(1-\mathrm{sim}(x,y)\bigr).
\label{eq:representative}
\end{equation}
Representatives are ordered into a coherent reasoning chain. On the first Delphi round, ordering follows the original question logic. On each subsequent round $t$, the ordering is updated by minimizing edit distance to the previous round's sequence, stabilizing step order across refinements. This clustering effectively implements self-consistency voting~\cite{wang2023selfconsistency} at the step level.

\noindent\textbf{Phase 3: Automated validation.}
A fifth MLLM (Molmo-72B~\cite{molmo}) validates logical soundness, sequence coherence, visual grounding in $I$, and consistency with answer $A$. Failed examples re-enter Phase~1 with fresh seeds (\emph{Iteration Loop~1}): new candidate steps are generated, re-clustered, and re-validated, mirroring Delphi consensus building where contributors revise proposals after aggregated feedback~\cite{nasa2021delphi}.

\noindent\textbf{Phase 4: Human quality gate.}
A trained annotator verifies that perceptual claims are visible in $I$, reasoning transitions are logically sound, and executing the steps yields $A$. Rejected examples restart from Phase~1 (\emph{Iteration Loop~2}); fewer than 5\% require re-iteration. Together, both loops form a closed refinement cycle that mirrors multi-annotator label aggregation~\cite{dawid1979mle,raykar2010jmlr}. The supplementary material visualizes the overall pipeline workflow and provides prompt templates, annotator guidelines, and the annotation interface.


\subsection{Evaluation Metrics}
\label{sec:metrics}

We evaluate models via two metrics: \textit{Match F1} measures step-level reasoning quality through semantic similarity matching, while \textit{Accuracy} measures final answer correctness.

\noindent\textbf{Match F1.}
Given predicted steps $\mathcal{P}_i=\{p_{ij}\}$ and reference steps $\mathcal{G}_i=\{g_{ik}\}$ for example $i$, we compute pairwise cosine similarity using sentence encoder $f(\cdot)$:
\begin{equation}
S_{jk} = \cos\!\bigl(f(p_{ij}),f(g_{ik})\bigr).
\label{eq:sim}
\end{equation}
Steps are matched via greedy 1:1 assignment over pairs with $S_{jk}\ge\tau$ (threshold), producing a set of matched pairs $\mathcal{A}_i$. We define true positives $\mathrm{TP}_i=|\mathcal{A}_i|$, false positives $\mathrm{FP}_i=|\mathcal{P}_i|-\mathrm{TP}_i$, and false negatives $\mathrm{FN}_i=|\mathcal{G}_i|-\mathrm{TP}_i$. Precision and recall are:
\begin{equation}
\mathrm{Prec}_i = \frac{\mathrm{TP}_i}{\max\{|\mathcal{P}_i|,1\}},\quad
\mathrm{Rec}_i = \frac{\mathrm{TP}_i}{\max\{|\mathcal{G}_i|,1\}}.
\label{eq:prec_rec}
\end{equation}
Per-example F1 combines precision and recall via harmonic mean (0 if both denominators are non-empty yet $\mathrm{TP}_i=0$; 1 if $|\mathcal{P}_i|=|\mathcal{G}_i|=0$). Dataset-level \textbf{Match F1} is the macro-average over all $N$ evaluation examples:
\begin{equation}
\mathrm{Match\text{-}F1} = \frac{1}{N}\sum_{i=1}^{N}\mathrm{F1}_i.
\label{eq:f1_macro}
\end{equation}
We use \texttt{all-distilroberta-v1}~\cite{reimers2019sbert} with $\tau=0.35$ (selected via ablation in Section~\ref{sec:ablation}).

\noindent\textbf{Ordered Match F1.}
We extend Match F1 with the Longest Increasing Subsequence (LIS) ratio~\cite{fredman1975lis} to penalize disordered chains. Given $m\!=\!|\mathcal{A}_i|$ matched pairs, we sort them by reference index and extract the corresponding prediction indices $(j_1,\ldots,j_m)$. The LIS ratio measures the fraction of these indices that form a non-decreasing subsequence:
\begin{equation}
\text{LIS\text{-}ratio}_i = \frac{\text{LIS}(j_1,\ldots,j_m)}{m}.
\label{eq:lis}
\end{equation}
Ordered Match F1 combines content quality with ordering:
\begin{equation}
\text{Ordered-F1}_i = \text{F1}_i \cdot \bigl((1 - \alpha) + \alpha \cdot \text{LIS\text{-}ratio}_i\bigr),
\label{eq:ordered_f1}
\end{equation}
with $\alpha\!=\!0.3$. When $\alpha\!=\!0$, this reduces to standard Match F1.

\noindent\textbf{Accuracy.}
Answer correctness uses type-adapted comparison: tolerance-based matching for numerics, exact matching for categoricals, and LLM-as-judge for free-form text, macro-averaged over all examples.

\subsection{Causal Process Reward}
\label{sec:cpr}

Standard reward formulations for reinforcement learning from reasoning combine accuracy and reasoning quality additively (e.g., $R = w_f R_{\text{fmt}} + w_a R_{\text{acc}} + w_r R_{\text{reason}}$, where $w_f$, $w_a$, $w_r$ weight format compliance, answer accuracy, and reasoning quality respectively; see the supplementary material), allowing the accuracy term to dominate and the model to maximize reward by guessing correctly while omitting reasoning steps. We propose the \textbf{Causal Process Reward (CPR)}, which uses a multiplicative interaction that \textit{causally} links answer correctness to step-level alignment:
\begin{equation}
R_{\text{CPR}} = \begin{cases}
a_w + s_w \cdot \text{F1}_{\text{step}} & \text{if answer correct} \\
s_w \cdot \text{F1}_{\text{step}} \cdot \lambda & \text{otherwise}
\end{cases}
\label{eq:cpr}
\end{equation}
where $a_w$ is the answer bonus, $s_w$ is the step bonus (weighting reasoning quality), $\text{F1}_{\text{step}}$ is the step-level matching F1 between predicted and reference steps (lexical similarity proxy during training; see supplementary), and $\lambda\!=\!0.3$ penalizes wrong answers. Under this formulation, a correct guess without reasoning receives only $a_w$ (no step bonus), while good reasoning with a wrong answer is heavily discounted, ensuring neither objective can be independently maximized. We extend CPR with \textbf{CPR-Curriculum}, a two-phase training strategy inspired by staged reward shaping~\cite{deepseekr1}. Phase~1 trains with only format and accuracy rewards (no reasoning signal) to establish stable answer generation. Phase~2 initializes from the Phase~1 checkpoint and introduces the full CPR reward at a lower learning rate, preserving the accuracy foundation while adding reasoning supervision. Within Phase~2, we apply progressive difficulty scheduling: training begins with examples having fewer reference steps (simpler reasoning chains) and gradually introduces examples with more steps (complex multi-hop reasoning), preventing early-stage collapse before the model learns to generate structured reasoning. Since accuracy and reasoning can produce conflicting gradients, both strategies use PCGrad~\cite{yu2020pcgrad} to project conflicting gradient components, preventing one objective from undermining the other during optimization.
\section{Experiments}
\label{sec:experiments}

We evaluate 20 MLLMs (16 open-source, 1B--38B; 4 commercial) on CRYSTAL. Our experiments address: \textbf{(i)}~step-level performance beyond accuracy, \textbf{(ii)}~cherry-picking in commercial frontier models, \textbf{(iii)}~reasoning step ordering, and \textbf{(iv)}~step-level rewards for training.

\subsection{Experimental Setup}
\label{sec:exp_setup}

\textbf{Dataset.} CRYSTAL contains 6,372 questions from five benchmarks (MathVision~\cite{wang2024mathvision} 47.7\%, ScienceQA-IMG~\cite{lu2022scienceqa} 31.7\%, RealWorldQA~\cite{realworldqa} 12.0\%, MMVP~\cite{tong2024mmvp} 4.7\%, PlotQA~\cite{plotqa} 3.9\%), averaging 11.6 reasoning steps per question (Table~\ref{tab:key_statistics}). Questions are stratified by complexity: easy (48.5\%, 8.6 steps), medium (44.1\%, 13.3 steps), hard (7.4\%, 20.5 steps), using model-agnostic features (step count, question length, linguistic markers).

\noindent\textbf{Baselines.} We evaluate 20 MLLMs spanning open-source and commercial systems.
\textit{Open-source} (16 models): Qwen2.5-VL (3B, 7B, 32B)~\cite{qwen2vl2024}, Qwen3-VL (2B, 8B, 32B)~\cite{qwen3vl2024}, InternVL3.5 (1B to 38B)~\cite{internvl35}, Gemma3 (4B, 12B)~\cite{gemma2024}, Llama~3.2-11B~\cite{llama32}, LLaVA-v1.6 (7B)~\cite{liu2024llavanext}, and MiniCPM-v2.6 (8B)~\cite{minicpmv2024}.
\textit{Commercial} (4 models): GPT-5, GPT-5-mini, GPT-5.2 Instant~\cite{gpt5}, and Gemini 2.5 Flash~\cite{gemini25flash}. All models use official endpoints or pretrained checkpoints without task-specific fine-tuning. Crucially, commercial models were \textit{not} used in the CRYSTAL reference generation pipeline, providing an unbiased assessment of whether observed patterns generalize to frontier systems.

\noindent\textbf{Metrics.} \textit{Accuracy} measures final answer correctness via fuzzy matching; \textit{Match F1} evaluates step-level reasoning through semantic similarity matching (all-distilroberta-v1, $\tau\!=\!0.35$). Match F1 combines precision (fraction of predicted steps aligned with references) and recall (fraction of reference steps covered).

\noindent\textbf{GRPO Training Data.} For the GRPO training experiment~\cite{grpo}, we use 30,312 examples from ScienceQA-IMG~\cite{lu2022scienceqa} (6,218 samples) and TextVQA~\cite{textvqa} (24,094 samples), generated via the pipeline from Section~\ref{sec:pipeline}. The supplementary material provides the complete pipeline workflow diagram, prompt templates, benchmarking settings, GRPO training configuration, and qualitative examples.

\subsection{Main Results}
\label{sec:main_results}

Table~\ref{tab:main_results} presents evaluation results across 20 MLLMs, revealing systematic gaps between answer correctness and reasoning transparency.


\begin{table}[!ht]
\caption{%
\textbf{Evaluation of 20 MLLMs on CRYSTAL.}
Match F1 (all-distilroberta-v1, $\tau\!=\!0.35$): step-level reasoning quality; P/R: step precision/recall; LIS: fraction of matched steps in correct order; Ord.\ F1 ($\alpha\!=\!0.3$): order-penalized Match F1. Top three per column in \best{green}, \second{blue}, \third{yellow}.
}
\label{tab:main_results}
\centering
\setlength{\tabcolsep}{3.5pt}
\resizebox{\linewidth}{!}{%
\begin{tabular}{l c c c c c c c c}
\toprule
\textbf{Model} & \textbf{Params} & \textbf{Accuracy} & \textbf{Match F1} & \textbf{P} & \textbf{R} & \textbf{Steps} & \textbf{LIS} & \textbf{Ord.\ F1} \\
\midrule
\multicolumn{9}{l}{\textit{Commercial MLLMs}} \\
GPT-5~\cite{gpt5}               & n/a  & \best{57.99\%} & 0.612 & 0.925 & 0.479 & 5.29 & 0.636 & 0.539 \\
GPT-5-mini~\cite{gpt5}           & n/a  & \third{55.59\%} & \best{0.773} & \best{0.978} & \third{0.669} & \third{7.57} & 0.560 & \best{0.670} \\
GPT-5.2 Instant~\cite{gpt5}      & n/a  & 47.35\% & 0.564 & \second{0.974} & 0.416 & 4.64 & 0.648 & 0.501 \\
Gemini 2.5 Flash~\cite{gemini25flash}     & n/a  & 53.95\% & \third{0.673} & 0.701 & \best{0.765} & \best{17.10} & 0.584 & \third{0.579} \\
\midrule
\multicolumn{9}{l}{\textit{Qwen Family}} \\
Qwen3-VL-8B~\cite{qwen3vl2024}         & 8B  & \second{57.66\%} & 0.659 & 0.827 & 0.590 & 7.37 & 0.624 & 0.572 \\
Qwen3-VL-32B~\cite{qwen3vl2024}        & 32B & 49.22\% & \second{0.718} & 0.819 & \second{0.704} & \second{10.56} & 0.581 & \second{0.617} \\
Qwen2.5-VL-32B~\cite{qwen2vl2024}      & 32B & 47.63\% & 0.653 & 0.943 & 0.524 & 5.86 & 0.619 & 0.572 \\
Qwen3-VL-2B~\cite{qwen3vl2024}         & 2B  & 34.15\% & 0.595 & 0.726 & 0.535 & 5.94 & 0.669 & 0.515 \\
Qwen2.5-VL-7B~\cite{qwen2vl2024}       & 7B  & 30.43\% & 0.475 & 0.765 & 0.365 & 4.07 & 0.717 & 0.422 \\
Qwen2.5-VL-3B~\cite{qwen2vl2024}       & 3B  & 39.85\% & 0.480 & 0.898 & 0.347 & 3.73 & 0.723 & 0.434 \\
\midrule
\multicolumn{9}{l}{\textit{InternVL Family}} \\
InternVL3.5-38B~\cite{internvl35}     & 38B & 51.21\% & 0.612 & 0.892 & 0.498 & 5.76 & 0.643 & 0.538 \\
InternVL3.5-8B~\cite{internvl35}      & 8B  & 51.98\% & 0.530 & 0.882 & 0.416 & 4.96 & 0.692 & 0.469 \\
InternVL3.5-4B~\cite{internvl35}      & 4B  & 37.61\% & 0.432 & 0.895 & 0.325 & 3.75 & 0.775 & 0.387 \\
InternVL3.5-2B~\cite{internvl35}      & 2B  & 33.02\% & 0.469 & 0.725 & 0.371 & 3.92 & 0.731 & 0.415 \\
InternVL3.5-1B~\cite{internvl35}      & 1B  & 30.13\% & 0.330 & 0.616 & 0.243 & 2.51 & 0.807 & 0.297 \\
\midrule
\multicolumn{9}{l}{\textit{Other Open-Source}} \\
Gemma3-12B~\cite{gemma2024}          & 12B & 33.83\% & 0.605 & 0.838 & 0.499 & 5.51 & 0.673 & 0.534 \\
Gemma3-4B~\cite{gemma2024}           & 4B  & 28.65\% & 0.618 & 0.878 & 0.506 & 5.72 & 0.668 & 0.547 \\
Llama~3.2-11B~\cite{llama32}       & 11B & 24.83\% & 0.471 & 0.713 & 0.379 & 4.19 & 0.726 & 0.415 \\
LLaVA-v1.6-7B~\cite{liu2024llavanext}       & 7B  & 24.66\% & 0.512 & \third{0.961} & 0.370 & 3.94 & 0.675 & 0.459 \\
MiniCPM-v2.6-8B~\cite{minicpmv2024}      & 8B  & 25.54\% & 0.215 & 0.709 & 0.134 & 1.31 & 0.854 & 0.186 \\
\bottomrule
\end{tabular}%
}
\end{table}


\noindent\textbf{Finding 1: Cherry-picking is near-universal.} 19 of 20 models exhibit precision substantially exceeding recall (ratios 1.2$\times$ to 5.3$\times$), including commercial systems absent from the reference pipeline: GPT-5 (P/R\,=\,0.925/0.479), GPT-5-mini (0.978/0.669), and GPT-5.2 Instant (0.974/0.416). The sole exception is Gemini~2.5~Flash, which generates 17.10 steps per question (vs.\ 11.6 reference average) with recall (0.765) exceeding precision (0.701). The persistence of cherry-picking across scales (1B to frontier) and 7 model families suggests a fundamental misalignment between training objectives and reasoning transparency~\cite{kalai2025languagemodelshallucinate}. Even GPT-5 (57.99\% accuracy) recovers only 47.9\% of reference steps. Importantly, commercial models were not part of the reference generation pipeline, yet GPT-5-mini achieves the highest F1 (0.773) and Gemini the highest recall across all 20 models, confirming that Match~F1 captures genuine reasoning quality without benchmark construction bias. Low recall reflects genuine omissions rather than over-complete references: Gemini~2.5~Flash attains 0.765 recall (coverage is achievable), while models generating few steps (LLaVA-v1.6: 3.94, GPT-5.2~Instant: 4.64) still reach $>$0.96 precision, aligning with references yet omitting intermediate reasoning.

\noindent\textbf{Finding 2: Accuracy and reasoning fidelity diverge.} GPT-5 leads accuracy (57.99\%) but ranks 8th in F1 (0.612), while GPT-5-mini leads F1 (0.773) at lower accuracy (55.59\%). Cross-family comparisons reveal architectural choices dominate scale: Gemma3-4B (0.618~F1) outperforms InternVL3.5-38B (0.612~F1) despite 9.5$\times$ fewer parameters~\cite{sun2025empirical,li2025small}. Models like LLaVA-v1.6 (0.961 precision, 3.94 steps) and GPT-5.2 Instant (0.974 precision, 4.64 steps) achieve near-perfect step accuracy by generating fewer, safer steps, masking reasoning failures through confident omissions~\cite{chen2024measuring,xu2024llavacot}. This divergence is enabled by the structure of the evaluation: 36.4\% of CRYSTAL questions are multiple-choice or binary (ScienceQA-IMG, MMVP), where models can exploit statistical regularities in answer distributions and surface-level visual cues to select correct answers without complete reasoning~\cite{goyal2017vqa,geirhos2020shortcut}. This motivates step-level rewards that explicitly incentivize reasoning coverage (Section~\ref{sec:grpo_results}).

\noindent\textbf{Finding 3: Scaling exhibits non-monotonic reasoning trade-offs.} Within model families, scaling parameters does not uniformly improve both accuracy and reasoning quality. Qwen3-VL-32B achieves 0.718~F1 (2nd overall) but lower accuracy (49.22\%) than Qwen3-VL-8B (57.66\%, 0.659~F1), indicating that the larger model produces more thorough reasoning chains at the cost of answer correctness. Conversely, InternVL3.5-4B yields lower F1 (0.432) than InternVL3.5-2B (0.469) despite higher accuracy (37.61\% vs.\ 33.02\%), suggesting that scale can improve answer extraction while suppressing reasoning coverage. These non-monotonic patterns demonstrate that accuracy and reasoning quality respond differently to parameter scaling~\cite{sun2025empirical}, reinforcing the necessity of joint evaluation through metrics like Match~F1.

\FloatBarrier
\subsection{Reasoning Order Analysis}
\label{sec:order_analysis}

Match F1 measures \textit{what} steps are present but not \textit{whether they appear in a logically coherent sequence}. Using Ordered Match F1 (Eq.~\ref{eq:ordered_f1}), which penalizes out-of-sequence steps via the LIS ratio (Eq.~\ref{eq:lis}), we evaluate whether models organize reasoning chains in the correct order. The last two columns of Table~\ref{tab:main_results} report LIS and Ordered F1 across all 20 MLLMs.

\noindent\textbf{Finding 4: No model preserves coherent reasoning order.}
Among the top reasoners by Match F1 ($>$ 0.66), no model preserves more than 60\% of matched steps in the correct relative order. GPT-5-mini achieves the highest Ordered F1 (0.670) but with LIS\,=\,0.560, meaning 44\% of its matched steps appear out of sequence. Qwen3-VL-32B (LIS\,=\,0.581) and Gemini~2.5~Flash (LIS\,=\,0.584) show similar ordering degradation despite strong Match F1. Notably, LIS alone can be misleading: smaller models such as InternVL3.5-1B (LIS\,=\,0.807) and MiniCPM-v2.6 (LIS\,=\,0.854) achieve high ratios simply because generating 2 to 3 matched steps is trivially ordered regardless of reasoning quality. Ordered Match F1 accounts for this by weighting LIS with Match F1, ensuring that high ordering scores require substantive step coverage.

\noindent\textbf{Finding 5: Ordering is orthogonal to cherry-picking.}
Comparing Findings 1 to 3 with ordering results reveals that cherry-picking (high precision, low recall) and disordered reasoning are independent failure modes. GPT-5.2~Instant (precision 0.974) cherry-picks aggressively yet maintains moderate order (LIS\,=\,0.648), while Gemini~2.5~Flash, the only model with recall exceeding precision, achieves comparable ordering (0.584). This suggests that current MLLMs retrieve relevant reasoning steps but fail to organize them into coherent chains regardless of their coverage strategy, pointing to a fundamental gap in sequential reasoning that training objectives do not address.

\FloatBarrier
\subsection{Ablation Studies}
\label{sec:ablation}

\begin{figure}[ht]
    \centering
    \includegraphics[width=\textwidth]{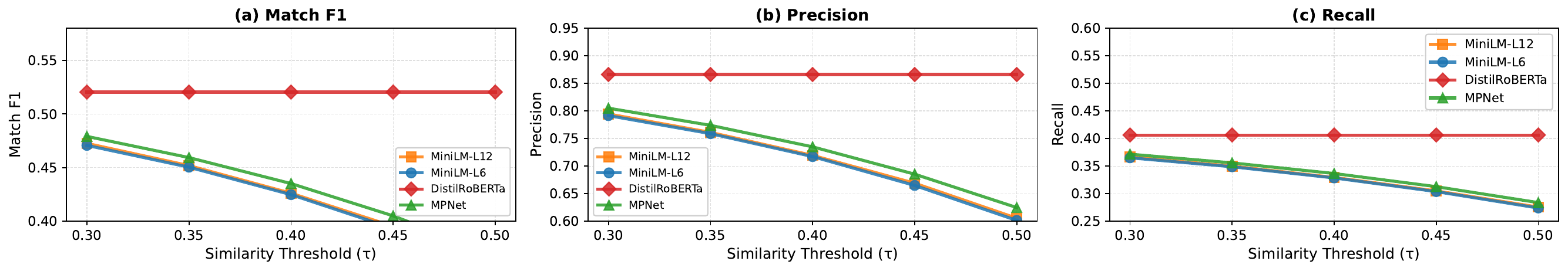}
    \caption{\textbf{Ablation: encoder and threshold.} 4 encoders across 5 thresholds, averaged over all models. (a) \texttt{all-distilroberta-v1} achieves the highest Match F1 (+4.9pp at $\tau=0.35$). (b--c) higher thresholds reduce precision and recall for most encoders, while DistilRoBERTa stays stable. We select $\tau=0.35$.}
    \label{fig:ablation_encoders}
\end{figure}

We conduct 100 experiments testing 4 sentence encoders across 5 thresholds ($\tau \in \{0.30, 0.35, 0.40, 0.45, 0.50\}$) on 5 baselines. Figure~\ref{fig:ablation_encoders} shows encoder choice dominates threshold selection: DistilRoBERTa-v1~\cite{reimers2019sbert} achieves 0.520 F1 versus 0.471--0.479 for competitors, with 4--8pp advantages transferring across all model families. Threshold variation yields only 2--3pp swings, confirming that evaluation reliability depends on encoder quality independent of the model being evaluated. Critically, model rankings remain stable across all 20 encoder-threshold configurations, indicating that the matching captures consistent semantic relationships rather than artifacts of a particular similarity function. This cross-encoder stability serves as an empirical proxy for human alignment: if the matching were unreliable, different encoders would produce contradictory rankings. We adopt all-distilroberta-v1 with $\tau\!=\!0.35$, and validate the matching directly against human judgment below.

\noindent\textbf{Human agreement.}
A benchmark metric is only as trustworthy as its agreement with human semantic judgment. We sample 100 (predicted step, reference step) pairs from three models spanning the performance spectrum: GPT-5~\cite{gpt5} (57.99\% acc.), Qwen2.5-VL-7B~\cite{qwen2vl2024} (30.43\%), and Llama~3.2-11B~\cite{llama32} (24.83\%). To stress-test the decision boundary, we stratify pairs by cosine similarity: 33 low ($<$0.20), 33 borderline (0.20--0.50), and 34 high ($>$0.50). A human annotator labels each pair as semantically equivalent or not, \textit{blinded} to the encoder's score and decision; this adversarial sampling over-represents hard cases, yielding a conservative estimate of agreement.

\begin{table}[!ht]
\centering
\caption{\textbf{Human agreement with embedding-based step matching.} Agreement between the encoder's decisions and a blinded human annotator on 100 step pairs stratified by cosine similarity. Agreement is perfect below the operating threshold ($\tau\!=\!0.35$); disagreements stay in the ambiguous borderline zone.}
\label{tab:human_agreement}
\small
\begin{tabular}{lccc}
\toprule
\textbf{Similarity Band} & \textbf{Pairs} & \textbf{Agreement} & \textbf{Note} \\
\midrule
$< 0.20$ & 33 & 100.0\% & No false matches \\
$[0.20, 0.35)$ & 14 & 100.0\% & No false matches \\
\rowcolor{gray!10}
$[0.35, 0.50)$ & 19 & 68.4\% & Borderline zone \\
\rowcolor{gray!10}
$[0.50, 0.70)$ & 25 & 64.0\% & Borderline zone \\
$\geq 0.70$ & 9 & 88.9\% & Clear matches \\
\midrule
\textbf{Overall} & \textbf{100} & \textbf{84.0\%} & $\kappa = 0.534$ \\
\bottomrule
\end{tabular}
\end{table}

Below the operating threshold ($\tau\!=\!0.35$), encoder and annotator agree on \textit{every} pair (47/47): the encoder \textbf{never produces a false match on semantically unrelated steps}, the most critical property for a benchmark metric, since false matches would inflate scores and mask reasoning deficiencies. Overall agreement is 84\% ($\kappa\!=\!0.534$), consistent with inter-annotator rates for comparable tasks~\cite{reimers2019sbert}; all 16 disagreements are false positives, cases where the encoder proposes a match the annotator declines, and they concentrate in the borderline band (15 at $0.35\!\leq\!\text{sim}\!<\!0.70$, one at $\text{sim}\!\geq\!0.70$) where semantic equivalence is genuinely ambiguous. The encoder produces no false negatives, and because this stratified sampling deliberately over-represents borderline pairs, effective agreement in normal evaluation exceeds 84\%.

\begin{figure}[!ht]
\centering
\includegraphics[width=0.5\linewidth]{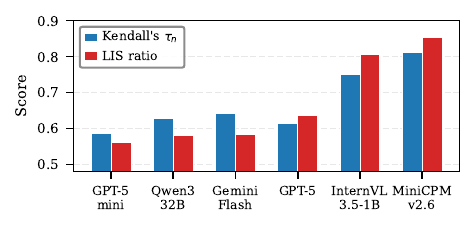}
\caption{\textbf{Order metric comparison on a representative subset of 6 models.} Both metrics rise for weak models generating few steps, but LIS ratio provides wider inter-group discrimination (0.56--0.85 vs.\ 0.58--0.81), clearly exposing trivially ordered few-step outputs.}
\label{fig:order_ablation}
\end{figure}

\noindent\textbf{Order metric selection.}
We compare two candidates for measuring step ordering in Ordered Match F1: Kendall's Tau (normalized to $[0,1]$)~\cite{kendall1938tau} and LIS ratio (Eq.~\ref{eq:lis}). Figure~\ref{fig:order_ablation} reports both metrics on a representative subset of six models deliberately spanning strong (GPT-5-mini, Qwen3-VL-32B), moderate (Gemini~2.5~Flash, GPT-5), and weak (InternVL3.5-1B, MiniCPM-v2.6-8B) reasoning profiles to stress-test discrimination across the full performance spectrum.


Kendall's $\tau_n$ shows limited discrimination: GPT-5-mini (7.57~steps, $\tau_n\!=\!0.586$) and InternVL3.5-1B (2.51~steps, $\tau_n\!=\!0.751$) differ by only 0.165 points. LIS ratio exposes this contrast more clearly (0.560 vs.\ 0.807, gap of 0.247), reflecting that models generating few matched steps achieve trivially high ordering. We adopt LIS for its wider dynamic range and direct interpretability as the fraction of steps readable in correct order.

\FloatBarrier
\subsection{Reinforcement Learning with Step-Level Rewards}
\label{sec:grpo_results}

A key question is whether CRYSTAL can guide model improvement through training, not just evaluation. We apply our Causal Process Reward (CPR) and CPR-Curriculum (Section~\ref{sec:cpr}) as GRPO~\cite{grpo} reward strategies to two architecturally distinct models under \textit{identical} CPR reward weights, Phase-2 learning rate, and data: Qwen2.5-VL-3B-Instruct~\cite{qwen2vl2024} and InternVL3.5-4B~\cite{internvl35}, the latter pairing an InternViT encoder with a Qwen3-based backbone (4$\times$A100, DeepSpeed ZeRO-3; full hyperparameters in the supplementary material). Both baselines already appear in Table~\ref{tab:main_results}, so the comparison rests on pre-existing references. Table~\ref{tab:grpo_results} shows CPR-Curriculum simultaneously improves accuracy \emph{and} Match~F1 on both: Qwen gains $+7.67$pp accuracy and $+0.153$~F1, while InternVL gains $+8.15$pp and $+0.401$~F1, its recall more than doubling (0.325$\to$0.811) as it expands from 3.75 to 9.49 steps, exceeding even Qwen3-VL-32B's coverage (0.704 recall, Table~\ref{tab:main_results}) at $8\times$ fewer parameters. Both show a small LIS drop (longer chains add ordering noise) yet higher Ordered~F1, so content gains outweigh the ordering cost. With no architecture-specific reward tuning, CPR-Curriculum is not a Qwen-specific artifact but a reward formulation that transfers across multimodal architectures, improving reasoning without manual step annotation (qualitative examples in the supplementary material).

\begin{table}[!ht]
\centering
\caption{\textbf{GRPO with step-level rewards.} CPR-Curriculum improves both accuracy and Match F1 over the base model on two architecturally distinct backbones, using identical reward weights and protocol. All metrics use all-distilroberta-v1, $\tau\!=\!0.35$.}
\label{tab:grpo_results}
\footnotesize
\setlength{\tabcolsep}{4pt}
\resizebox{\linewidth}{!}{%
\begin{tabular}{l c c c c c c c}
\toprule
\textbf{Configuration} & \textbf{Acc (\%)} & \textbf{Match F1} & \textbf{Prec} & \textbf{Rec} & \textbf{Steps} & \textbf{LIS} & \textbf{Ord.\ F1} \\
\midrule
\multicolumn{8}{l}{\textit{Qwen2.5-VL-3B}} \\
\quad Baseline & 39.85 & 0.480 & 0.898 & 0.347 & 3.73 & 0.723 & 0.434 \\
\quad \textbf{CPR-Curriculum} & \textbf{47.52} & \textbf{0.633} & \textbf{0.963} & \textbf{0.493} & \textbf{5.10} & 0.626 & \textbf{0.560} \\
\quad $\Delta$ & +7.67 & +0.153 & +0.065 & +0.146 & +1.37 & $-$0.097 & +0.126 \\
\midrule
\multicolumn{8}{l}{\textit{InternVL3.5-4B}} \\
\quad Baseline & 37.61 & 0.432 & 0.895 & 0.325 & 3.75 & 0.775 & 0.387 \\
\quad \textbf{CPR-Curriculum} & \textbf{45.76} & \textbf{0.833} & \textbf{0.903} & \textbf{0.811} & \textbf{9.49} & 0.562 & \textbf{0.719} \\
\quad $\Delta$ & +8.15 & +0.401 & +0.008 & +0.486 & +5.74 & $-$0.213 & +0.332 \\
\bottomrule
\end{tabular}}
\end{table}


\begin{figure}[!ht]
\centering
\begin{minipage}[t]{0.48\linewidth}
    \centering
    \includegraphics[width=\linewidth]{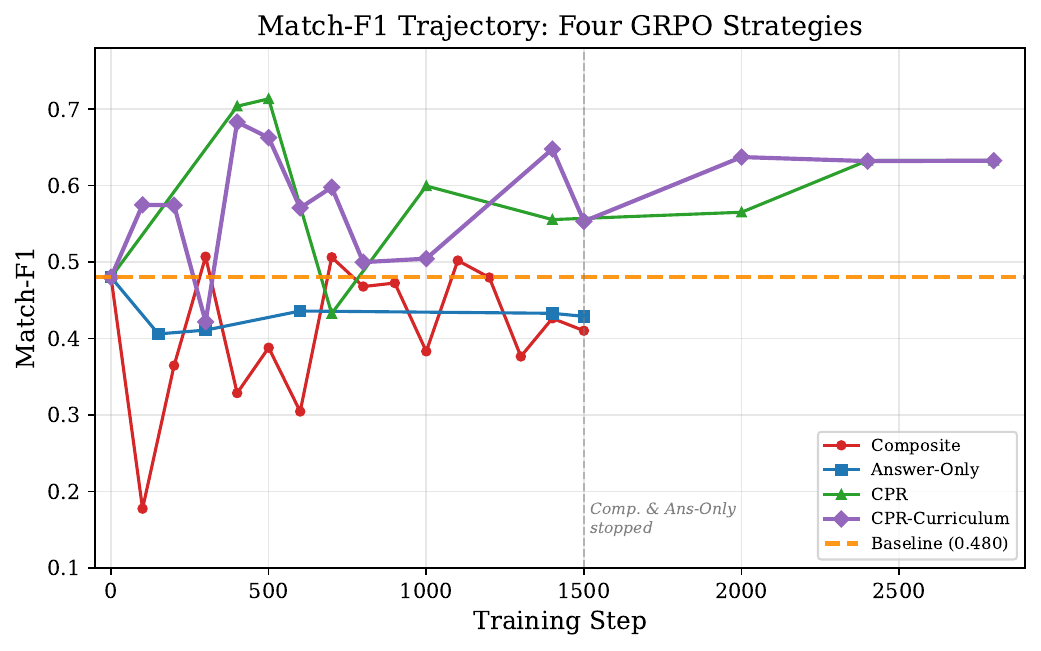}
\end{minipage}
\hfill
\begin{minipage}[t]{0.48\linewidth}
    \centering
    \includegraphics[width=\linewidth]{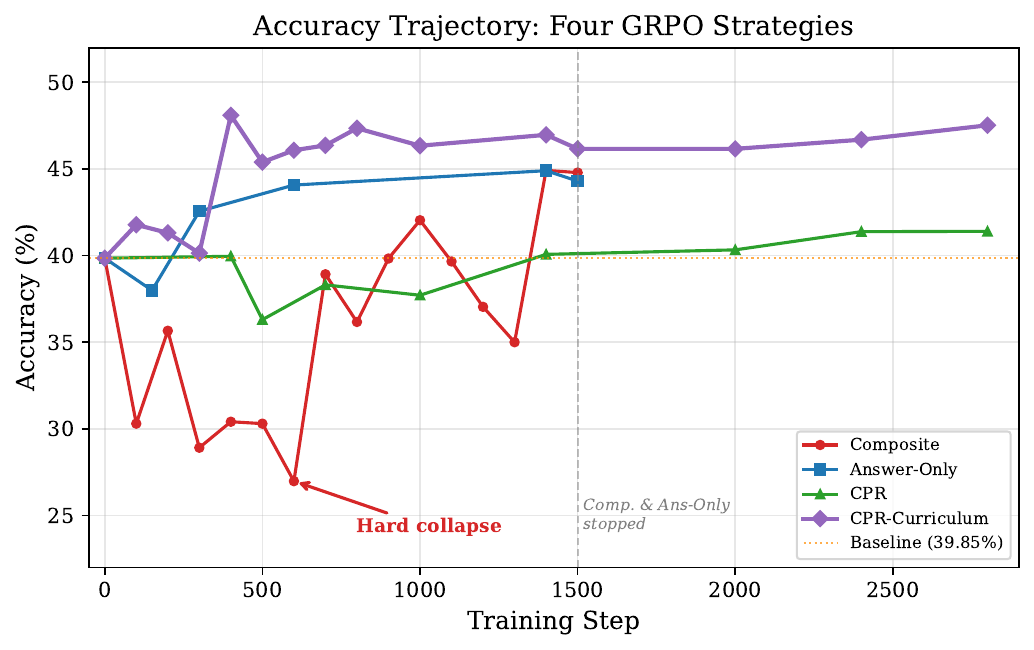}
\end{minipage}
\caption{\textbf{Training dynamics.} \textit{Left:} Match F1; \textit{Right:} Accuracy. Composite and Answer-Only diverge to NaN gradients by step~1,500 (Composite collapses at~600), while CPR variants train stably through 2,800 steps.}
\label{fig:grpo_trajectories}
\end{figure}



\subsubsection{Reward Strategy Comparison.}

We compare four reward strategies (Table~\ref{tab:grpo_strategy}) to isolate the effect of step-level supervision. The \textit{Composite} reward combines format, accuracy, and reasoning terms additively ($R = w_f R_{\text{fmt}} + w_a R_{\text{acc}} + w_r R_{\text{reason}}$), allowing the model to maximize each independently, achieving high accuracy through guessing while producing minimal reasoning. \textit{Answer-Only} removes reasoning ($w_r\!=\!0$) as a control. CPR and CPR-Curriculum use multiplicative coupling (Eq.~\ref{eq:cpr}) with initial weights $a_w\!=\!0.65, s_w\!=\!0.35$; we ablate this choice in Table~\ref{tab:grpo_weights}.

\begin{table}[!ht]
\centering
\caption{\textbf{Reward strategy comparison.} Best checkpoint per strategy on CRYSTAL test set. CPR strategies use weights $a_w\!=\!0.65, s_w\!=\!0.35$; a full weight ablation is presented in Table~\ref{tab:grpo_weights}. Composite and Answer-Only reach comparable accuracy but fail to improve reasoning (F1 $\leq$ 0.43). CPR-based strategies improve both.}
\label{tab:grpo_strategy}
\small
\setlength{\tabcolsep}{3.5pt}
\begin{tabular}{l c c c c c c}
\toprule
\textbf{Strategy} & \textbf{Acc (\%)} & \textbf{Match F1} & \textbf{Prec} & \textbf{Rec} & \textbf{LIS} & \textbf{Ord.\ F1} \\
\midrule
Baseline & 39.85 & 0.480 & 0.898 & 0.347 & 0.723 & 0.434 \\
Composite & \underline{44.92} & 0.426 & \textbf{0.983} & 0.284 & 0.743 & 0.392 \\
Answer-Only & 44.30 & 0.429 & 0.803 & 0.308 & 0.700 & 0.380 \\
CPR & 41.40 & \textbf{0.633} & \underline{0.975} & \underline{0.489} & 0.631 & \textbf{0.560} \\
CPR-Curriculum & \textbf{47.52} & \textbf{0.633} & 0.963 & \textbf{0.493} & 0.626 & \textbf{0.560} \\
\bottomrule
\end{tabular}
\end{table}


Answer-Only and Composite reach similar accuracy (44.30\% and 44.92\%) while reasoning stays flat (F1: 0.429 and 0.426), confirming the reasoning term is ignored under additive optimization. Composite collapses at step 600 (recall: 0.284); both strategies diverged to NaN gradients at step 1,500, requiring early termination (Figure~\ref{fig:grpo_trajectories}). CPR's multiplicative coupling resolves this: the model must produce both correct answers and aligned reasoning, achieving F1\,=\,0.633 (+32\%) and the highest Ordered F1 (0.560), while training stably through 2,800 steps. CPR-Curriculum further improves accuracy while maintaining comparable F1 and ordering quality~\cite{zhang2024mmcot,jiang2025mmecot}.

\subsubsection{Reward Weight Sensitivity.}
\label{sec:weight_sensitivity}

We vary $a_w$ and $s_w$ in Eq.~\ref{eq:cpr} across six configurations on a 15\% held-out validation split (4,546 samples; Table~\ref{tab:grpo_weights}). Beyond downstream performance, we report two diagnostics: the \emph{discrimination gap} (mean reward difference between correct and incorrect outputs) and the \emph{reward--reasoning correlation} $r$(F1) (Pearson correlation between reward and Match~F1). These reveal a fundamental trade-off: higher $s_w$ increases $r$(F1) but reduces discrimination. Both extremes are unstable: too answer-heavy ($a_w\!=\!0.80$) causes KL spikes from overconfident updates, while too step-heavy ($s_w\!\geq\!0.50$) elevates gradient norms from insufficient discrimination. Our configuration ($a_w\!=\!0.65,\,s_w\!=\!0.35$) achieves the highest accuracy (82.70\%), the lowest KL$_{\max}$ (0.372), and the highest $r$(F1) among stable configs (0.448), confirming it as the optimal balance.

\begin{table}[!ht]
\centering
\vspace{-0.3cm}
\caption{\textbf{Reward weight sensitivity.} Six configurations on a 15\% held-out split (4,546 samples). Both extremes are unstable ($\nabla\!\uparrow$); KL$_{\max}$ is U-shaped with its minimum at $a_w\!=\!0.65$, which also gives the highest accuracy and $r$(F1) in the stable range.}
\label{tab:grpo_weights}
\small
\setlength{\tabcolsep}{2.8pt}
\begin{tabular}{cc ccc cc cc}
\toprule
& & \multicolumn{3}{c}{\textbf{Val Performance}} & \multicolumn{2}{c}{\textbf{Reward Signal}} & \multicolumn{2}{c}{\textbf{Stability}} \\
\cmidrule(lr){3-5} \cmidrule(lr){6-7} \cmidrule(lr){8-9}
$a_w$ & $s_w$ & Acc (\%) & M-F1 & Rec & Discrim. & $r$(F1) & KL$_{\max}$ & $\nabla$ \\
\midrule
0.80 & 0.20 & 81.79 & 0.676 & 0.553 & 0.865 & 0.338 & 1.274 & \textcolor{red}{$\uparrow$} \\
0.70 & 0.30 & 82.61 & 0.731 & 0.627 & 0.797 & 0.410 & 0.456 & $\downarrow$ \\
\rowcolor{gray!12}
0.65 & 0.35 & \textbf{82.70} & 0.728 & 0.634 & 0.763 & \underline{0.448} & \textbf{0.372} & $\downarrow$ \\
\midrule
0.50 & 0.50 & 79.39 & 0.730 & 0.629 & 0.662 & 0.563 & 1.218 & \textcolor{red}{$\uparrow$} \\
0.35 & 0.65 & 78.53 & \textbf{0.736} & 0.642 & 0.560 & 0.670 & 1.684 & \textcolor{red}{$\uparrow$} \\
0.20 & 0.80 & 74.81 & 0.721 & \textbf{0.647} & 0.459 & \textbf{0.752} & 1.521 & \textcolor{red}{$\uparrow$} \\
\bottomrule
\end{tabular}
\\[2pt]
{\scriptsize \textcolor{red}{$\uparrow$}\,=\,increasing gradient norms (unstable). \textbf{Bold}\,=\,best per column; \underline{underline}\,=\,best within viable range (gap\,$\geq$\,0.70). \colorbox{gray!12}{Shaded}\,=\,selected.}
\end{table}
\vspace{-0.3cm}

\section{Conclusion}
\label{sec:conclusion}

We introduced CRYSTAL, a benchmark that evaluates multimodal reasoning step by step via Match~F1 and Ordered Match~F1. Across 20 MLLMs, we find universal cherry-picking, non-monotonic scaling trade-offs, and disordered reasoning. CPR-Curriculum achieves +32\% Match~F1 via GRPO where additive strategies fail. We publicly release the CRYSTAL benchmark, reference steps, and evaluation code to support reproducible research.

\noindent\textbf{Limitations.} References may not capture all valid paths. The fixed encoder and threshold are robust (Section~\ref{sec:ablation}) but domain-specific alternatives may help. We validate CPR on two architectures (Qwen2.5-VL-3B, InternVL3.5-4B) but only at the 3--4B scale; behavior at larger scale remains open, and Ordered Match~F1 does not yet model causal step dependencies.

\section*{Acknowledgments}
This work was supported by startup funds provided by Dartmouth College. The authors also acknowledge support from the National Science Foundation under CAREER Award No. 2541968.

\bibliographystyle{splncs04}
\bibliography{main}

\ifx\ifstandalonesupp\undefined
  \clearpage
  \renewcommand{\thepage}{S\arabic{page}}
  \setcounter{page}{1}
  \begin{center}
    {\Large\bfseries \papertitle\par}
    \vspace{0.5em}
    {\large\bfseries Supplementary Material\par}
  \end{center}
  \vspace{1em}
\fi

\renewcommand{\thefigure}{S\arabic{figure}}
\renewcommand{\thetable}{S\arabic{table}}
\renewcommand{\thesection}{S\arabic{section}}
\renewcommand{\thesubsection}{S\arabic{section}.\arabic{subsection}}
\renewcommand{\theequation}{S\arabic{equation}}
\renewcommand{\theHsection}{S\arabic{section}}
\renewcommand{\theHequation}{S\arabic{equation}}

\setcounter{figure}{0}
\setcounter{table}{0}
\setcounter{equation}{0}
\setcounter{section}{0}
\setcounter{subsection}{0}

\noindent This supplementary material complements the main paper with full derivations, implementation details, and additional examples. Sections~\ref{sec:training_data}--\ref{sec:implementation_details} cover the GRPO training dataset construction and the multi-agent reasoning generation pipeline with complete workflow diagrams and prompt templates. Section~\ref{sec:complexity_details} formalizes the complexity stratification, and Section~S4 specifies the evaluation protocol. Section~\ref{sec:rl-grpo} provides the full GRPO mathematical formulation, reward design, and training protocol. Section~S6 presents qualitative examples from both benchmark evaluation and GRPO before/after comparisons. Section~\ref{sec:impl_details} consolidates all hyperparameters and infrastructure for reproducibility. Sections~S8--S9 provide additional CRYSTAL and GRPO training examples.

\section{Training Dataset Construction}
\label{sec:training_data}

We construct a training dataset of 30,312 examples from ScienceQA-IMG~\cite{lu2022scienceqa} (6,218) and TextVQA~\cite{textvqa} (24,094), each containing an image, question, and ground-truth answer. For each example, the same four generators used in Section~3.1 independently produce candidate reasoning steps, which are pooled and semantically clustered into representative sequences (Phases~1 and 2). We omit Phases~3 and 4 (automated validation and human review) since the GRPO reward function (Section~\ref{sec:rl-grpo}) provides implicit quality control: poorly aligned steps receive lower Match~F1 scores, naturally down-weighting their influence during optimization.

\section{Multi-Agent Implementation Details}
\label{sec:implementation_details}

We provide the specific model configurations and hyperparameters used in our multi-agent reasoning generation pipeline (Section 3.1), ensuring reproducibility and transparency in benchmark construction.

\subsection{Complete Pipeline Workflow}
\label{sec:complete_workflow}

Figure~\ref{fig:workflow_complete} illustrates the complete pipeline referenced in Section~3.1. We detail each phase below, focusing on hyperparameters and implementation choices not covered in the main paper.

\begin{figure}[t]
\centering
\includegraphics[trim=0 64bp 64bp 0,clip,width=\textwidth]{images/workflow_dataset.png}
\caption{\textbf{Complete multi-agent reasoning generation workflow.} The pipeline processes input triplets $(Q, I, A)$ through four phases with two quality gates. \textit{Phase 1} generates candidate steps from four independent MLLMs. \textit{Phase 2} clusters steps via connected components and selects representative medoids. \textit{Phase 3} validates with a fifth MLLM; failures trigger Iteration Loop~1. \textit{Phase 4} applies human review; failures trigger Iteration Loop~2 (full restart). This iterative refinement implements Delphi-style consensus building~\cite{nasa2021delphi}.}
\label{fig:workflow_complete}
\end{figure}

\paragraph{Phase 1: Independent Generation.}
Four generators from distinct families (Qwen-VL, InternVL, Gemma, Llama; 17B to 76B parameters) each receive the triplet $(Q, I, A)$ and produce atomic reasoning steps as a JSON list. All generators use temperature $T\!=\!1.0$ with independent random seeds per example to maximize step diversity. The structured prompt requests minimal, logically ordered steps (full template in Section~\ref{sec:prompt_specs}).

\paragraph{Phase 2: Semantic Clustering.}
Candidate steps from all generators (typically 20 to 60 per question) are embedded with \texttt{roberta-large}~\cite{roberta} and connected into an undirected similarity graph at threshold $\tau\!=\!0.45$ (higher than the evaluation threshold $\tau\!=\!0.35$ to ensure only strongly equivalent steps merge). Connected components define semantic clusters; for each cluster, we select the medoid (Eq.~1 in the main paper) as the representative step, preserving natural language quality. Representatives are ordered by minimizing edit distance to the previous Delphi round's sequence, with topological sorting enforcing hard dependencies (e.g., ``identify object $X$'' before ``count instances of $X$''). Figure~\ref{fig:clustering_example} illustrates this process.

\begin{figure}[t]
\centering
\includegraphics[width=0.75\textwidth]{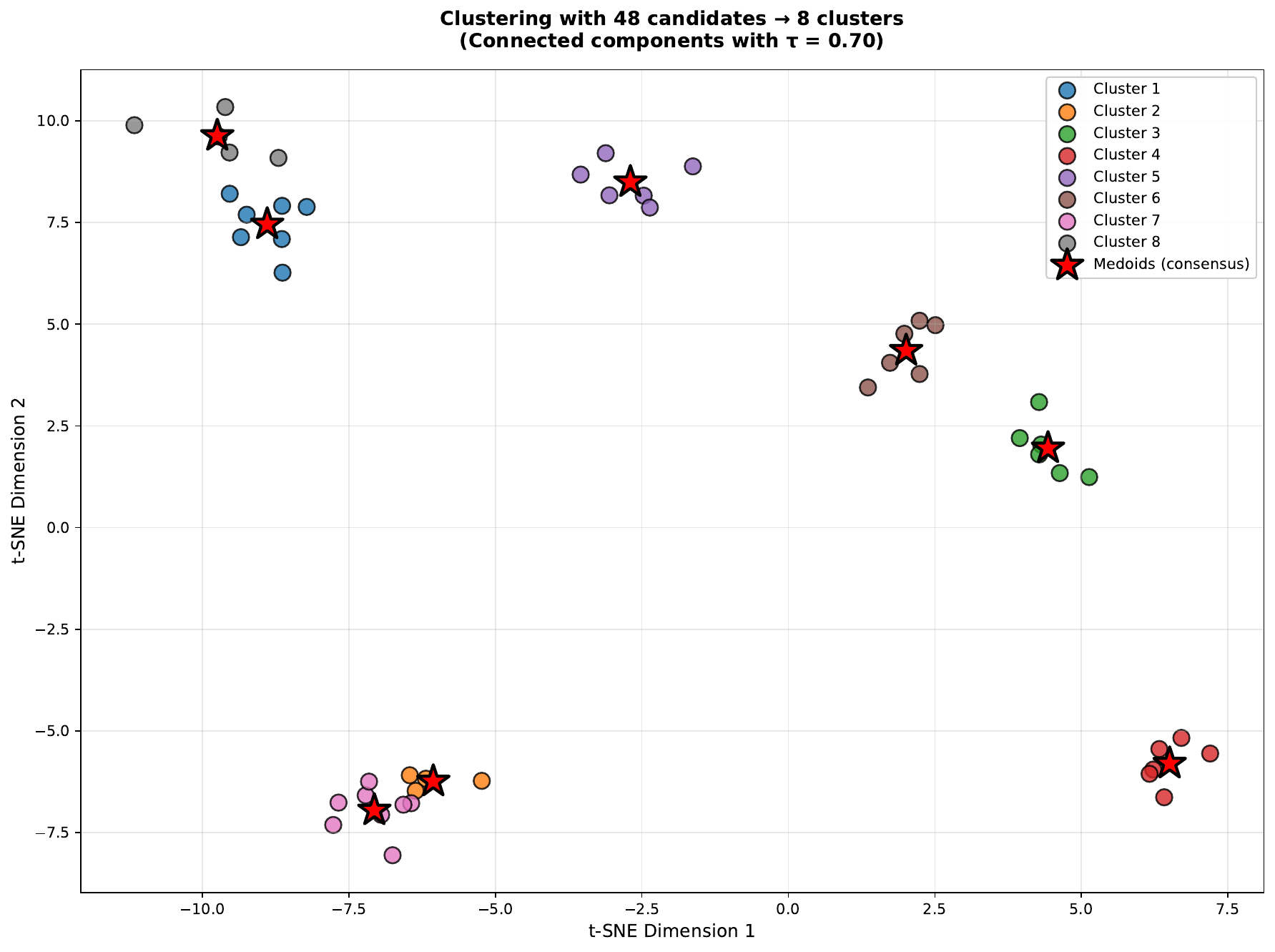}
\caption{\textbf{Semantic clustering visualization (Phase 2).} Scatter plot showing 48 candidate reasoning steps from four VLM generators clustered via connected components. Each colored point represents a candidate step embedded in 768-dimensional space and projected to 2D using t-SNE. Points are colored by cluster membership, with 8 clusters formed where edges connect steps with cosine similarity $\geq \tau = 0.70$. Red stars mark cluster medoids, selected to minimize average dissimilarity within each cluster. Each cluster contains multiple paraphrastic variations (average 6 steps per cluster), demonstrating how semantically equivalent steps naturally group together. This example illustrates redundancy elimination ($48 \rightarrow 8$ steps, 83.3\% reduction) while preserving distinct reasoning operations.}
\label{fig:clustering_example}
\end{figure}

\paragraph{Phase 3: Automated Validation.}
A fifth MLLM (Molmo-72B~\cite{molmo}), architecturally distinct from the generators, validates the reasoning chain against four criteria:
\begin{enumerate}[leftmargin=*,nosep]
\item \textbf{Logical soundness:} no contradictions between steps.
\item \textbf{Sequential coherence:} no unexplained reasoning jumps.
\item \textbf{Visual grounding:} perceptual claims verifiable in image $I$.
\item \textbf{Answer consistency:} executing the steps yields answer $A$.
\end{enumerate}
The validator outputs a structured JSON response with a pass/fail decision and step-level justifications. Failed examples re-enter Phase~1 with fresh seeds (Iteration Loop~1, max 2 attempts).

\paragraph{Phase 4: Human Quality Gate.}
Trained annotators verify the same four criteria using a custom Gradio~\cite{abid2019gradio} interface (Figure~\ref{fig:annotation_ui}), which displays the input triplet alongside candidate steps with inline editing and navigation controls. Annotation consistency is verified through dual-annotation on a subset with adjudication by a senior annotator. Rejected examples restart the full pipeline (Iteration Loop~2). Fewer than 5\% of examples require any re-iteration.

\begin{figure}[t]
\centering
\begin{tabular}{@{}c@{\hspace{0.5cm}}c@{}}
\includegraphics[width=0.48\textwidth]{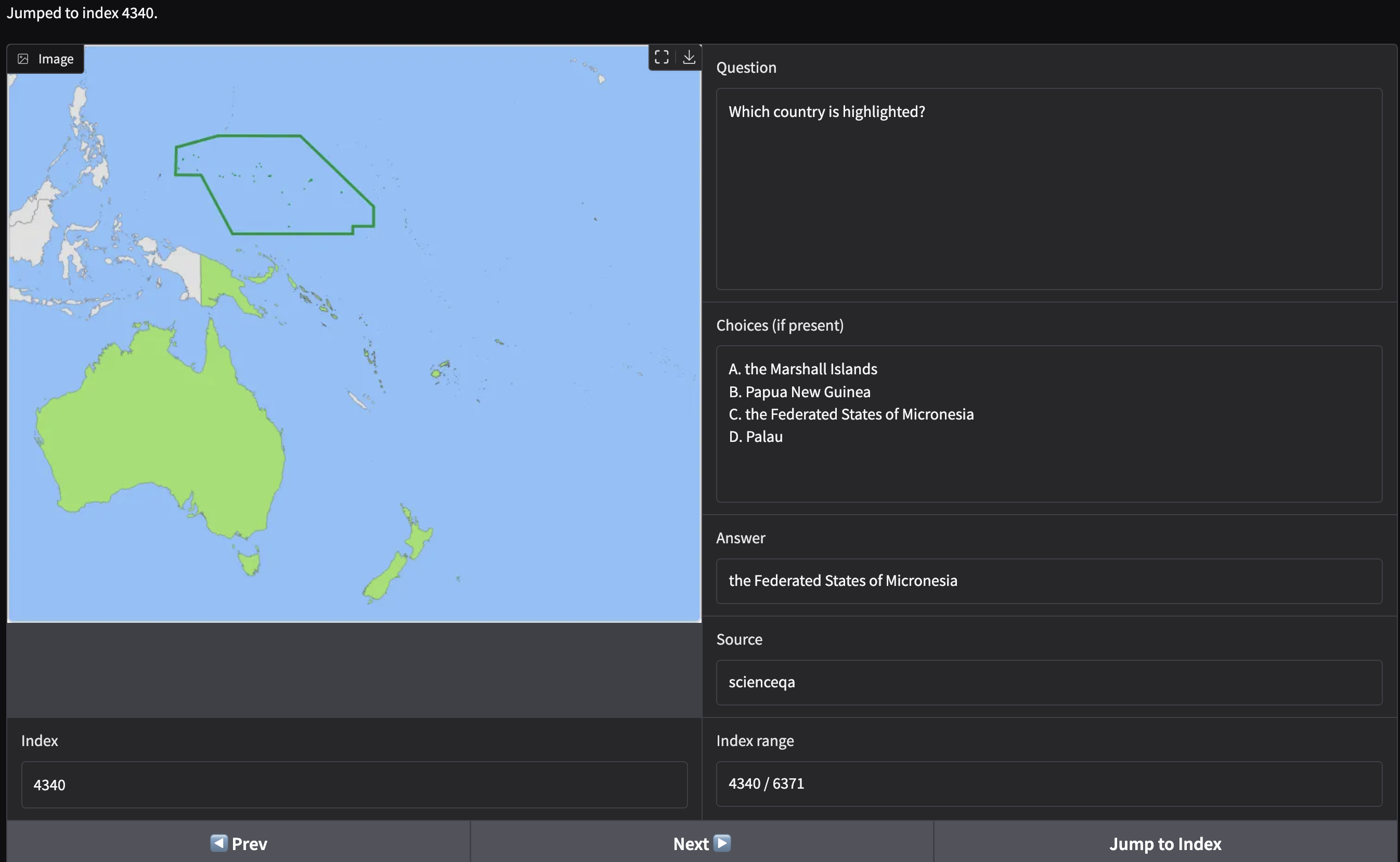} &
\includegraphics[width=0.48\textwidth]{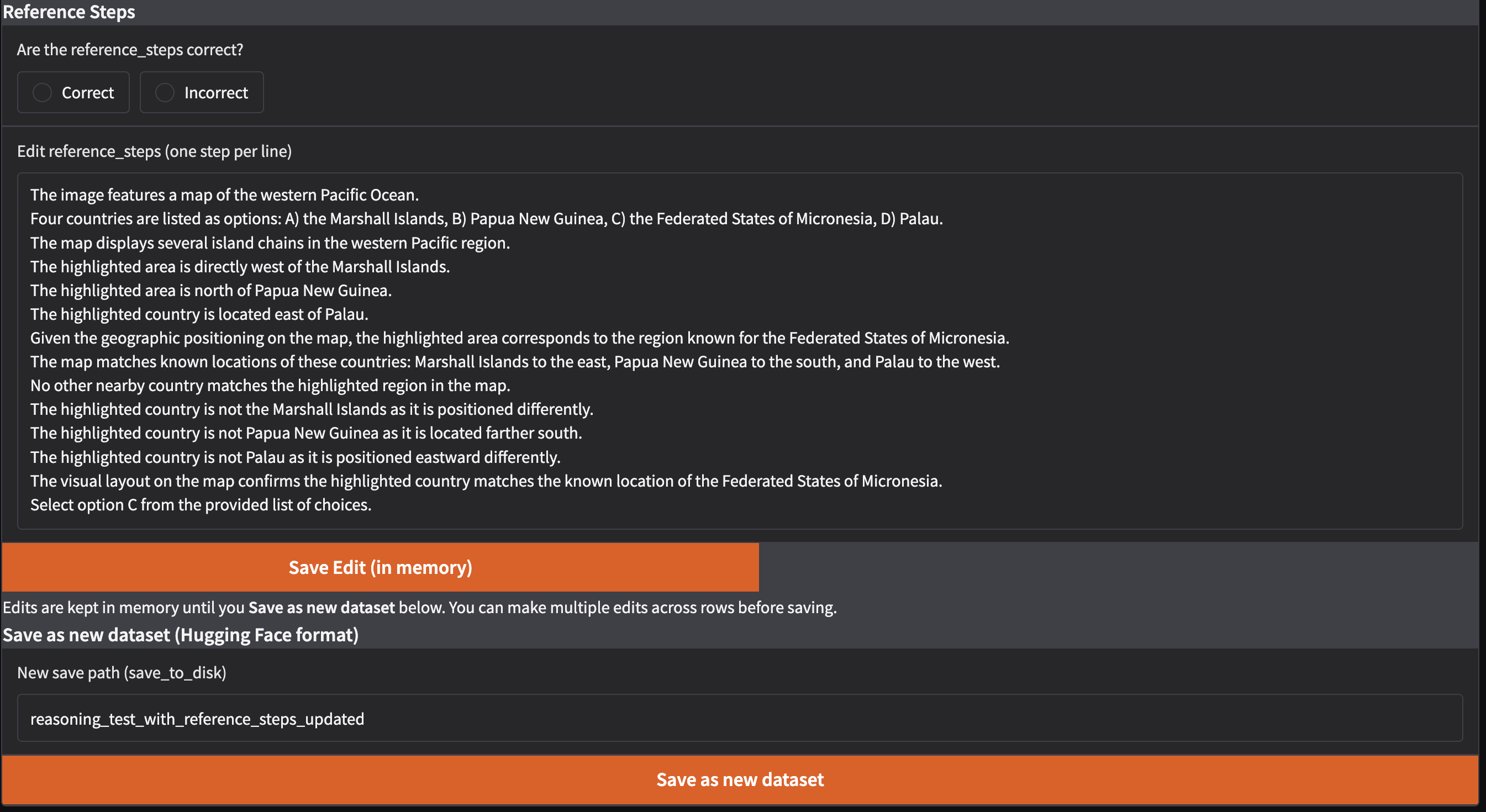} \\
(a) Question display panel & (b) Validation and editing panel
\end{tabular}
\caption{\textbf{Human validation interface (Phase 4).} Screenshot of the Gradio-based annotation tool used for quality gate verification. \textbf{(a)} The left panel displays the input triplet: image, question text, multiple-choice options (if present), ground-truth answer, and dataset source. Navigation controls (Prev/Next buttons and index slider) allow annotators to traverse the dataset sequentially or jump to specific examples. \textbf{(b)} The right panel presents the candidate reasoning steps produced by Phases 1--3, with radio buttons to mark correctness and a text area for editing steps. Annotators can save edits in memory and batch-export corrected examples to a new dataset. This interface enables efficient human review of 6,372 validation examples.}
\label{fig:annotation_ui}
\end{figure}

\subsection{Prompt Specifications}
\label{sec:prompt_specs}

For reproducibility, we provide the exact prompts used in dataset construction and model evaluation. The two prompts differ in three key ways: (1)~the generation prompt receives the ground-truth answer $A$ while the evaluation prompt does not, (2)~generation outputs a Python list (\texttt{reference\_steps = [...]}) while evaluation enforces JSON, and (3)~generation uses $T\!=\!1.0$ with independent seeds for diversity while evaluation uses greedy decoding ($T\!=\!0.0$) for determinism. Both prompts are applied identically across all models in their respective contexts. Domain-specific variants (MathVision, ScienceQA, MMVP, PlotQA) adjust the context section but maintain the same output format and grounding constraints.

Figure~\ref{fig:generation_prompt} shows the generation prompt (RealWorldQA variant) used by the four Phase~1 generators. Figure~\ref{fig:evaluation_prompt} shows the evaluation prompt applied to all 20 benchmarked models in Table~2 in the main paper.

\begin{figure}[p]
\centering
\begin{tcolorbox}[
    colback=blue!5,
    colframe=blue!40,
    title=System Prompt for Dataset Generation (RealWorldQA Example),
    fonttitle=\small\bfseries,
    width=\textwidth,
]
\begin{scriptsize}
\begin{verbatim}
You are given:
1) User question/instruction: {{QUESTION}}
2) A natural, real-world image (or image pair), often captured
   from vehicles or everyday outdoor scenes. No separate
   annotations are provided.
3) Correct final answer to the question: {{ANSWER}}

CONTEXT
The RealWorldQA benchmark evaluates real-world visual understanding.
Images depict genuine environments: urban streets, highways, rural
areas, parking lots, interiors, or other everyday contexts.

ROLE
Inspect the photo(s) directly and report only concise, observable
evidence items that justify the answer. Evidence should include:
- Objects & attributes: colors, shapes, textures, materials, sizes,
  distinct markings.
- Scene layout: road type, background structures, horizon, sky
  conditions, environment type.
- Spatial relations: left/right, above/below, foreground/background,
  near/far, blocking/occluding.
- Object relationships: car behind truck, pedestrian crossing in
  front of vehicle, bike leaning against wall.
- Commonsense & physics: object motion implied by posture/blur,
  shadow direction implying sun position, physical feasibility.
- Lighting & environment cues: daylight, nighttime, reflections,
  shadows, occlusions, perspective.

TASK
Produce a compact decision trace: a short list of atomic evidence
checks that justify the provided answer. Each item must be a
standalone verification grounded in what is visible.

HARD CONSTRAINTS
- Output ONLY: reference_steps = [ "<item 1>", "<item 2>", ... ]
- Every element MUST be a double-quoted string.
- No code fences, no headings, no paragraphs, no bullets.
- Do NOT restate {{ANSWER}} verbatim in any item.
- Keep each item concise and factual; 15-20 items is typical.
- The final item should state the required final action per the
  question format (e.g., "Select option B", "Return the numeric
  value", "Answer True/False").
\end{verbatim}
\end{scriptsize}
\end{tcolorbox}
\caption{\textbf{Dataset generation prompt (RealWorldQA variant).} The four Phase~1 generators receive the question and ground-truth answer and output a Python list of atomic evidence checks. Similar templates are adapted per source dataset.}
\label{fig:generation_prompt}
\end{figure}

\begin{figure}[p]
\centering
\begin{tcolorbox}[
    colback=gray!5,
    colframe=gray!40,
    title=System Prompt for Model Evaluation (Prediction Generation),
    fonttitle=\small\bfseries,
    width=\textwidth,
]
\begin{scriptsize}
\begin{verbatim}
You are a vision-language model. Analyze the provided image(s)
and user text silently. Do NOT reveal internal reasoning.

Return ONLY a valid JSON object with this schema:
{"reasoning_steps": [], "answer": ""}

Rules for "reasoning_steps":
- Include enough steps to make the answer evident without filler.
- Write single-clause sentences, each adding a new, directly
  checkable fact or cue-based inference anchored to visible cues.
- No multi-sentence items. No internal monologue.

Rules for "answer":
- Ground strictly in visible content and given text.
- If ambiguous, set "answer" to "insufficient information".
- Multiple-choice: return only the best LETTER (e.g., "B").
- Numeric: include units; obey requested rounding.

What to notice in steps (as sentences, not labels):
- Objects & attributes (colors, materials, states, logos).
- Spatial relations (left/right, above/below, foreground/
  background, proximity).
- Depth cues (relative size, occlusion order, shadow contact).
- Scene & lighting cues (indoor/outdoor, daylight vs. night).
- Text/OCR with exact casing ("Read text: 'SPEED LIMIT 25'.").
- Counts & quantities; approximate only if visually justified.
- Graphics/plots: axes, ticks, units, legends; read exact values.

Output: only the JSON object. No extra keys, comments, or prose.

User instruction: {USER_INSTRUCTION}
\end{verbatim}
\end{scriptsize}
\end{tcolorbox}
\caption{\textbf{Model evaluation prompt.} Unlike generation, this prompt provides only the question (no answer) and enforces JSON output. Applied to all 20 models with greedy decoding ($T\!=\!0.0$).}
\label{fig:evaluation_prompt}
\end{figure}

\section{Complexity Stratification}
\label{sec:complexity_details}

Questions are stratified into three complexity tiers (\textit{easy}, \textit{medium}, \textit{hard}) using a model-agnostic scoring function based exclusively on ground-truth features. We compute a weighted complexity score $c \in [0,1]$ for each question as:
\begin{equation}
c = 0.4 \cdot s_{\text{steps}} + 0.2 \cdot s_{\text{length}} + 0.2 \cdot s_{\text{ling}} + 0.2 \cdot s_{\text{type}},
\end{equation}
where $s_{\text{steps}}$ reflects the normalized reference step count (primary indicator), $s_{\text{length}}$ captures question length and word count, $s_{\text{ling}}$ aggregates linguistic complexity markers (conditionals such as ``if/when'', causals such as ``because/therefore'', comparisons, negations), and $s_{\text{type}}$ adjusts for question type (e.g., counting questions reduce complexity by $-0.15$, ``why'' questions increase by $+0.20$). Questions with $c < 0.27$ are classified as \textit{easy}, $0.27 \le c < 0.42$ as \textit{medium}, and $c \ge 0.42$ as \textit{hard}. This approach ensures complexity assignment is independent of model predictions, enabling fair cross-model comparison and preventing circular evaluation bias. The resulting complexity distribution is: \textit{easy} (48.5\%, avg.~8.6 steps), \textit{medium} (44.1\%, avg.~13.3 steps), and \textit{hard} (7.4\%, avg.~20.5 steps), providing balanced coverage across difficulty levels.

\section{Benchmarking Settings}

\textbf{Evaluation Protocol.}
We assess models using two complementary metrics defined in Section~3.2 of the main paper: \textit{Accuracy} measures final answer correctness through fuzzy matching strategies (exact match, numeric comparison, choice-based matching, yes/no detection), while \textit{Match F1} evaluates alignment between predicted and reference reasoning steps via semantic similarity matching.
Match~F1 embeds reasoning steps with \texttt{all-distil\-roberta-v1}~\cite{reimers2019sbert} and applies a cosine similarity threshold $\tau\!=\!0.35$ for step matching (encoder and threshold selected via ablation, Section~4.4).

\noindent\textbf{Metric behavior.}
Match F1 focuses on the fidelity of the predicted reasoning at the level of atomic steps. It penalizes omissions and hallucinations through the precision and recall terms in Eq.~3 of the main paper. The score is sensitive to verbosity because unmatched steps increase the false positive count which lowers precision, and the macro-averaged F1 in Eq.~4 of the main paper reflects these per-example scores. Accuracy measures only whether the final answer is judged correct. It is insensitive to the path of reasoning that led to the answer. High Accuracy with low Match F1 indicates correct answers with misaligned or speculative steps. High Match F1 with low Accuracy indicates faithful reasoning traces that do not culminate in the correct final prediction.

\noindent Full implementation details (software versions, hardware, fixed parameters) are consolidated in Section~\ref{sec:impl_details} for reproducibility.



\section{Post-Training via Group Relative Policy Optimization}
\label{sec:rl-grpo}

This section provides the full mathematical formulation behind the GRPO training experiments in Section~4.5 of the main paper. We first describe the optimization framework (Section~\ref{sec:grpo_framework}), then detail the composite and CPR reward functions (Section~\ref{sec:grpo_reward}), and the training protocol (Section~\ref{sec:grpo_protocol}).

\subsection{GRPO Framework}
\label{sec:grpo_framework}

Given an input consisting of image $\mathcal{I}$ and question $q$, the MLLM policy $\pi_\theta$ generates a response containing atomic reasoning steps and a final answer. At each iteration, we sample $K$ candidates $\{o^{(k)}\}_{k=1}^{K}$ from the policy. Each candidate is parsed to extract predicted steps $\mathcal{P}_i$ and answer $\hat{y}_i$, which are evaluated against ground truth $\mathcal{G}_i$ and $y_i$ using the metrics in Section~3.2 of the main paper. GRPO~\cite{grpo} computes advantages relative to the group mean, eliminating a separate value model. The clipped surrogate objective is:
\begin{equation}
\begin{aligned}
\mathcal{L}_{\mathrm{GRPO}}(\theta)
= {}& \mathbb{E}_{(\mathcal{I},q)\sim\mathcal{D}}\Biggl[
  \frac{1}{K}\sum_{k=1}^{K}
  \min\Bigl(
    \rho^{(k)} A^{(k)},\\
  &\qquad\operatorname{clip}\bigl(\rho^{(k)}, 1-\epsilon, 1+\epsilon\bigr) A^{(k)}
  \Bigr)
\Biggr],
\end{aligned}
\end{equation}
where $\rho^{(k)} = \pi_\theta(o^{(k)} \mid \mathcal{I}, q) / \pi_{\theta_{\mathrm{old}}}(o^{(k)} \mid \mathcal{I}, q)$ is the importance ratio and $\epsilon$ the clipping threshold. We add a KL penalty $\lambda_{\mathrm{KL}} \, \mathbb{D}_{\mathrm{KL}}[\pi_\theta \,\|\, \pi_{\mathrm{ref}}]$ to prevent drift from the pretrained policy $\pi_{\mathrm{ref}}$.

\subsection{Reward Design}
\label{sec:grpo_reward}

\noindent\textbf{Output format.} The model must produce structured JSON:
\begin{equation}
\bigl\{\;\texttt{"reasoning\_steps"}:\,[\ldots],\;\texttt{"answer"}:\,\text{``...''}\;\bigr\},
\label{eq:format_template}
\end{equation}
where \texttt{reasoning\_steps} contains atomic step strings and \texttt{answer} the final prediction. We define a binary format indicator $F^{(k)} \in \{0,1\}$ that returns 1 if candidate $k$ is valid JSON conforming to Equation~\eqref{eq:format_template} with both required fields present.

\noindent\textbf{Composite reward.} Let $\mathrm{F1}_i^{(k)}$ denote the Match~F1 score for candidate $k$ (Eqs.~2--4 in the main paper) and $C(\hat{y}_i^{(k)}, y_i)$ the binary answer correctness predicate. To ensure commensurability, we normalize Match~F1 via batch statistics. The per-token reward is:
\begin{equation}
r^{(k)} \;=\; \alpha \, \frac{\mathrm{F1}_i^{(k)} - \mu_{\mathrm{F1}}}{\sigma_{\mathrm{F1}} + \delta} \;+\; \beta \, C(\hat{y}_i^{(k)}, y_i) \;+\; \gamma \, F^{(k)},
\label{eq:composite_reward}
\end{equation}
where $\mu_{\mathrm{F1}}$, $\sigma_{\mathrm{F1}}$ are batch mean and standard deviation, $\delta$ prevents division by zero, and $\alpha, \beta, \gamma \in \mathbb{R}_{>0}$ control emphasis on reasoning fidelity, correctness, and format compliance. The format term provides strong supervision during early training when the model may produce free-form text, with influence diminishing as compliance saturates. This additive formulation serves as the baseline reward (``Composite'' in Table~5 of the main paper).

\noindent\textbf{Causal Process Reward (CPR).} The composite reward allows the model to maximize each term independently, leading to correct answers with minimal reasoning. CPR replaces additive combination with multiplicative coupling that conditions the step-level reward on answer correctness:
\begin{equation}
r_{\text{CPR}}^{(k)} = \begin{cases}
  a_w \cdot 1.0 + s_w \cdot \mathrm{F1}_i^{(k)} & \text{if } C(\hat{y}_i^{(k)}, y_i) = 1, \\
  s_w \cdot \mathrm{F1}_i^{(k)} \cdot 0.3 & \text{otherwise},
\end{cases}
\label{eq:cpr_reward_supp}
\end{equation}
where $a_w$ and $s_w$ are the causal answer and step weights controlling the relative emphasis on correctness versus reasoning fidelity, and $\mathrm{F1}_i^{(k)}$ is a step-matching F1 that follows the same greedy 1:1 formulation as our evaluation metric (Eqs.~2--4 in the main paper) but replaces the sentence encoder with a lexical word-overlap similarity: common tokens counted with multiplicity and normalized by the geometric mean of the two step lengths, matched above a threshold of $0.3$. We use this lexical surrogate inside the training loop for two reasons: a sentence encoder in the reward path is unstable under DeepSpeed's multi-process generation, and the reward is evaluated for all $K$ candidates at every optimization step, where a deterministic token-level score is substantially cheaper. Evaluation always uses the encoder-based Match~F1 of Section~3.2 of the main paper, so the reward acts as a fast surrogate for the reported metric rather than the metric itself. The 0.3 discount factor for incorrect answers prevents the model from learning faithful reasoning chains that lead to wrong conclusions. We apply PCGrad~\cite{yu2020pcgrad} to resolve gradient conflicts between the accuracy and reasoning objectives during joint optimization.

\subsection{Training Protocol}
\label{sec:grpo_protocol}

CPR-Curriculum (Section~3.3 of the main paper) proceeds in two phases: an answer-only warm-up that establishes format compliance and basic accuracy, followed by CPR fine-tuning that introduces step-level reasoning supervision. This staged approach prevents the reasoning objective from destabilizing training before the model has learned to produce well-formatted, accurate answers. The training dataset comprises 30,312 examples with expert-generated step-by-step solutions produced by our multi-agent annotation pipeline. We apply both phases to Qwen2.5-VL-3B-Instruct~\cite{qwen2vl2024} and InternVL3.5-4B~\cite{internvl35} (training infrastructure and per-model hyperparameters in Section~\ref{sec:impl_details}).

At each iteration, we sample $K$ candidates and parse each as JSON per Equation~\eqref{eq:format_template}. Malformed outputs receive $F^{(k)}=0$ and minimal reward but still contribute gradients to discourage malformation. Valid completions are processed to extract $\mathcal{P}_i$ and $\hat{y}_i$ for evaluation. We monitor step count distributions and penalize anomalous verbosity relative to a running median to guard against reward hacking. Within Phase~2, progressive difficulty scheduling orders training examples by reference step count, starting with simpler chains and gradually introducing complex multi-hop reasoning. Validation tracking of Match~F1, Accuracy, and format compliance ensures no metric degrades during optimization.


\section{Qualitative Analysis}

We present four representative examples drawn from the full set of model predictions on CRYSTAL, illustrating the failure modes identified in Findings~1--5 of the main paper. Each example maps to a distinct region in the accuracy--F1 space, together spanning the full spectrum from perfect alignment to catastrophic breakdown.

\begin{figure}[t]
\centering

{\small\bfseries Input Image}\par
\vspace{1pt}
\fcolorbox{black!40}{white}{%
  \includegraphics[width=0.5\linewidth]{images/supp_perfect_sample412.jpg}%
}

\vspace{0.08cm}

\begin{tcolorbox}[
    width=\linewidth,
    colback=blue!5,
    colframe=blue!50,
    boxrule=1pt,
    arc=2mm,
    left=3pt,
    right=3pt,
    top=2pt,
    bottom=2pt
]
{\small
\centering
\textbf{\textcolor{blue!70!black}{Question:}} What color is the traffic light? Please answer directly with a single word or number.\\[2pt]
\textbf{Ground Truth:} \colorbox{green!20}{\textbf{Green}}
}
\end{tcolorbox}

\vspace{0.08cm}

\begin{tcolorbox}[
    width=\linewidth,
    colback=green!5,
    colframe=green!60!black,
    boxrule=1.2pt,
    arc=2mm,
    left=3pt,
    right=3pt,
    top=2pt,
    bottom=2pt,
    title={\centering\small\textbf{Qwen3-VL-32B Prediction}},
    fonttitle=\bfseries
]
{\small
\textbf{Answer:} \colorbox{green!20}{\textbf{green}} $\checkmark$ \quad
\textbf{Match F1:} 1.000 \quad \textbf{Precision:} 1.000 \quad \textbf{Recall:} 1.000

\vspace{0.04cm}
\textbf{Reasoning Steps (8 generated, 8 reference):}
\begin{enumerate}[leftmargin=10pt,itemsep=0pt,parsep=0pt,topsep=1pt]
  \item Two traffic lights are visible above the intersection, mounted on a horizontal pole.
  \item Each traffic light has three circular lenses arranged vertically.
  \item The top and middle lenses are dark and unlit.
  \item The bottom lens in each traffic light is illuminated with a bright green light.
  \item The green illumination is consistent across both traffic lights.
  \item The green color is clearly visible despite the overcast lighting conditions.
  \item The traffic lights are positioned above the crosswalk and street, indicating they are for vehicular traffic.
  \item The green light suggests that vehicles may proceed through the intersection.
\end{enumerate}

\vspace{0.06cm}
\centering
\fcolorbox{green!70}{green!15}{%
  \parbox{0.96\linewidth}{%
    \centering\small
    \textbf{\textcolor{green!60!black}{Perfect Alignment:}} All 8 steps match references with comprehensive coverage
  }%
}
\vspace{-2pt}
}
\end{tcolorbox}

\vspace{0.04cm}

\caption{\textbf{Perfect reasoning example.}
The model achieves F1 = 1.0 by systematically identifying visual elements, verifying illumination state, and concluding with the correct answer. All 8 predicted steps match the 8 references, demonstrating comprehensive reasoning without shortcuts.}
\label{fig:supp_perfect}
\end{figure}

\noindent\textbf{Perfect reasoning is achievable but rare.} Figure~\ref{fig:supp_perfect} shows Qwen3-VL-32B on a traffic light question, generating exactly 8 steps that match all 8 references (Precision = Recall = 1.000, F1 = 1.0). This outcome is uncommon and concentrates on perceptually simple questions with short reasoning chains, confirming that current models \textit{can} produce faithful reasoning when cognitive load is low~\cite{chen2024measuring,xu2024llavacot}. The rarity of perfect alignment, even for the strongest models, underscores how demanding full step coverage is relative to answer correctness alone.

\begin{figure}[t]
\centering

{\small\bfseries Input Image}\par
\vspace{1pt}
\fcolorbox{black!40}{white}{%
  \includegraphics[width=0.75\linewidth]{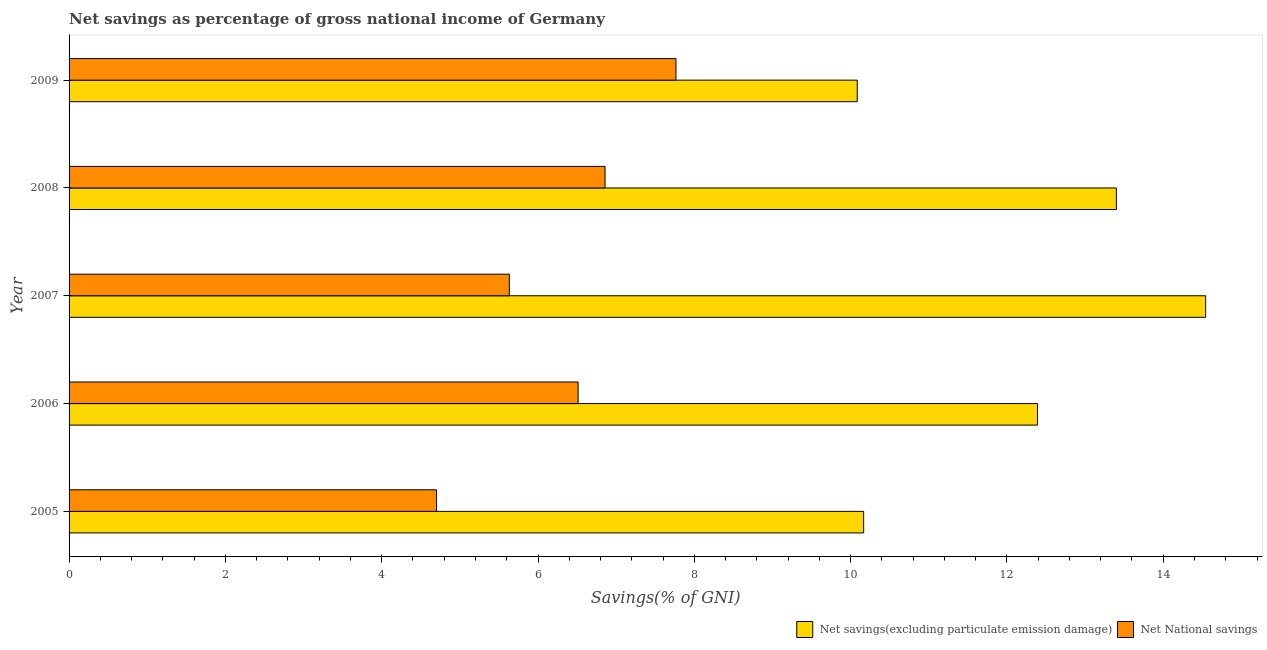}%
}

\vspace{0.08cm}

\begin{tcolorbox}[
    width=\linewidth,
    colback=blue!5,
    colframe=blue!50,
    boxrule=1pt,
    arc=2mm,
    left=3pt,
    right=3pt,
    top=2pt,
    bottom=2pt
]
{\small
\centering
\textbf{\textcolor{blue!70!black}{Question:}} In how many years is the net national savings greater than the average net national savings taken over all years?

\vspace{0.06cm}
\textbf{Ground Truth:} \colorbox{green!20}{\textbf{3 years (2006, 2008, 2009)}}
}
\end{tcolorbox}

\vspace{0.08cm}

\begin{tcolorbox}[
    width=\linewidth,
    colback=red!5,
    colframe=red!60,
    boxrule=1.2pt,
    arc=2mm,
    left=3pt,
    right=3pt,
    top=2pt,
    bottom=2pt,
    title={\centering\small\textbf{Gemma3-12B Prediction}},
    fonttitle=\bfseries
]
{\small
\textbf{Answer:} \colorbox{red!20}{\textbf{2 years}} $\times$ \quad
\textbf{Match F1:} 0.857 \quad \textbf{Precision:} 0.750 \quad \textbf{Recall:} 1.000

\vspace{0.04cm}
\textbf{Reasoning Steps (8 generated, 6 reference):}
\begin{enumerate}[leftmargin=10pt,itemsep=0pt,parsep=0pt,topsep=1pt]
  \item The graph displays net national savings and net savings over several years.
  \item The x-axis represents the years from 2005 to 2009.
  \item The y-axis represents savings as a percentage of GNI.
  \item \textcolor{red!70!black}{\textbf{[ERROR] The average net national savings appears to be approximately 9.5\% of GNI.}} \quad \textit{[Incorrect: true average is 6.294\%]}
  \item \textcolor{red!70!black}{\textbf{[ERROR] In 2007, the net national savings are approximately 13.5\% of GNI.}} \quad \textit{[Incorrect: true value is 5.633\%]}
  \item In 2008, the net national savings are approximately 13.5\% of GNI.
  \item The net national savings are greater than the average in 2007 and 2008.
  \item Therefore, the net national savings are greater than the average in two years.
\end{enumerate}

\vspace{0.06cm}
\centering
\fcolorbox{orange!70}{orange!15}{%
  \parbox{0.96\linewidth}{%
    \centering\small
    \textbf{\textcolor{orange!80!black}{Wrong+Sound (F1=0.857, Precision=0.750, Recall=1.000):}} Follows complete reasoning pipeline (identify→read→calculate→compare→count), matching ALL 6 reference step types. However, \textbf{Precision=0.750 reveals 2 of 8 steps contain numerical errors}: miscalculates average as 9.5\% (true: 6.294\%, 51\% off) and misreads 2007 value as 13.5\% (true: 5.633\%). These errors propagate to final count. \textbf{Demonstrates Recall measures reasoning structure, Precision detects quantitative accuracy.}
  }%
}
\vspace{-2pt}
}
\end{tcolorbox}

\vspace{0.04cm}

\caption{\textbf{Sound reasoning structure with numerical errors (F1\,=\,0.857).} Gemma3-12B covers all 6 reference step types (Recall\,=\,1.000) but two steps contain numerical errors (Precision\,=\,0.750): miscalculating the average as 9.5\% (true: 6.294\%) and misreading 2007 as 13.5\% (true: 5.633\%). These perceptual errors propagate through valid comparison logic, yielding answer ``2'' instead of ``3''. Recall confirms correct algorithmic structure; Precision detects execution failures in visual data extraction.}
\label{fig:supp_wrongsound}
\end{figure}

\noindent\textbf{Sound reasoning does not guarantee correct answers.} Figure~\ref{fig:supp_wrongsound} shows Gemma3-12B on a chart analysis task, covering the complete reasoning pipeline (Recall = 1.000) yet answering incorrectly. The model follows the right structure (identify chart type, compute average, read yearly values, compare, count) but two steps contain numerical errors: miscomputing the average as 9.5\% (true: 6.294\%) and misreading a data point as 13.5\% (true: 5.633\%). These perceptual-computational errors propagate through otherwise valid logic, yielding Precision = 0.750 and F1 = 0.857. This pattern is notably common, more frequent than fully perfect reasoning, and reveals a failure mode invisible to accuracy metrics: the model possesses the correct algorithmic structure but fails at visual data extraction~\cite{chen2024measuring,tong2024mmvp}. Such cases suggest that targeted improvements in numerical grounding could recover accuracy without retraining the reasoning framework.

\begin{figure}[t]
\centering

{\small\bfseries Input Image}\par
\vspace{1pt}
\fcolorbox{black!40}{white}{%
  \includegraphics[width=0.6\linewidth]{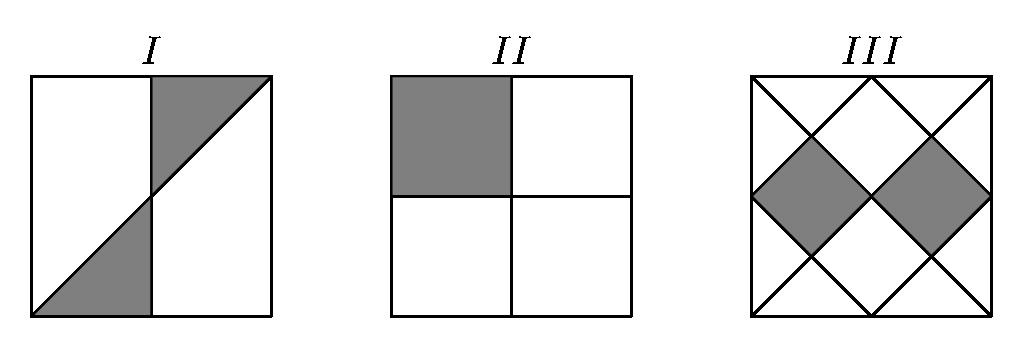}%
}

\vspace{0.08cm}

\begin{tcolorbox}[
    width=\linewidth,
    colback=blue!5,
    colframe=blue!50,
    boxrule=1pt,
    arc=2mm,
    left=3pt,
    right=3pt,
    top=2pt,
    bottom=2pt
]
{\small
\centering
\textbf{\textcolor{blue!70!black}{Question:}} Each of the three large squares shown below is the same size. Segments that intersect the sides of the squares intersect at the midpoints of the sides. How do the shaded areas of these squares compare?

\vspace{0.08cm}
\begin{tabular}{@{}l@{}}
A) $\text{The shaded areas in all three are equal.}$ \\
B) $\text{Only the shaded areas of }I\text{ and }II\text{ are equal.}$ \\
C) $\text{Only the shaded areas of }I\text{ and }III\text{ are equal.}$ \\
D) $\text{Only the shaded areas of }II\text{ and }III\text{ are equal.}$ \\
E) $\text{The shaded areas of }I, II\text{ and }III\text{ are all different.}$
\end{tabular}

\vspace{0.06cm}
\textbf{Ground Truth:} \colorbox{green!20}{\textbf{A}}
}
\end{tcolorbox}

\vspace{0.08cm}

\begin{tcolorbox}[
    width=\linewidth,
    colback=green!5,
    colframe=green!60!black,
    boxrule=1.2pt,
    arc=2mm,
    left=3pt,
    right=3pt,
    top=2pt,
    bottom=2pt,
    title={\centering\small\textbf{Qwen3-VL-32B Prediction}},
    fonttitle=\bfseries
]
{\small
\textbf{Answer:} \colorbox{green!20}{\textbf{A}} $\checkmark$ \quad
\textbf{Match F1:} 0.324 \quad \textbf{Precision:} 1.000 \quad \textbf{Recall:} 0.194

\vspace{0.04cm}
\textbf{Reasoning Steps (6 generated, 31 expected):}
\begin{enumerate}[leftmargin=10pt,itemsep=0pt,parsep=0pt,topsep=1pt]
  \item The three squares are identical in size, as stated in the problem.
  \item In square I, the shaded region consists of two right triangles. The area of each triangle is $s^2/8$. Together, the shaded area is $s^2/4$.
  \item In square II, the top-left small square is shaded. Since each small square has area $s^2/4$, the shaded area is $s^2/4$.
  \item In square III, the shaded region consists of two triangles. Each triangle has area $s^2/8$. Two such triangles give a shaded area of $s^2/4$.
  \item All three shaded areas are equal to $s^2/4$.
  \item Therefore, the shaded areas in all three squares are equal.
\end{enumerate}

\vspace{0.06cm}
\centering
\fcolorbox{orange!70}{orange!15}{%
  \parbox{0.96\linewidth}{%
    \centering\small
    \textbf{\textcolor{orange!80!black}{Cherry-Picking:}} Correct answer with only 19\% recall (model omits 25 detailed geometric steps)
  }%
}
\vspace{-2pt}
}
\end{tcolorbox}

\vspace{0.04cm}

\caption{\textbf{Correct answer with cherry-picked reasoning (Precision = 1.0, Recall = 0.194).}
The model achieves correct answer (A) with perfect precision by generating 6 high-confidence calculation steps. However, it skips 25 reference steps detailing midpoint identification, diagonal construction, and triangle congruence proofs. This exemplifies universal cherry-picking: high precision (1.0) through selective reasoning, but low recall (0.194) by omitting intermediate logical transitions. The model answers correctly through efficient shortcuts rather than comprehensive verification.}
\label{fig:supp_cherrypicking}
\end{figure}

\noindent\textbf{Cherry-picking masks reasoning failures behind correct answers.} Figure~\ref{fig:supp_cherrypicking} illustrates Finding~1 from the main paper at the individual-example level: Qwen3-VL-32B answers a geometry problem correctly with perfect precision (1.000) but covers only 6 of 31 reference steps (Recall = 0.194, F1 = 0.324), omitting midpoint identification, diagonal construction, and triangle congruence proofs. The model reaches the right answer through calculation shortcuts rather than geometric verification, exemplifying the precision-recall asymmetry observed across all 20 models (ratios 1.2$\times$ to 5.3$\times$). This pattern recurs across a nontrivial fraction of correct predictions and demonstrates that accuracy-only evaluation systematically overestimates model capabilities by rewarding confident omissions over transparent reasoning~\cite{kalai2025languagemodelshallucinate}.

\begin{figure}[t]
\centering

{\small\bfseries Input Image}\par
\vspace{1pt}
\fcolorbox{black!40}{white}{%
  \includegraphics[width=0.5\linewidth]{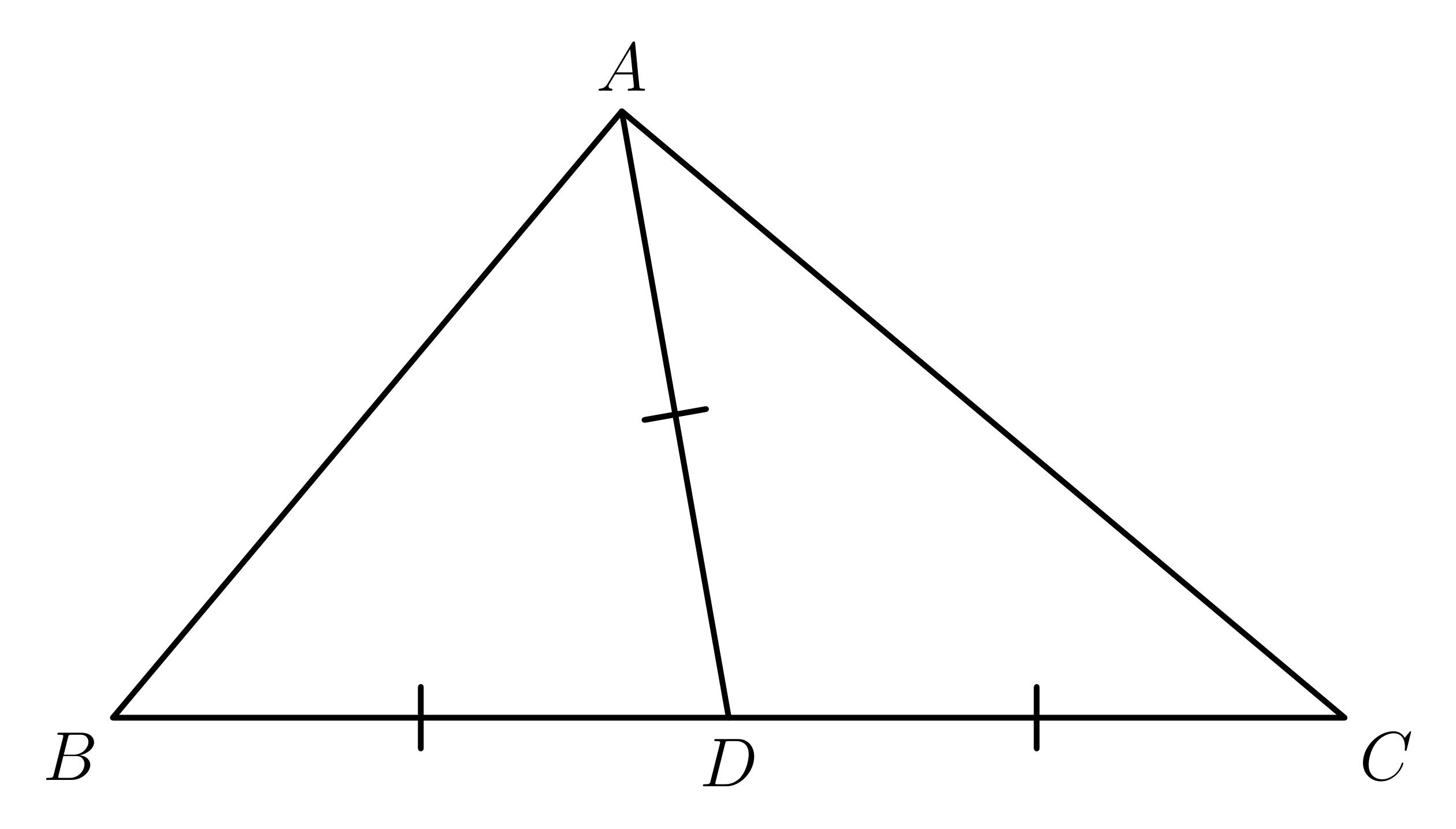}%
}

\vspace{0.08cm}

\begin{tcolorbox}[
    width=\linewidth,
    colback=blue!5,
    colframe=blue!50,
    boxrule=1pt,
    arc=2mm,
    left=3pt,
    right=3pt,
    top=2pt,
    bottom=2pt
]
{\small
\centering
\textbf{\textcolor{blue!70!black}{Question:}} In the diagram, $AD=BD=CD$ and $\angle BCA = 40^\circ$. What is the measure of $\angle BAC$?\\[2pt]
\textbf{Ground Truth:} \colorbox{green!20}{\textbf{90°}}
}
\end{tcolorbox}

\vspace{0.08cm}

\begin{tcolorbox}[
    width=\linewidth,
    colback=red!5,
    colframe=red!60,
    boxrule=1.2pt,
    arc=2mm,
    left=3pt,
    right=3pt,
    top=2pt,
    bottom=2pt,
    title={\centering\small\textbf{MiniCPM-v2.6-8B Prediction}},
    fonttitle=\bfseries
]
{\small
\textbf{Answer:} \colorbox{red!20}{\textbf{60°}} $\times$ \quad
\textbf{Match F1:} 0.214 \quad \textbf{Precision:} 1.000 \quad \textbf{Recall:} 0.120

\vspace{0.04cm}
\textbf{Reasoning Steps (3 generated, 25 expected):}
\begin{enumerate}[leftmargin=10pt,itemsep=0pt,parsep=0pt,topsep=1pt]
  \item \textcolor{red!70!black}{\textbf{Triangle ABD is equilateral since AD=BD}} \quad \textit{[FALSE assumption]}
  \item Therefore, each internal angle in triangle ABD equals 60°.
  \item Since ABC is also an equilateral triangle and all angles are congruent to those in a regular hexagon (which has equal interior angles), the measure of BAC would be 60 degrees. \quad \textit{[Fabricated reasoning]}
\end{enumerate}

\vspace{0.06cm}
\centering
\fcolorbox{red!70}{red!15}{%
  \parbox{0.96\linewidth}{%
    \centering\small
    \textbf{\textcolor{red!80!black}{Fabricated Reasoning:}} False assumption (2 equal sides $\neq$ equilateral) leads to completely incorrect logic
  }%
}
\vspace{-2pt}
}
\end{tcolorbox}

\vspace{0.04cm}

\caption{\textbf{Complete failure with fabricated reasoning (F1 = 0.214).}
MiniCPM-v2.6-8B commits a fundamental geometric error: assuming triangle ABD is equilateral when only $AD=BD$ (two sides equal does NOT make a triangle equilateral; requires three equal sides). This false premise propagates through all subsequent reasoning. Step 3 fabricates connections to ``regular hexagon'' concepts that are irrelevant to the problem. The correct solution requires recognizing that $AD=BD=CD$ implies D is the circumcenter, leading to $\angle BAC = 90°$. This represents the 14.4\% of predictions classified as ``Failure'' where both perception and logical reasoning break down catastrophically.}
\label{fig:supp_failure}
\end{figure}

\noindent\textbf{Catastrophic failures combine perceptual and logical breakdown.} Figure~\ref{fig:supp_failure} shows MiniCPM-v2.6-8B on a geometry problem, generating only 3 steps matching 3 of 25 references (Recall = 0.120, F1 = 0.214). The model assumes triangle ABD is equilateral when only $AD=BD$ holds, then fabricates connections to ``regular hexagons,'' predicting 60\textdegree{} instead of the correct 90\textdegree. Unlike the previous example, where Gemma3-12B covers the full pipeline but errs numerically (F1 = 0.857), this failure exhibits both perceptual misidentification and logical fabrication, abandoning systematic reasoning entirely. This distinction (F1: 0.214 vs.\ 0.857) is invisible to accuracy metrics, which label both as equally ``incorrect.'' Predictions falling in this failure mode require foundational capability enhancement rather than the targeted refinement that would suffice for sound-but-wrong reasoning~\cite{chen2024measuring,xu2024llavacot}.

\noindent These four examples map to distinct diagnostic regions: perfect alignment, sound reasoning with wrong answers, correct answers with cherry-picked reasoning, and catastrophic failures. Together they confirm that accuracy and reasoning quality are largely orthogonal (Finding~2 of the main paper), motivating the step-level reward approach in Section~4.5 of the main paper that explicitly incentivizes comprehensive reasoning coverage~\cite{grpo,zhang2024mmcot}.

\subsection{GRPO Training: Before and After CPR-Curriculum}

We complement the aggregate results in Section~4.5 with per-example comparisons between baseline and CPR-trained models for both Qwen2.5-VL-3B and InternVL3.5-4B. Each pair shows the same sample evaluated before and after CPR-Curriculum training, illustrating both improvements and regressions at the individual prediction level.

\begin{figure}[t]
\centering

{\small\bfseries ScienceQA (Sample 5572)}\par
\vspace{1pt}
\fcolorbox{black!40}{white}{%
  \includegraphics[width=0.55\linewidth]{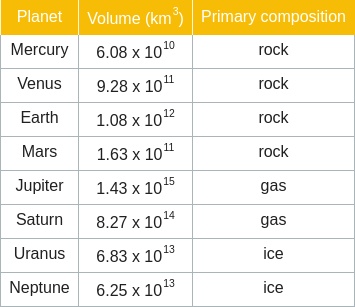}%
}

\vspace{0.08cm}

\begin{tcolorbox}[
    width=\linewidth,
    colback=blue!5,
    colframe=blue!50,
    boxrule=1pt,
    arc=2mm,
    left=3pt, right=3pt, top=2pt, bottom=2pt
]
{\small
\textbf{\textcolor{blue!70!black}{Question:}} Is the following statement about our solar system true or false? The smallest planet is made mainly of rock.\\[2pt]
\textbf{Ground Truth:} \colorbox{green!20}{\textbf{true}}
}
\end{tcolorbox}

\vspace{0.06cm}

\begin{tcolorbox}[
    width=\linewidth,
    colback=red!5,
    colframe=red!60,
    boxrule=1pt,
    arc=2mm,
    left=3pt, right=3pt, top=2pt, bottom=2pt,
    title={\centering\small\textbf{Qwen2.5-VL-3B Baseline}},
    fonttitle=\bfseries
]
{\small
\textbf{Answer:} \colorbox{red!20}{\textbf{A}} $\times$ \quad
\textbf{Match F1:} 0.333 \quad \textbf{P:} 1.000 \quad \textbf{R:} 0.200

\vspace{0.04cm}
\textbf{Reasoning Steps (2 generated, 10 reference):}
\begin{enumerate}[leftmargin=10pt,itemsep=0pt,parsep=0pt,topsep=1pt]
  \item Noted the smallest planet is Mercury.
  \item Noted Mercury is made mainly of rock.
\end{enumerate}
}
\end{tcolorbox}

\vspace{0.04cm}

\begin{tcolorbox}[
    width=\linewidth,
    colback=green!5,
    colframe=green!60!black,
    boxrule=1pt,
    arc=2mm,
    left=3pt, right=3pt, top=2pt, bottom=2pt,
    title={\centering\small\textbf{Qwen2.5-VL-3B + CPR-Curriculum}},
    fonttitle=\bfseries
]
{\small
\textbf{Answer:} \colorbox{green!20}{\textbf{true}} $\checkmark$ \quad
\textbf{Match F1:} 0.889 \quad \textbf{P:} 1.000 \quad \textbf{R:} 0.800

\vspace{0.04cm}
\textbf{Reasoning Steps (8 generated, 10 reference):}
\begin{enumerate}[leftmargin=10pt,itemsep=0pt,parsep=0pt,topsep=1pt]
  \item Observe the table.
  \item Identify the volume of each planet.
  \item Compare the volumes.
  \item Verify the composition of Mercury.
  \item Conclude that Mercury is the smallest planet made mainly of rock.
  \item Determine the volume of Mercury.
  \item Compare the volume of Mercury to other planets.
  \item Conclude that Mercury is indeed the smallest planet made mainly of rock.
\end{enumerate}
}
\end{tcolorbox}

\caption{\textbf{CPR-Curriculum improves both answer and reasoning on Qwen2.5-VL-3B.} The baseline outputs only 2 steps and answers ``A'' (wrong format). After CPR training, the model produces 8 steps covering volume comparison, composition verification, and correct conclusion (F1: 0.333$\to$0.889, Recall: 0.200$\to$0.800).}
\label{fig:grpo_qwen_pos}
\end{figure}

\noindent\textbf{Finding: CPR supervision transforms terse guesses into structured reasoning chains.} Figure~\ref{fig:grpo_qwen_pos} shows a science table question where the Qwen2.5-VL-3B baseline outputs only 2 steps and answers with the wrong format (``A'' instead of ``true''), achieving F1 = 0.333 with Recall = 0.200. After CPR-Curriculum training, the model produces 8 steps that systematically identify planetary volumes, verify Mercury's composition, and conclude correctly, reaching F1 = 0.889 and Recall = 0.800. The 4$\times$ increase in recall demonstrates that CPR's multiplicative coupling between answer correctness and step alignment (Eq.~\ref{eq:cpr_reward_supp}) directly incentivizes comprehensive reasoning rather than confident shortcuts.

\begin{figure}[t]
\centering

{\small\bfseries MathVision (Sample 2949)}\par
\vspace{1pt}
\fcolorbox{black!40}{white}{%
  \includegraphics[width=0.55\linewidth]{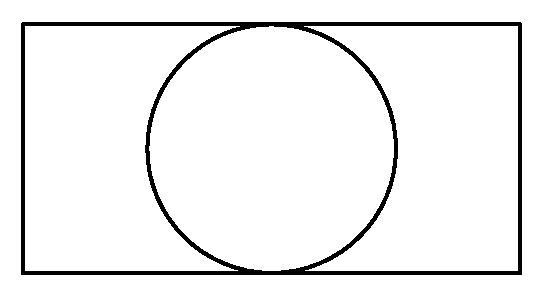}%
}

\vspace{0.08cm}

\begin{tcolorbox}[
    width=\linewidth,
    colback=blue!5,
    colframe=blue!50,
    boxrule=1pt,
    arc=2mm,
    left=3pt, right=3pt, top=2pt, bottom=2pt
]
{\small
\textbf{\textcolor{blue!70!black}{Question:}} A circle of radius 5 is inscribed in a rectangle. The ratio of the length to its width is $2:1$. What is the area of the rectangle?\\[2pt]
\textbf{Ground Truth:} \colorbox{green!20}{\textbf{200}}
}
\end{tcolorbox}

\vspace{0.06cm}

\begin{tcolorbox}[
    width=\linewidth,
    colback=green!5,
    colframe=green!60!black,
    boxrule=1pt,
    arc=2mm,
    left=3pt, right=3pt, top=2pt, bottom=2pt,
    title={\centering\small\textbf{Qwen2.5-VL-3B Baseline}},
    fonttitle=\bfseries
]
{\small
\textbf{Answer:} \colorbox{green!20}{\textbf{200}} $\checkmark$ \quad
\textbf{Match F1:} 0.870 \quad \textbf{P:} 1.000 \quad \textbf{R:} 0.769

\vspace{0.04cm}
\textbf{Reasoning Steps (10 generated, 13 reference):}
\begin{enumerate}[leftmargin=10pt,itemsep=0pt,parsep=0pt,topsep=1pt]
  \item Noted the circle's radius is 5 units.
  \item Noted the ratio of length to width is 2:1.
  \item Noted the circle is inscribed in the rectangle.
  \item Noted the circle touches all four sides of the rectangle.
  \item Noted the rectangle's width is equal to the circle's diameter.
  \item Noted the rectangle's length is twice the width.
  \item Noted the circle's diameter is 10 units.
  \item Noted the rectangle's width is 10 units.
  \item Noted the rectangle's length is 20 units.
  \item Noted the area of the rectangle is length times width.
\end{enumerate}
}
\end{tcolorbox}

\vspace{0.04cm}

\begin{tcolorbox}[
    width=\linewidth,
    colback=red!5,
    colframe=red!60,
    boxrule=1pt,
    arc=2mm,
    left=3pt, right=3pt, top=2pt, bottom=2pt,
    title={\centering\small\textbf{Qwen2.5-VL-3B + CPR-Curriculum}},
    fonttitle=\bfseries
]
{\small
\textbf{Answer:} \colorbox{red!20}{\textbf{100}} $\times$ \quad
\textbf{Match F1:} 0.632 \quad \textbf{P:} 1.000 \quad \textbf{R:} 0.462

\vspace{0.04cm}
\textbf{Reasoning Steps (6 generated, 13 reference):}
\begin{enumerate}[leftmargin=10pt,itemsep=0pt,parsep=0pt,topsep=1pt]
  \item Observe the circle.
  \item Identify the radius of the circle.
  \item Determine the width of the rectangle.
  \item Calculate the length of the rectangle.
  \item Verify the ratio of the length to the width.
  \item \textcolor{red!70!black}{\textbf{[ERROR]} Find the area of the rectangle.} \quad \textit{[Outputs 100 instead of 200]}
\end{enumerate}
}
\end{tcolorbox}

\caption{\textbf{CPR-Curriculum regression on a geometry problem.} The baseline produces 10 detailed steps with explicit calculations (diameter$\to$width$\to$length$\to$area) and answers correctly (F1: 0.870). After CPR training, the model generates only 6 generic steps without showing intermediate computations, misapplying the 2:1 ratio and arriving at 100 instead of 200 (F1: 0.632). This illustrates that CPR can encourage step generation at the expense of computational precision on multi-step math problems.}
\label{fig:grpo_qwen_neg}
\end{figure}

\noindent\textbf{Finding: CPR can trade computational precision for step coverage.} Figure~\ref{fig:grpo_qwen_neg} reveals a failure mode specific to multi-step mathematical problems. The baseline correctly solves the inscribed circle problem with 10 detailed steps showing explicit calculations (diameter = 10, width = 10, length = 20, area = 200), achieving F1 = 0.870. After CPR training, the model generates 6 steps that name the right operations (``identify radius,'' ``determine width,'' ``calculate length'') but omit the numerical computations, misapplying the 2:1 ratio and outputting 100 instead of 200 (F1: 0.632). This pattern suggests that the step-level reward successfully encourages the model to articulate its reasoning pipeline, but the word-overlap reward used during training (Section~4.5 of the main paper) cannot verify whether intermediate calculations are numerically correct, a limitation that future work could address through tool-augmented verification~\cite{gou2024tora}.

\begin{figure}[t]
\centering

{\small\bfseries MathVision (Sample 1241)}\par
\vspace{1pt}
\fcolorbox{black!40}{white}{%
  \includegraphics[width=0.3\linewidth]{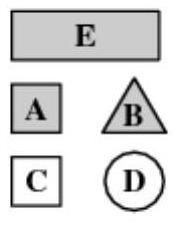}%
}

\vspace{0.08cm}

\begin{tcolorbox}[
    width=\linewidth,
    colback=blue!5,
    colframe=blue!50,
    boxrule=1pt,
    arc=2mm,
    left=3pt, right=3pt, top=2pt, bottom=2pt
]
{\small
\textbf{\textcolor{blue!70!black}{Question:}} Maria describes one of these five shapes: ``It is not a square. It is grey. It is either round or three sided.'' Which shape did she describe?\\[2pt]
\textbf{Ground Truth:} \colorbox{green!20}{\textbf{B}}
}
\end{tcolorbox}

\vspace{0.06cm}

\begin{tcolorbox}[
    width=\linewidth,
    colback=red!5,
    colframe=red!60,
    boxrule=1pt,
    arc=2mm,
    left=3pt, right=3pt, top=2pt, bottom=2pt,
    title={\centering\small\textbf{InternVL3.5-4B Baseline}},
    fonttitle=\bfseries
]
{\small
\textbf{Answer:} \colorbox{red!20}{\textbf{E}} $\times$ \quad
\textbf{Match F1:} 0.667 \quad \textbf{P:} 1.000 \quad \textbf{R:} 0.500

\vspace{0.04cm}
\textbf{Reasoning Steps (5 generated, 10 reference):}
\begin{enumerate}[leftmargin=10pt,itemsep=0pt,parsep=0pt,topsep=1pt]
  \item Identified the shapes: A (square), B (triangle), C (square), D (circle), E (rectangle).
  \item Filtered out squares: A and C are squares, so they are not the answer.
  \item Checked grey color: A, C, and E are grey.
  \item Checked round or three-sided: B is three-sided, D is round.
  \item Matched criteria: E is grey and not a square, fitting the description.
\end{enumerate}
}
\end{tcolorbox}

\vspace{0.04cm}

\begin{tcolorbox}[
    width=\linewidth,
    colback=green!5,
    colframe=green!60!black,
    boxrule=1pt,
    arc=2mm,
    left=3pt, right=3pt, top=2pt, bottom=2pt,
    title={\centering\small\textbf{InternVL3.5-4B + CPR-Curriculum}},
    fonttitle=\bfseries
]
{\small
\textbf{Answer:} \colorbox{green!20}{\textbf{B}} $\checkmark$ \quad
\textbf{Match F1:} 0.741 \quad \textbf{P:} 0.588 \quad \textbf{R:} 1.000

\vspace{0.04cm}
\textbf{Reasoning Steps (17 generated, 10 reference):}
\begin{enumerate}[leftmargin=10pt,itemsep=0pt,parsep=0pt,topsep=1pt]
  \item Identify the shape that is not a square.
  \item Identify the grey shape.
  \item Compare the options A, B, C, D, and E.
  \item A is a square, so it is not the answer.
  \item B is a triangle (three-sided), possible answer.
  \item C is a square, so it is not the answer.
  \item D is a circle (round), possible answer.
  \item[\vdots] \textit{[10 more elimination steps leading to B]}
\end{enumerate}
}
\end{tcolorbox}

\caption{\textbf{CPR-Curriculum enables systematic elimination on InternVL3.5-4B.} The baseline jumps to ``E'' without verifying all constraints (F1: 0.667). After CPR training, the model eliminates each option against all three constraints, covering all 10 reference types (R: 1.000) and arriving at the correct answer B (F1: 0.741).}
\label{fig:grpo_iv_pos}
\end{figure}

\noindent\textbf{Finding: CPR unlocks systematic elimination in InternVL3.5-4B.} Figure~\ref{fig:grpo_iv_pos} demonstrates the cross-model transfer discussed in Section~4.5 of the main paper. On a shape classification task requiring three-constraint elimination, the InternVL baseline produces 5 steps and selects ``E'' without verifying all constraints (Recall = 0.500). After CPR training, the model generates 17 steps that explicitly evaluate each option against each constraint (A: square, eliminate; B: triangle and grey, candidate; C: square, eliminate; D: round but wrong color), achieving Recall = 1.000 and the correct answer B. The step count increase from 5 to 17 mirrors the aggregate trend (3.75$\to$9.49 steps) reported in the main paper and confirms that InternVL3.5's latent reasoning capacity is substantially underutilized by the base model.

\begin{figure}[t]
\centering

{\small\bfseries MathVision (Sample 1286)}\par
\vspace{1pt}
\fcolorbox{black!40}{white}{%
  \includegraphics[width=0.65\linewidth]{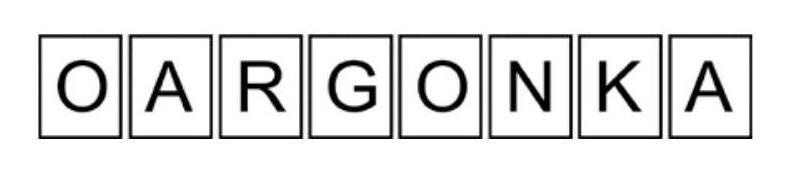}%
}

\vspace{0.08cm}

\begin{tcolorbox}[
    width=\linewidth,
    colback=blue!5,
    colframe=blue!50,
    boxrule=1pt,
    arc=2mm,
    left=3pt, right=3pt, top=2pt, bottom=2pt
]
{\small
\textbf{\textcolor{blue!70!black}{Question:}} Elisabeth sorts the following cards. With each move she is allowed to swap any two cards. What is the smallest number of moves she needs to get the word KANGAROO?\\[2pt]
\textbf{Ground Truth:} \colorbox{green!20}{\textbf{3}}
}
\end{tcolorbox}

\vspace{0.06cm}

\begin{tcolorbox}[
    width=\linewidth,
    colback=green!5,
    colframe=green!60!black,
    boxrule=1pt,
    arc=2mm,
    left=3pt, right=3pt, top=2pt, bottom=2pt,
    title={\centering\small\textbf{InternVL3.5-4B Baseline}},
    fonttitle=\bfseries
]
{\small
\textbf{Answer:} \colorbox{green!20}{\textbf{3}} $\checkmark$ \quad
\textbf{Match F1:} 0.889 \quad \textbf{P:} 0.800 \quad \textbf{R:} 1.000

\vspace{0.04cm}
\textbf{Reasoning Steps (15 generated, 12 reference):}
\begin{enumerate}[leftmargin=10pt,itemsep=0pt,parsep=0pt,topsep=1pt]
  \item Identified the current word: O A R G O N K A.
  \item Identified the target word: K A N G A R O O.
  \item Matched each letter of the target word with current positions.
  \item K is at position 7, needs to be at position 1.
  \item A is at position 2, already in place.
  \item[\vdots] \textit{[7 more position mappings]}
  \item Swap O (pos.\ 1) with K (pos.\ 7).
  \item Swap A (pos.\ 8) with O (pos.\ 5).
  \item Swap N (pos.\ 6) with R (pos.\ 3). Total: 3.
\end{enumerate}
}
\end{tcolorbox}

\vspace{0.04cm}

\begin{tcolorbox}[
    width=\linewidth,
    colback=red!5,
    colframe=red!60,
    boxrule=1pt,
    arc=2mm,
    left=3pt, right=3pt, top=2pt, bottom=2pt,
    title={\centering\small\textbf{InternVL3.5-4B + CPR-Curriculum}},
    fonttitle=\bfseries
]
{\small
\textbf{Answer:} \colorbox{red!20}{\textbf{4}} $\times$ \quad
\textbf{Match F1:} 0.857 \quad \textbf{P:} 0.750 \quad \textbf{R:} 1.000

\vspace{0.04cm}
\textbf{Reasoning Steps (16 generated, 12 reference):}
\begin{enumerate}[leftmargin=10pt,itemsep=0pt,parsep=0pt,topsep=1pt]
  \item Identify the target word `KANGAROO'.
  \item Identify the current arrangement of the cards.
  \item Identify the positions of each letter in the target word.
  \item Compare the current positions with the target positions.
  \item Identify the letters that are out of place.
  \item Count the number of misplaced letters.
  \item Identify the minimum number of swaps needed.
  \item Analyze the sequence of swaps to achieve the target word.
  \item[\vdots] \textit{[8 more steps, concluding with 4 swaps]}
\end{enumerate}
}
\end{tcolorbox}

\caption{\textbf{CPR-Curriculum regression on a combinatorial problem.} The baseline correctly identifies the optimal 3-swap solution with high F1 (0.889). After CPR training, the model covers the same reasoning types (R: 1.000) but overcounts the required swaps, answering 4 instead of 3. Both models follow the same algorithmic structure (identify$\to$compare$\to$count$\to$minimize), yet the CPR model fails at the final counting step, suggesting that step-level supervision improves reasoning coverage but does not guarantee computational accuracy on combinatorial tasks.}
\label{fig:grpo_iv_neg}
\end{figure}

\noindent\textbf{Finding: Step-level supervision does not guarantee counting accuracy.} Figure~\ref{fig:grpo_iv_neg} shows a combinatorial task where the InternVL baseline correctly identifies the 3-swap solution (F1 = 0.889). After CPR training, the model covers all 12 reference reasoning types (Recall = 1.000) and follows the same algorithmic structure (identify arrangement, compare positions, minimize swaps) but overcounts the required moves, answering 4 instead of 3 (F1 = 0.857). The regression occurs specifically at the counting step, where the model identifies the right approach but executes it imprecisely. Combined with the Qwen geometry regression (Figure~\ref{fig:grpo_qwen_neg}), this suggests a consistent limitation: CPR effectively teaches \textit{what} reasoning steps to produce but cannot enforce \textit{numerical correctness within those steps} using word-overlap rewards alone.

\section{Implementation Details for Reproducibility}
\label{sec:impl_details}

This section consolidates all fixed parameters across evaluation and training to facilitate reproducibility.

\subsection{Evaluation}

All 20 models are evaluated on the CRYSTAL test set (6,372 examples) using the HuggingFace Transformers library~\cite{wolf2020huggingface} on NVIDIA A100 GPUs with mixed-precision inference (FP16). Inference uses greedy decoding ($T=0.0$) to ensure deterministic outputs. Each model generates structured JSON containing \texttt{reasoning\_steps} and \texttt{answer} fields; malformed outputs receive placeholder steps so that all examples contribute to metrics. Match~F1 is computed with sentence encoder \texttt{all-distilroberta-v1}~\cite{reimers2019sbert}, cosine similarity threshold $\tau=0.35$, and greedy 1:1 matching (Section~3.2 of the main paper). Ordered Match~F1 uses the LIS ratio with $\alpha=0.3$ (Eq.~6 in the main paper). Accuracy uses type-adapted fuzzy matching: tolerance-based for numerics, exact for categoricals, and choice-letter normalization for multiple-choice. All metrics are macro-averaged over the 6,372 examples.

\subsection{GRPO Training}

Both Qwen2.5-VL-3B and InternVL3.5-4B train on 4$\times$NVIDIA A100 (80\,GB) with DeepSpeed ZeRO-3~\cite{rajbhandari2020zero}, mixed-precision BF16, and gradient checkpointing. Shared GRPO hyperparameters: KL coefficient $\beta=0.04$, clipping threshold $\epsilon=0.2$, sampling temperature $T=0.9$, top-$p=1.0$, top-$k=50$, random seed 42. Checkpoints are saved every 100 steps. The training dataset contains 30,312 examples from ScienceQA-IMG~\cite{lu2022scienceqa} (6,218) and TextVQA~\cite{textvqa} (24,094).

Table~\ref{tab:grpo_hyperparams} summarizes the per-model, per-phase hyperparameters. Both models share an identical Phase~2 configuration (reward weights, learning rate, candidate count, batch size), so the CPR stage is untuned across architectures; their Phase~1 warm-ups differ, as InternVL3.5-4B reached format stability with a larger candidate group, a higher learning rate, and a 5\% linear warm-up. The remaining difference is training duration. Qwen2.5-VL-3B requires 1,400 warm-up steps to reach format stability (42.56\% accuracy) before 2,800 CPR steps (4,200 total). InternVL3.5-4B stabilizes format compliance within 200 warm-up steps and shows best validation accuracy and Match~F1 at Phase~2 step 200 (400 total). We attribute the shorter horizon to InternVL3.5's larger effective batch throughput per step under dynamic resolution (up to 12 tiles at 448$\times$448 each, vs.\ Qwen's fixed resolution range of 3,136--602,112 pixels), which exposes the model to more visual tokens per gradient update. Both models are evaluated at their respective best validation checkpoint; no further training produced higher metrics for either model at the time of submission.

\begin{table}[h]
\centering
\caption{\textbf{Per-model GRPO hyperparameters.} Phase~1 trains answer-only; Phase~2 introduces CPR. All unlisted hyperparameters are shared (see text).}
\label{tab:grpo_hyperparams}
\footnotesize
\setlength{\tabcolsep}{4pt}
\begin{tabular}{l cc cc}
\toprule
& \multicolumn{2}{c}{\textbf{Qwen2.5-VL-3B}} & \multicolumn{2}{c}{\textbf{InternVL3.5-4B}} \\
\cmidrule(lr){2-3} \cmidrule(lr){4-5}
\textbf{Hyperparameter} & Phase~1 & Phase~2 & Phase~1 & Phase~2 \\
\midrule
Steps & 1,400 & 2,800 & 200 & 200 \\
Learning rate $\eta$ & $3\!\times\!10^{-6}$ & $5\!\times\!10^{-6}$ & $5\!\times\!10^{-6}$ & $5\!\times\!10^{-6}$ \\
Per-device batch size & 4 & 5 & 5 & 5 \\
Candidates $K$ & 2 & 5 & 5 & 5 \\
Gradient accumulation & 2 & 2 & 2 & 2 \\
$w_{\text{format}}$ / $w_{\text{accuracy}}$ & 2.0 / 3.0 & 2.0 / 2.0 & 2.0 / 2.0 & 2.0 / 2.0 \\
$w_{\text{reasoning}}$ & -- & 2.0 & -- & 2.0 \\
CPR weights $a_w$ / $s_w$ & -- & 0.65 / 0.35 & -- & 0.65 / 0.35 \\
\midrule
\textbf{Total steps} & \multicolumn{2}{c}{4,200} & \multicolumn{2}{c}{400} \\
\bottomrule
\end{tabular}
\end{table}

\subsection{Software and Hardware}

All experiments run on NVIDIA A100 80\,GB GPUs: 4$\times$A100 for GRPO training and 1--4$\times$A100 for inference depending on model size. The software stack comprises PyTorch 2.1+, HuggingFace Transformers 4.46+~\cite{wolf2020huggingface}, and DeepSpeed 0.15+~\cite{rajbhandari2020zero} for distributed training.

\noindent Step-level evaluation uses \texttt{all-distilroberta-v1} via \texttt{sentence-transformers} 2.2+~\cite{reimers2019sbert} with greedy 1:1 cosine-similarity matching at threshold $\tau\!=\!0.35$. Ordered F1 computes the LIS ratio via patience sorting~\cite{fredman1975lis} with $\alpha\!=\!0.3$. Gradient conflicts between accuracy and reasoning objectives are resolved with PCGrad~\cite{yu2020pcgrad}. All experiments use random seed 42 for both training and inference; the multi-agent pipeline (Section~\ref{sec:implementation_details}) uses independent seeds per generator to maximize step diversity.

\section{Additional CRYSTAL Benchmark Examples}

To further demonstrate the diversity and coverage of the CRYSTAL benchmark, we present 12 additional representative examples spanning mathematical reasoning, scientific understanding, and visual perception tasks. Figures~\ref{fig:crystal_ex1} through~\ref{fig:crystal_ex12} showcase samples from MathVision (mathematical problem-solving with geometric and numerical reasoning), ScienceQA (scientific knowledge and multi-hop inference), RealWorldQA (spatial understanding in real-world contexts), MMVP (fine-grained visual perception), and PlotQA (chart interpretation). Each figure displays a single example with complete question context (including multiple-choice options where applicable), ground truth answer, and reference reasoning steps demonstrating the expected solution path.

\begin{figure}[t]
\centering

{{\small\bfseries Mathvision (Sample 1265)}}\par
\vspace{1pt}
\fcolorbox{black!40}{white}{%
  \includegraphics[width=0.6\linewidth,height=0.35\textheight,keepaspectratio]{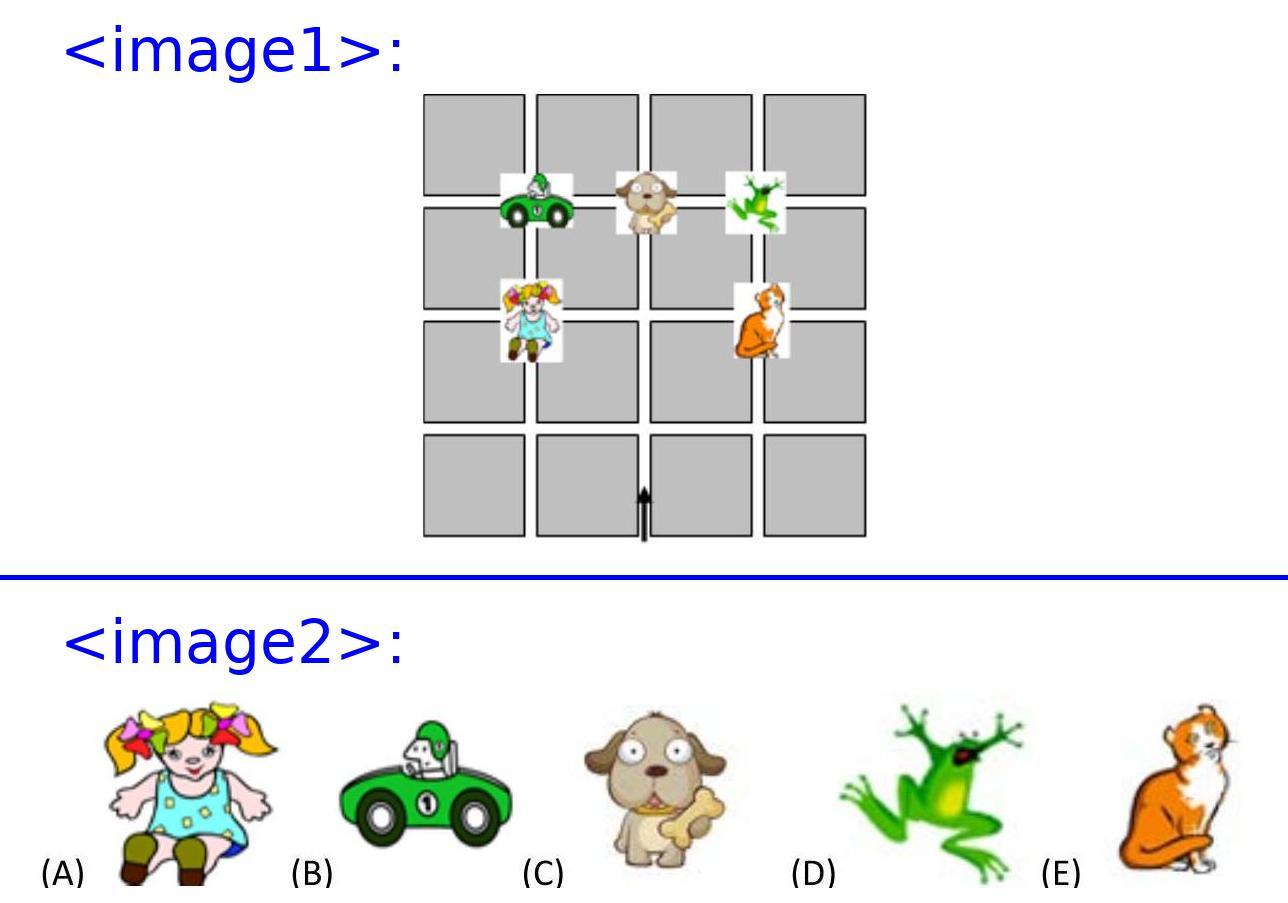}%
}

\vspace{0.08cm}

\begin{tcolorbox}[
    width=\linewidth,
    colback=blue!5,
    colframe=blue!50,
    boxrule=1pt,
    arc=2mm,
    left=3pt,
    right=3pt,
    top=2pt,
    bottom=2pt
]
{\small
\textbf{\textcolor{blue!70!black}{Question:}} Anna starts in the direction of the arrow. At each crossing she turns either right or left. At the first crossing she turns right, at the next left, then left again, then right, then left and left again. What will she find at the next crossing that she comes to?   \\ A) A \\ B) B \\ C) C \\ D) D \\ E) E

\vspace{2pt}
\textbf{Ground Truth:} \colorbox{green!20}{\textbf{A}}
}
\end{tcolorbox}

\vspace{0.08cm}

\begin{tcolorbox}[
    width=\linewidth,
    colback=yellow!8,
    colframe=orange!60,
    boxrule=1pt,
    arc=2mm,
    left=3pt,
    right=3pt,
    top=2pt,
    bottom=2pt,
    title={\centering\small\textbf{Reference Reasoning Steps (12 steps)}},
    fonttitle=\bfseries
]
{\small
\begin{enumerate}[leftmargin=10pt,itemsep=0pt,parsep=0pt,topsep=1pt]
  \item Text says Anna starts in the direction of an arrow.
  \item She turns at each crossing either right or left.
  \item First crossing instruction in text: 'turns right'.
  \item Second crossing instruction in text: 'left'.
  \item Third crossing instruction in text: 'left again'.
  \item Fourth crossing instruction in text: 'right'.
  \item Fifth crossing instruction in text: 'left'.
  \item Sixth crossing instruction in text: 'left again'.
  \item Question asks what she will find at the next crossing reached after this sequence.
  \item Answer format is multiple-choice with options 'A, B, C, D, E'.
\end{enumerate}
\textit{[2 more steps omitted for brevity]}
}
\end{tcolorbox}

\caption{\textbf{Mathvision example from CRYSTAL benchmark (Sample 1265).} Shows input image, question with multiple-choice options (if applicable), ground truth answer, and reference reasoning steps demonstrating the expected step-by-step solution path.}
\label{fig:crystal_ex1}
\end{figure}

\begin{figure}[t]
\centering

{{\small\bfseries Mathvision (Sample 899)}}\par
\vspace{1pt}
\fcolorbox{black!40}{white}{%
  \includegraphics[width=0.6\linewidth,height=0.35\textheight,keepaspectratio]{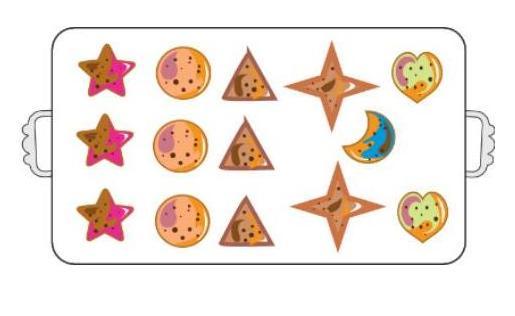}%
}

\vspace{0.08cm}

\begin{tcolorbox}[
    width=\linewidth,
    colback=blue!5,
    colframe=blue!50,
    boxrule=1pt,
    arc=2mm,
    left=3pt,
    right=3pt,
    top=2pt,
    bottom=2pt
]
{\small
\textbf{\textcolor{blue!70!black}{Question:}} Each participant in a cooking contest baked one tray of cookies like the one shown beside. What is the smallest number of trays of cookies needed to make the following plate? \\ 
\vspace{2pt}
\textbf{Ground Truth:} \colorbox{green!20}{\textbf{3}}
}
\end{tcolorbox}

\vspace{0.08cm}

\begin{tcolorbox}[
    width=\linewidth,
    colback=yellow!8,
    colframe=orange!60,
    boxrule=1pt,
    arc=2mm,
    left=3pt,
    right=3pt,
    top=2pt,
    bottom=2pt,
    title={\centering\small\textbf{Reference Reasoning Steps (12 steps)}},
    fonttitle=\bfseries
]
{\small
\begin{enumerate}[leftmargin=10pt,itemsep=0pt,parsep=0pt,topsep=1pt]
  \item The prompt asks for the minimum number of trays needed to assemble a shown plate from identical trays.
  \item In image, six distinct cookie designs are displayed as the contents of a standard tray.
  \item Image shows two orientations of the same set of designs on cubes, indicating rotations don’t change the design type.
  \item Each tray therefore contributes at most one cookie of each design type.
  \item The plate in image features multiple cookies, including repeated occurrences of one design (diamond-in-square motif).
  \item The repeated design on the plate appears exactly three times.
  \item Since one tray contains only one of that design, at least three trays are necessary to get three copies.
  \item No other design on the plate requires more than one copy, so the lower bound from the repeated design is tight.
  \item Taking the three repeated-design cookies from three trays and the remaining needed designs from these trays is feasible.
  \item Thus the minimal trays required equals the maximum multiplicity of any single design on the plate.
\end{enumerate}
\textit{[2 more steps omitted for brevity]}
}
\end{tcolorbox}

\caption{\textbf{Mathvision example from CRYSTAL benchmark (Sample 899).} Shows input image, question with multiple-choice options (if applicable), ground truth answer, and reference reasoning steps demonstrating the expected step-by-step solution path.}
\label{fig:crystal_ex2}
\end{figure}

\begin{figure}[t]
\centering

{{\small\bfseries Mathvision (Sample 1947)}}\par
\vspace{1pt}
\fcolorbox{black!40}{white}{%
  \includegraphics[width=0.6\linewidth,height=0.35\textheight,keepaspectratio]{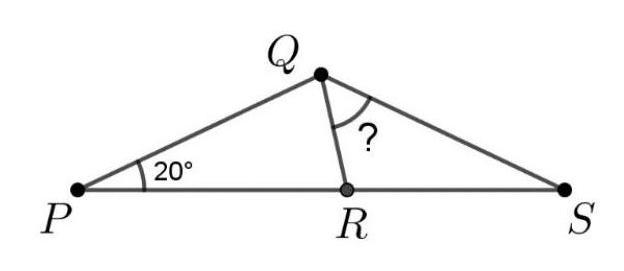}%
}

\vspace{0.08cm}

\begin{tcolorbox}[
    width=\linewidth,
    colback=blue!5,
    colframe=blue!50,
    boxrule=1pt,
    arc=2mm,
    left=3pt,
    right=3pt,
    top=2pt,
    bottom=2pt
]
{\small
\textbf{\textcolor{blue!70!black}{Question:}} The following information is known about triangle PSQ: \$\textbackslash{}angle Q P S=20\^{}{\textbackslash{}circ}\$. The triangle PSQ has been split up into two smaller triangles by the line \$Q R\$ as shown. It is known that \$P Q=P R=Q S\$. How big is the angle RQS?   \\ A) \$50\^{}{\textbackslash{}circ}\$ \\ B) \$60\^{}{\textbackslash{}circ}\$ \\ C) \$65\^{}{\textbackslash{}circ}\$ \\ D) \$70\^{}{\textbackslash{}circ}\$ \\ E) \$75\^{}{\textbackslash{}circ}\$

\vspace{2pt}
\textbf{Ground Truth:} \colorbox{green!20}{\textbf{B}}
}
\end{tcolorbox}

\vspace{0.08cm}

\begin{tcolorbox}[
    width=\linewidth,
    colback=yellow!8,
    colframe=orange!60,
    boxrule=1pt,
    arc=2mm,
    left=3pt,
    right=3pt,
    top=2pt,
    bottom=2pt,
    title={\centering\small\textbf{Reference Reasoning Steps (13 steps)}},
    fonttitle=\bfseries
]
{\small
\begin{enumerate}[leftmargin=10pt,itemsep=0pt,parsep=0pt,topsep=1pt]
  \item Text states ∠QPS = 20°.
  \item Triangle PSQ is split by line QR into triangles PQR and RQS (R on PS).
  \item Given PQ = QS, triangle PQS is isosceles with base PS.
  \item So base angles at P and S are equal: ∠QPS = ∠QSP = 20°.
  \item Triangle sum gives ∠PQS = 180° − 20° − 20° = 140°.
  \item Given PQ = PR, triangle PQR is isosceles with base QR.
  \item ∠QPR equals ∠QPS since R lies on PS, so ∠QPR = 20°.
  \item Thus base angles ∠PQR and ∠PRQ are equal and each = (180° − 20°)/2 = 80°.
  \item Angle at Q of the big triangle is split: ∠PQS = ∠PQR + ∠RQS.
  \item Substitute ∠PQS = 140° and ∠PQR = 80°.
\end{enumerate}
\textit{[3 more steps omitted for brevity]}
}
\end{tcolorbox}

\caption{\textbf{Mathvision example from CRYSTAL benchmark (Sample 1947).} Shows input image, question with multiple-choice options (if applicable), ground truth answer, and reference reasoning steps demonstrating the expected step-by-step solution path.}
\label{fig:crystal_ex3}
\end{figure}

\begin{figure}[t]
\centering

{{\small\bfseries Plotqa (Sample 6193)}}\par
\vspace{1pt}
\fcolorbox{black!40}{white}{%
  \includegraphics[width=0.6\linewidth,height=0.35\textheight,keepaspectratio]{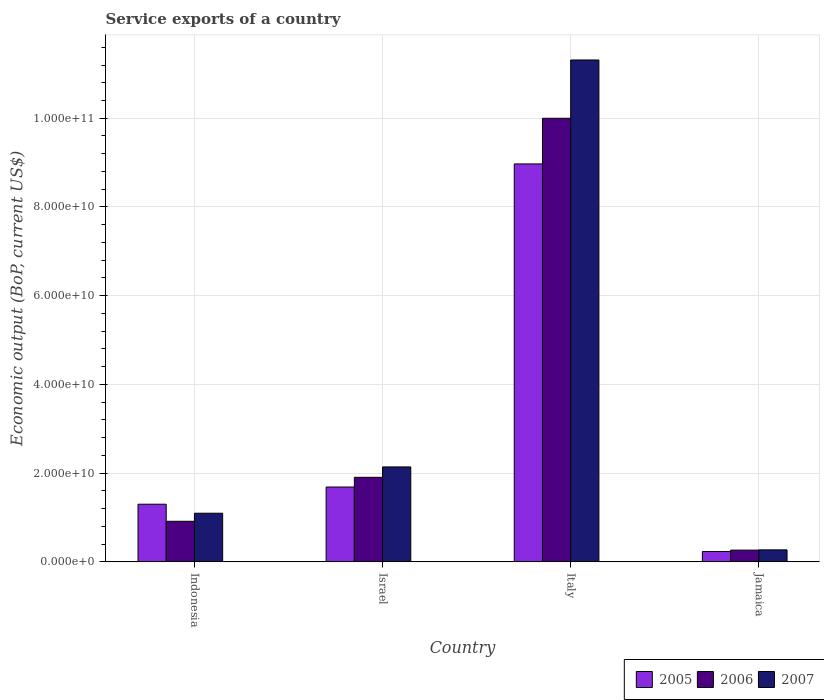}%
}

\vspace{0.08cm}

\begin{tcolorbox}[
    width=\linewidth,
    colback=blue!5,
    colframe=blue!50,
    boxrule=1pt,
    arc=2mm,
    left=3pt,
    right=3pt,
    top=2pt,
    bottom=2pt
]
{\small
\textbf{\textcolor{blue!70!black}{Question:}} How many bars are there on the 4th tick from the left ?

\vspace{2pt}
\textbf{Ground Truth:} \colorbox{green!20}{\textbf{3}}
}
\end{tcolorbox}

\vspace{0.08cm}

\begin{tcolorbox}[
    width=\linewidth,
    colback=yellow!8,
    colframe=orange!60,
    boxrule=1pt,
    arc=2mm,
    left=3pt,
    right=3pt,
    top=2pt,
    bottom=2pt,
    title={\centering\small\textbf{Reference Reasoning Steps (8 steps)}},
    fonttitle=\bfseries
]
{\small
\begin{enumerate}[leftmargin=10pt,itemsep=0pt,parsep=0pt,topsep=1pt]
  \item The question asks for the number of bars located at the fourth tick from the left on the x‑axis of the bar chart.
  \item The x‑axis major labels are listed in order: Indonesia, Israel, Italy, Jamaica, then the same four countries repeat for the next group.
  \item The fourth tick from the left therefore corresponds to the first occurrence of the label 'Jamaica'.
  \item The chart contains three data series (2005, 2006, 2007), each represented by a set of bars with distinct colors.
  \item For each series, a bar is drawn at the x‑position that aligns with each country label; thus at the 'Jamaica' tick there is one bar from 2005, one from 2006, and one from 2007.
  \item Counting these bars gives a total of three bars at the fourth tick.
  \item Validate that the counted total (3) matches the expected answer value.
  \item Return the numeric value.
\end{enumerate}
}
\end{tcolorbox}

\caption{\textbf{Plotqa example from CRYSTAL benchmark (Sample 6193).} Shows input image, question with multiple-choice options (if applicable), ground truth answer, and reference reasoning steps demonstrating the expected step-by-step solution path.}
\label{fig:crystal_ex4}
\end{figure}

\begin{figure}[t]
\centering

{{\small\bfseries Scienceqa (Sample 4306)}}\par
\vspace{1pt}
\fcolorbox{black!40}{white}{%
  \includegraphics[width=0.6\linewidth,height=0.35\textheight,keepaspectratio]{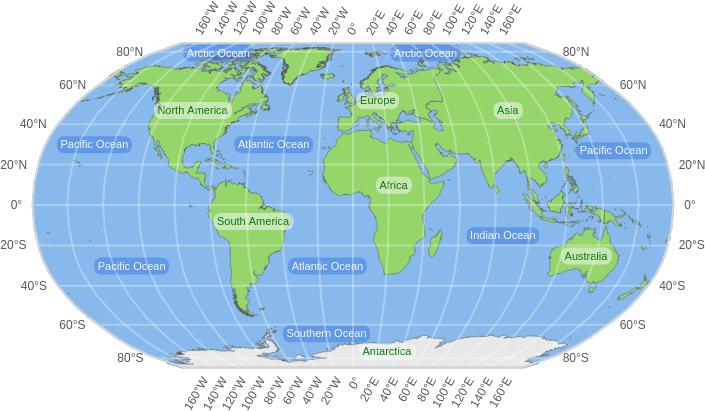}%
}

\vspace{0.08cm}

\begin{tcolorbox}[
    width=\linewidth,
    colback=blue!5,
    colframe=blue!50,
    boxrule=1pt,
    arc=2mm,
    left=3pt,
    right=3pt,
    top=2pt,
    bottom=2pt
]
{\small
\textbf{\textcolor{blue!70!black}{Question:}} Which of these continents does the equator intersect?   \\ A) North America \\ B) Africa \\ C) Europe

\vspace{2pt}
\textbf{Ground Truth:} \colorbox{green!20}{\textbf{Africa}}
}
\end{tcolorbox}

\vspace{0.08cm}

\begin{tcolorbox}[
    width=\linewidth,
    colback=yellow!8,
    colframe=orange!60,
    boxrule=1pt,
    arc=2mm,
    left=3pt,
    right=3pt,
    top=2pt,
    bottom=2pt,
    title={\centering\small\textbf{Reference Reasoning Steps (8 steps)}},
    fonttitle=\bfseries
]
{\small
\begin{enumerate}[leftmargin=10pt,itemsep=0pt,parsep=0pt,topsep=1pt]
  \item The equator is the line located at 0° latitude and shows how far north or south a place is.
  \item Lines of latitude, including the equator, go all the way around the Earth.
  \item The equator bisects the Earth into the Northern and Southern Hemispheres.
  \item Africa spans both the northern and southern hemispheres.
  \item Europe is located entirely north of the equator.
  \item North America is also located entirely north of the equator.
  \item Africa is crossed by the equator, making it one of the continents intersected by the equator.
  \item Select the continent intersected by the equator among the options.
\end{enumerate}
}
\end{tcolorbox}

\caption{\textbf{Scienceqa example from CRYSTAL benchmark (Sample 4306).} Shows input image, question with multiple-choice options (if applicable), ground truth answer, and reference reasoning steps demonstrating the expected step-by-step solution path.}
\label{fig:crystal_ex5}
\end{figure}

\begin{figure}[t]
\centering

{{\small\bfseries Scienceqa (Sample 4116)}}\par
\vspace{1pt}
\fcolorbox{black!40}{white}{%
  \includegraphics[width=0.6\linewidth,height=0.35\textheight,keepaspectratio]{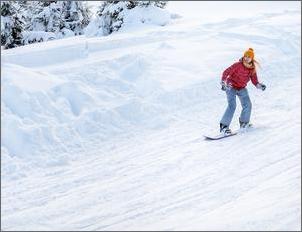}%
}

\vspace{0.08cm}

\begin{tcolorbox}[
    width=\linewidth,
    colback=blue!5,
    colframe=blue!50,
    boxrule=1pt,
    arc=2mm,
    left=3pt,
    right=3pt,
    top=2pt,
    bottom=2pt
]
{\small
\textbf{\textcolor{blue!70!black}{Question:}} Identify the question that Heather and Tanner’s experiment can best answer. \\  \\ A) Does Heather’s snowboard slide down a hill in less time when it has a layer of wax or when it does not have a layer of wax? \\ B) Does Heather’s snowboard slide down a hill in less time when it has a thin layer of wax or a thick layer of wax?

\vspace{2pt}
\textbf{Ground Truth:} \colorbox{green!20}{\textbf{B) Does Heather’s snowboard slide down down a hill...}}
}
\end{tcolorbox}

\vspace{0.08cm}

\begin{tcolorbox}[
    width=\linewidth,
    colback=yellow!8,
    colframe=orange!60,
    boxrule=1pt,
    arc=2mm,
    left=3pt,
    right=3pt,
    top=2pt,
    bottom=2pt,
    title={\centering\small\textbf{Reference Reasoning Steps (11 steps)}},
    fonttitle=\bfseries
]
{\small
\begin{enumerate}[leftmargin=10pt,itemsep=0pt,parsep=0pt,topsep=1pt]
  \item Heather alternated between riding with and without wax on her snowboard.
  \item The experiment compared average times of rides with wax and without wax.
  \item The measured outcome was the time it took to slide down the hill.
  \item No mention of varying wax thickness in the experiment.
  \item The tested variable was the presence or absence of wax on the snowboard.
  \item The experiment could not compare different wax thickness levels.
  \item The purpose was to measure effect of wax on slide time.
  \item The experiment answered about wax presence affecting slide time.
  \item Option B refers to different wax thicknesses, not tested here.
  \item Option A addresses presence or absence of wax, which was tested.
\end{enumerate}
\textit{[1 more steps omitted for brevity]}
}
\end{tcolorbox}

\caption{\textbf{Scienceqa example from CRYSTAL benchmark (Sample 4116).} Shows input image, question with multiple-choice options (if applicable), ground truth answer, and reference reasoning steps demonstrating the expected step-by-step solution path.}
\label{fig:crystal_ex6}
\end{figure}

\begin{figure}[t]
\centering

{{\small\bfseries Scienceqa (Sample 5429)}}\par
\vspace{1pt}
\fcolorbox{black!40}{white}{%
  \includegraphics[width=0.6\linewidth,height=0.35\textheight,keepaspectratio]{images/crystal_ex7_sample5429.jpg}%
}

\vspace{0.08cm}

\begin{tcolorbox}[
    width=\linewidth,
    colback=blue!5,
    colframe=blue!50,
    boxrule=1pt,
    arc=2mm,
    left=3pt,
    right=3pt,
    top=2pt,
    bottom=2pt
]
{\small
\textbf{\textcolor{blue!70!black}{Question:}} What can Joey and Darell trade to each get what they want? \\  \\ A) Joey can trade his tomatoes for Darell's sandwich. \\ B) Darell can trade his broccoli for Joey's oranges. \\ C) Darell can trade his almonds for Joey's tomatoes. \\ D) Joey can trade his tomatoes for Darell's broccoli.

\vspace{2pt}
\textbf{Ground Truth:} \colorbox{green!20}{\textbf{D) Joey can trade his tomatoes for Darell's broccoli.}}
}
\end{tcolorbox}

\vspace{0.08cm}

\begin{tcolorbox}[
    width=\linewidth,
    colback=yellow!8,
    colframe=orange!60,
    boxrule=1pt,
    arc=2mm,
    left=3pt,
    right=3pt,
    top=2pt,
    bottom=2pt,
    title={\centering\small\textbf{Reference Reasoning Steps (11 steps)}},
    fonttitle=\bfseries
]
{\small
\begin{enumerate}[leftmargin=10pt,itemsep=0pt,parsep=0pt,topsep=1pt]
  \item Joey wants broccoli in his lunch
  \item Darell wants tomatoes in his lunch
  \item Joey has tomatoes in his lunch box
  \item Darell has broccoli in his lunch box
  \item Trading goods directly to satisfy both preferences describes barter
  \item Option A involves Joey trading tomatoes for Darell's sandwich, which isn't what either wants
  \item Option B involves Darell trading broccoli for Joey's oranges, which isn't what either wants
  \item Option C involves Darell trading almonds for Joey's tomatoes, which isn't what either wants
  \item Option D involves Joey trading tomatoes for Darell's broccoli, satisfying both preferences
  \item Only option D directly results in both Joey and Darell getting what they want
\end{enumerate}
\textit{[1 more steps omitted for brevity]}
}
\end{tcolorbox}

\caption{\textbf{Scienceqa example from CRYSTAL benchmark (Sample 5429).} Shows input image, question with multiple-choice options (if applicable), ground truth answer, and reference reasoning steps demonstrating the expected step-by-step solution path.}
\label{fig:crystal_ex7}
\end{figure}

\begin{figure}[t]
\centering

{{\small\bfseries Scienceqa (Sample 4031)}}\par
\vspace{1pt}
\fcolorbox{black!40}{white}{%
  \includegraphics[width=0.6\linewidth,height=0.35\textheight,keepaspectratio]{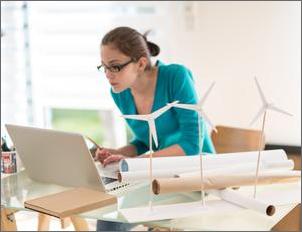}%
}

\vspace{0.08cm}

\begin{tcolorbox}[
    width=\linewidth,
    colback=blue!5,
    colframe=blue!50,
    boxrule=1pt,
    arc=2mm,
    left=3pt,
    right=3pt,
    top=2pt,
    bottom=2pt
]
{\small
\textbf{\textcolor{blue!70!black}{Question:}} Which of the following could Rose's test show? \\  \\ A) if the new turbine could turn easily \\ B) whether the new turbine could produce 10\% more electricity \\ C) how much the new turbine would weigh

\vspace{2pt}
\textbf{Ground Truth:} \colorbox{green!20}{\textbf{B) Whether the new turbine could produce...}}
}
\end{tcolorbox}

\vspace{0.08cm}

\begin{tcolorbox}[
    width=\linewidth,
    colback=yellow!8,
    colframe=orange!60,
    boxrule=1pt,
    arc=2mm,
    left=3pt,
    right=3pt,
    top=2pt,
    bottom=2pt,
    title={\centering\small\textbf{Reference Reasoning Steps (13 steps)}},
    fonttitle=\bfseries
]
{\small
\begin{enumerate}[leftmargin=10pt,itemsep=0pt,parsep=0pt,topsep=1pt]
  \item Rose created a computer model to test her design for a new turbine.
  \item The goal was to produce 10\% more electricity than older turbines.
  \item Rose calculated how much more electricity the new turbine could produce compared to older ones.
  \item The test focused on measuring the electrical output, not ease of turning or weight.
  \item Tests measure criteria like increased electricity production, which was the primary objective.
  \item The description specifies that the test showed the potential increase in electricity production.
  \item Identify the test variable: electricity production capability.
  \item Identify the measurement: increase in electricity production.
  \item The passage emphasizes calculating the electricity production gain, not mechanical ease or weight.
  \item Rose's model was used to compare new and old turbine performance specifically for electricity output.
\end{enumerate}
\textit{[3 more steps omitted for brevity]}
}
\end{tcolorbox}

\caption{\textbf{Scienceqa example from CRYSTAL benchmark (Sample 4031).} Shows input image, question with multiple-choice options (if applicable), ground truth answer, and reference reasoning steps demonstrating the expected step-by-step solution path.}
\label{fig:crystal_ex8}
\end{figure}

\begin{figure}[t]
\centering

{{\small\bfseries Realwordqa (Sample 714)}}\par
\vspace{1pt}
\fcolorbox{black!40}{white}{%
  \includegraphics[width=0.6\linewidth,height=0.35\textheight,keepaspectratio]{images/crystal_ex9_sample714.jpg}%
}

\vspace{0.08cm}

\begin{tcolorbox}[
    width=\linewidth,
    colback=blue!5,
    colframe=blue!50,
    boxrule=1pt,
    arc=2mm,
    left=3pt,
    right=3pt,
    top=2pt,
    bottom=2pt
]
{\small
\textbf{\textcolor{blue!70!black}{Question:}} Is there a stop sign facing us? \\ Please answer directly with a single word or number.

\vspace{2pt}
\textbf{Ground Truth:} \colorbox{green!20}{\textbf{Yes}}
}
\end{tcolorbox}

\vspace{0.08cm}

\begin{tcolorbox}[
    width=\linewidth,
    colback=yellow!8,
    colframe=orange!60,
    boxrule=1pt,
    arc=2mm,
    left=3pt,
    right=3pt,
    top=2pt,
    bottom=2pt,
    title={\centering\small\textbf{Reference Reasoning Steps (9 steps)}},
    fonttitle=\bfseries
]
{\small
\begin{enumerate}[leftmargin=10pt,itemsep=0pt,parsep=0pt,topsep=1pt]
  \item Scene: nighttime residential street with curbs and sidewalks visible on the right.
  \item Locate a red octagonal sign mounted on a pole near the right curb at the intersection ahead.
  \item The sign has a white border and white letter shapes consistent with the word ``STOP''.
  \item The sign face is oriented toward the camera (frontal view of the octagon), not edge-on.
  \item The sign is positioned at the near-right corner of the intersection in our lane of approach.
  \item Lighting and reflections allow the sign face to be visually inspected from the camera viewpoint.
  \item No other stop sign faces are clearly visible that would be mistaken for the one near-right corner.
  \item From the visible red octagonal face with white letters oriented toward the camera, conclude a stop sign is facing us.
  \item Answer: ``Yes''
\end{enumerate}
}
\end{tcolorbox}

\caption{\textbf{Realwordqa example from CRYSTAL benchmark (Sample 714).} Shows input image, question with multiple-choice options (if applicable), ground truth answer, and reference reasoning steps demonstrating the expected step-by-step solution path.}
\label{fig:crystal_ex9}
\end{figure}

\begin{figure}[t]
\centering

{{\small\bfseries Realwordqa (Sample 580)}}\par
\vspace{1pt}
\fcolorbox{black!40}{white}{%
  \includegraphics[width=0.6\linewidth,height=0.35\textheight,keepaspectratio]{images/crystal_ex10_sample580.jpg}%
}

\vspace{0.08cm}

\begin{tcolorbox}[
    width=\linewidth,
    colback=blue!5,
    colframe=blue!50,
    boxrule=1pt,
    arc=2mm,
    left=3pt,
    right=3pt,
    top=2pt,
    bottom=2pt
]
{\small
\textbf{\textcolor{blue!70!black}{Question:}} How many boats are in this scene? \\  \\ A. 1 \\ B. 4 \\ C. 0 \\ Please answer directly with only the letter of the correct option and nothing else.

\vspace{2pt}
\textbf{Ground Truth:} \colorbox{green!20}{\textbf{A) 1}}
}
\end{tcolorbox}

\vspace{0.08cm}

\begin{tcolorbox}[
    width=\linewidth,
    colback=yellow!8,
    colframe=orange!60,
    boxrule=1pt,
    arc=2mm,
    left=3pt,
    right=3pt,
    top=2pt,
    bottom=2pt,
    title={\centering\small\textbf{Reference Reasoning Steps (9 steps)}},
    fonttitle=\bfseries
]
{\small
\begin{enumerate}[leftmargin=10pt,itemsep=0pt,parsep=0pt,topsep=1pt]
  \item Observe urban residential street scene with houses and parked vehicles on both sides
  \item Note no visible water body (lake, river, sea) in the image
  \item Scan left curb: see parked cars, no boat-shaped objects
  \item Scan center roadway: see only cars, no boats
  \item Scan right curb in the midground: identify a hull-shaped object on a small trailer or rack near a parked vehicle
  \item Confirm the object has a pointed bow-like front and curved hull consistent with a small boat
  \item Confirm no other boat-shaped objects along either curb or in the distance
  \item Therefore the total visible boats = 1
  \item Select option (A) based on the verified count
\end{enumerate}
}
\end{tcolorbox}

\caption{\textbf{Realwordqa example from CRYSTAL benchmark (Sample 580).} Shows input image, question with multiple-choice options (if applicable), ground truth answer, and reference reasoning steps demonstrating the expected step-by-step solution path.}
\label{fig:crystal_ex10}
\end{figure}

\begin{figure}[t]
\centering

{{\small\bfseries MMVP (Sample 5866)}}\par
\vspace{1pt}
\fcolorbox{black!40}{white}{%
  \includegraphics[width=0.6\linewidth,height=0.35\textheight,keepaspectratio]{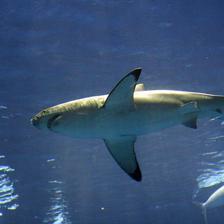}%
}

\vspace{0.08cm}

\begin{tcolorbox}[
    width=\linewidth,
    colback=blue!5,
    colframe=blue!50,
    boxrule=1pt,
    arc=2mm,
    left=3pt,
    right=3pt,
    top=2pt,
    bottom=2pt
]
{\small
\textbf{\textcolor{blue!70!black}{Question:}} Is the shark's belly visible in this image? \\  \\ A) Yes \\ B) No

\vspace{2pt}
\textbf{Ground Truth:} \colorbox{green!20}{\textbf{A) Yes}}
}
\end{tcolorbox}

\vspace{0.08cm}

\begin{tcolorbox}[
    width=\linewidth,
    colback=yellow!8,
    colframe=orange!60,
    boxrule=1pt,
    arc=2mm,
    left=3pt,
    right=3pt,
    top=2pt,
    bottom=2pt,
    title={\centering\small\textbf{Reference Reasoning Steps (9 steps)}},
    fonttitle=\bfseries
]
{\small
\begin{enumerate}[leftmargin=10pt,itemsep=0pt,parsep=0pt,topsep=1pt]
  \item A single shark is centered against blue water.
  \item The shark is viewed from slightly below, exposing its underside.
  \item The ventral surface appears pale compared to the darker top.
  \item The lower body from snout to tail is unobstructed.
  \item Undersides of the pectoral fins are visible.
  \item Lighting highlights the pale belly distinctly from the dorsal side.
  \item No objects or shadows hide the underside.
  \item These visible cues indicate the belly is in view.
  \item Select option A
\end{enumerate}
}
\end{tcolorbox}

\caption{\textbf{MMVP example from CRYSTAL benchmark (Sample 5866).} Shows input image, question with multiple-choice options (if applicable), ground truth answer, and reference reasoning steps demonstrating the expected step-by-step solution path.}
\label{fig:crystal_ex11}
\end{figure}

\begin{figure}[t]
\centering

{{\small\bfseries MMVP (Sample 6038)}}\par
\vspace{1pt}
\fcolorbox{black!40}{white}{%
  \includegraphics[width=0.6\linewidth,height=0.35\textheight,keepaspectratio]{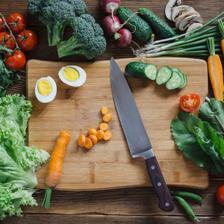}%
}

\vspace{0.08cm}

\begin{tcolorbox}[
    width=\linewidth,
    colback=blue!5,
    colframe=blue!50,
    boxrule=1pt,
    arc=2mm,
    left=3pt,
    right=3pt,
    top=2pt,
    bottom=2pt
]
{\small
\textbf{\textcolor{blue!70!black}{Question:}} Is the knife laid flat or standing up? \\  \\ A) Flat \\ B) Standing up

\vspace{2pt}
\textbf{Ground Truth:} \colorbox{green!20}{\textbf{A) Flat}}
}
\end{tcolorbox}

\vspace{0.08cm}

\begin{tcolorbox}[
    width=\linewidth,
    colback=yellow!8,
    colframe=orange!60,
    boxrule=1pt,
    arc=2mm,
    left=3pt,
    right=3pt,
    top=2pt,
    bottom=2pt,
    title={\centering\small\textbf{Reference Reasoning Steps (10 steps)}},
    fonttitle=\bfseries
]
{\small
\begin{enumerate}[leftmargin=10pt,itemsep=0pt,parsep=0pt,topsep=1pt]
  \item Knife lies on wooden cutting board near center
  \item Broad blade face visible from top-down perspective
  \item Handle rests flush against the board surface
  \item Both edge and spine visible, not an edge-on view
  \item Tip oriented toward upper area while touching the board
  \item Shadow falls closely beneath blade on the board
  \item No stand or block holding knife upright
  \item Surrounding sliced vegetables sit on the same flat board plane
  \item No visible gap under blade or handle suggesting lift
  \item Select option A
\end{enumerate}
}
\end{tcolorbox}

\caption{\textbf{MMVP example from CRYSTAL benchmark (Sample 6038).} Shows input image, question with multiple-choice options (if applicable), ground truth answer, and reference reasoning steps demonstrating the expected step-by-step solution path.}
\label{fig:crystal_ex12}
\end{figure}

\section{GRPO Training Dataset Examples}

To provide transparency into the training data quality used for reinforcement learning experiments (Section~4.5 of the main paper), we present 10 representative examples from the GRPO training dataset. Figures~\ref{fig:grpo_train_ex1} through~\ref{fig:grpo_train_ex10} showcase samples from ScienceQA-IMG (5 examples) and TextVQA (5 examples), illustrating the diversity of question types and reasoning complexity. Each training sample contains expert-generated reference reasoning steps produced by our multi-agent annotation pipeline (Section~3.1), enabling step-level reward computation without manual annotation during deployment. These examples demonstrate that the training data covers diverse reasoning scenarios including material identification (Fig.~\ref{fig:grpo_train_ex1}), biological adaptation (Fig.~\ref{fig:grpo_train_ex2}), taxonomy (Fig.~\ref{fig:grpo_train_ex3}), geography (Fig.~\ref{fig:grpo_train_ex5}), and text recognition in natural images (Figs.~\ref{fig:grpo_train_ex6}--\ref{fig:grpo_train_ex10}). The step annotations provide detailed reasoning chains (7--15 steps per example) that guide the model toward transparent, verifiable reasoning patterns during reinforcement learning.

\begin{figure}[t]
\centering

{{\small\bfseries Scienceqa Training Sample (ID 29750)}}\par
\vspace{1pt}
\fcolorbox{black!40}{white}{%
  \includegraphics[width=0.6\linewidth,height=0.35\textheight,keepaspectratio]{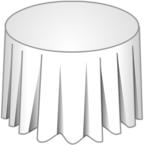}%
}

\vspace{0.08cm}

\begin{tcolorbox}[
    width=\linewidth,
    colback=blue!5,
    colframe=blue!50,
    boxrule=1pt,
    arc=2mm,
    left=3pt,
    right=3pt,
    top=2pt,
    bottom=2pt
]
{\small
\textbf{\textcolor{blue!70!black}{Question:}} Which material is this tablecloth made of? \\  \\ A) ceramic \\ B) linen

\vspace{2pt}
\textbf{Ground Truth:} \colorbox{green!20}{\textbf{B) linen}}
}
\end{tcolorbox}

\vspace{0.08cm}

\begin{tcolorbox}[
    width=\linewidth,
    colback=yellow!8,
    colframe=orange!60,
    boxrule=1pt,
    arc=2mm,
    left=3pt,
    right=3pt,
    top=2pt,
    bottom=2pt,
    title={\centering\small\textbf{Expert Reference Steps (7 steps)}},
    fonttitle=\bfseries
]
{\small
\begin{enumerate}[leftmargin=10pt,itemsep=0pt,parsep=0pt,topsep=1pt]
  \item The question asks about the material of the tablecloth.
  \item Common materials mentioned include wood, glass, metal, and plastic.
  \item Ceramic is not listed as a common material in the context.
  \item Linen is a type of fabric typically used for items like tablecloths.
  \item Tablecloths are generally made from fabric materials like linen, not hard materials like ceramic.
  \item The context supports linen being a reasonable material for a tablecloth.
  \item Select option B
\end{enumerate}
}
\end{tcolorbox}

\caption{\textbf{Scienceqa training example (ID 29750).} Representative sample from GRPO training dataset showing input image, question with ground truth answer, and expert-generated reference reasoning steps. These step-level annotations enable reinforcement learning with Match F1 rewards without requiring manual step annotation during deployment.}
\label{fig:grpo_train_ex1}
\end{figure}

\begin{figure}[t]
\centering

{{\small\bfseries Scienceqa Training Sample (ID 25092)}}\par
\vspace{1pt}
\fcolorbox{black!40}{white}{%
  \includegraphics[width=0.6\linewidth,height=0.35\textheight,keepaspectratio]{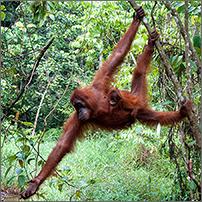}%
}

\vspace{0.08cm}

\begin{tcolorbox}[
    width=\linewidth,
    colback=blue!5,
    colframe=blue!50,
    boxrule=1pt,
    arc=2mm,
    left=3pt,
    right=3pt,
    top=2pt,
    bottom=2pt
]
{\small
\textbf{\textcolor{blue!70!black}{Question:}} Which animal's limbs are also adapted for climbing trees? \\  \\ A) okapi \\ B) three-toed sloth

\vspace{2pt}
\textbf{Ground Truth:} \colorbox{green!20}{\textbf{B) three-toed sloth}}
}
\end{tcolorbox}

\vspace{0.08cm}

\begin{tcolorbox}[
    width=\linewidth,
    colback=yellow!8,
    colframe=orange!60,
    boxrule=1pt,
    arc=2mm,
    left=3pt,
    right=3pt,
    top=2pt,
    bottom=2pt,
    title={\centering\small\textbf{Expert Reference Steps (12 steps)}},
    fonttitle=\bfseries
]
{\small
\begin{enumerate}[leftmargin=10pt,itemsep=0pt,parsep=0pt,topsep=1pt]
  \item The question asks which animal's limbs are adapted for climbing trees.
  \item The supporting context mentions adaptations of limbs for different functions like running, swimming, and flying.
  \item The provided context specifically states that some animals' limbs are adapted for climbing trees.
  \item A Sumatran orangutan is mentioned, but not as an option in the question.
  \item The supporting context explains that adaptations help organisms survive or reproduce, and includes limb adaptations.
  \item The question lists two options: okapi and three-toed sloth.
  \item The context from the Sumatran orangutan implies the importance of limbs adapted for climbing in a rainforest environment.
  \item Three-toed sloths are known to live in rainforest environments similar to the Sumatran orangutan.
  \item Three-toed sloths have limbs adapted for climbing trees, as implied by the supporting text about rainforest adaptations.
  \item Three-toed sloths use their limbs for climbing, fitting the description in the context about limb adaptations.
\end{enumerate}
\textit{[2 more steps omitted for brevity]}
}
\end{tcolorbox}

\caption{\textbf{Scienceqa training example (ID 25092).} Representative sample from GRPO training dataset showing input image, question with ground truth answer, and expert-generated reference reasoning steps. These step-level annotations enable reinforcement learning with Match F1 rewards without requiring manual step annotation during deployment.}
\label{fig:grpo_train_ex2}
\end{figure}

\begin{figure}[t]
\centering

{{\small\bfseries Scienceqa Training Sample (ID 24320)}}\par
\vspace{1pt}
\fcolorbox{black!40}{white}{%
  \includegraphics[width=0.6\linewidth,height=0.35\textheight,keepaspectratio]{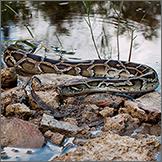}%
}

\vspace{0.08cm}

\begin{tcolorbox}[
    width=\linewidth,
    colback=blue!5,
    colframe=blue!50,
    boxrule=1pt,
    arc=2mm,
    left=3pt,
    right=3pt,
    top=2pt,
    bottom=2pt
]
{\small
\textbf{\textcolor{blue!70!black}{Question:}} Select the organism in the same species as the Burmese python. \\  \\ A) Python reticulatus \\ B) Cervus canadensis \\ C) Python bivittatus

\vspace{2pt}
\textbf{Ground Truth:} \colorbox{green!20}{\textbf{ C) Python bivittatus}}
}
\end{tcolorbox}

\vspace{0.08cm}

\begin{tcolorbox}[
    width=\linewidth,
    colback=yellow!8,
    colframe=orange!60,
    boxrule=1pt,
    arc=2mm,
    left=3pt,
    right=3pt,
    top=2pt,
    bottom=2pt,
    title={\centering\small\textbf{Expert Reference Steps (9 steps)}},
    fonttitle=\bfseries
]
{\small
\begin{enumerate}[leftmargin=10pt,itemsep=0pt,parsep=0pt,topsep=1pt]
  \item The given organism is identified as a Burmese .
  \item The scientific name of the Burmese  is provided as Python bivittatus.
  \item The question asks for the organism in the same species as the Burmese .
  \item The lecture explains that the second word in the scientific name identifies the species within the genus.
  \item Among the options, Python bivittatus matches the species name of the Burmese .
  \item Python reticulatus has a different species name, reticulatus.
  \item Cervus canadensis is from a different genus and species, not related to s.
  \item The correct answer is the organism with the scientific name Python bivittatus.
  \item Select option C
\end{enumerate}
}
\end{tcolorbox}

\caption{\textbf{Scienceqa training example (ID 24320).} Representative sample from GRPO training dataset showing input image, question with ground truth answer, and expert-generated reference reasoning steps. These step-level annotations enable reinforcement learning with Match F1 rewards without requiring manual step annotation during deployment.}
\label{fig:grpo_train_ex3}
\end{figure}

\begin{figure}[t]
\centering

{{\small\bfseries Scienceqa Training Sample (ID 26538)}}\par
\vspace{1pt}
\fcolorbox{black!40}{white}{%
  \includegraphics[width=0.6\linewidth,height=0.35\textheight,keepaspectratio]{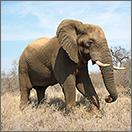}%
}

\vspace{0.08cm}

\begin{tcolorbox}[
    width=\linewidth,
    colback=blue!5,
    colframe=blue!50,
    boxrule=1pt,
    arc=2mm,
    left=3pt,
    right=3pt,
    top=2pt,
    bottom=2pt
]
{\small
\textbf{\textcolor{blue!70!black}{Question:}} What is the African bush elephant's scientific name? \\  \\ A) Loxodonta cyclotis \\ B) Loxodonta africana

\vspace{2pt}
\textbf{Ground Truth:} \colorbox{green!20}{\textbf{B) Loxodonta africana}}
}
\end{tcolorbox}

\vspace{0.08cm}

\begin{tcolorbox}[
    width=\linewidth,
    colback=yellow!8,
    colframe=orange!60,
    boxrule=1pt,
    arc=2mm,
    left=3pt,
    right=3pt,
    top=2pt,
    bottom=2pt,
    title={\centering\small\textbf{Expert Reference Steps (14 steps)}},
    fonttitle=\bfseries
]
{\small
\begin{enumerate}[leftmargin=10pt,itemsep=0pt,parsep=0pt,topsep=1pt]
  \item The question asks for the scientific name of the African bush elephant.
  \item Option A is listed as ``Loxodonta cyclotis''.
  \item Option B is listed as ``Loxodonta africana''.
  \item Context notes that the African bush elephant's scientific name refers to Africa.
  \item Scientific names often include geographical information.
  \item The word ``africana'' in option B refers to ``Africa''.
  \item The word ``cyclotis'' does not refer to ``Africa''.
  \item Both elephants, bush and forest, live in Africa.
  \item The scientific name ``Loxodonta africana'' should logically belong to the African bush elephant.
  \item Supporting context specifies correct naming conventions for organisms.
\end{enumerate}
\textit{[4 more steps omitted for brevity]}
}
\end{tcolorbox}

\caption{\textbf{Scienceqa training example (ID 26538).} Representative sample from GRPO training dataset showing input image, question with ground truth answer, and expert-generated reference reasoning steps. These step-level annotations enable reinforcement learning with Match F1 rewards without requiring manual step annotation during deployment.}
\label{fig:grpo_train_ex4}
\end{figure}

\begin{figure}[t]
\centering

{{\small\bfseries Scienceqa Training Sample (ID 26273)}}\par
\vspace{1pt}
\fcolorbox{black!40}{white}{%
  \includegraphics[width=0.6\linewidth,height=0.35\textheight,keepaspectratio]{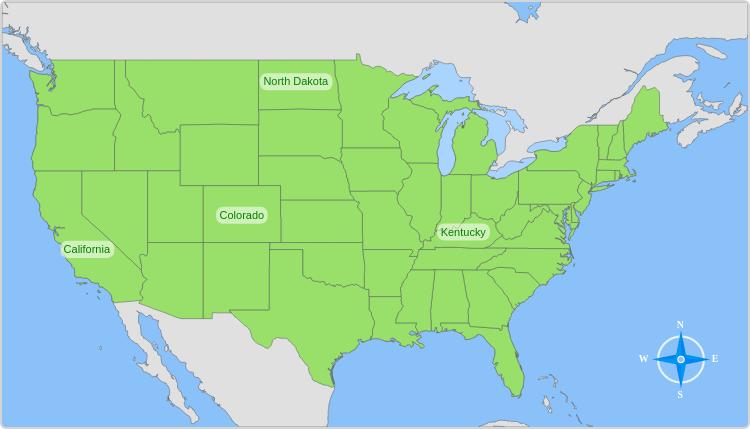}%
}

\vspace{0.08cm}

\begin{tcolorbox}[
    width=\linewidth,
    colback=blue!5,
    colframe=blue!50,
    boxrule=1pt,
    arc=2mm,
    left=3pt,
    right=3pt,
    top=2pt,
    bottom=2pt
]
{\small
\textbf{\textcolor{blue!70!black}{Question:}} Which of these states is farthest east? \\  \\ A) North Dakota \\ B) Colorado \\ C) California \\ D) Kentucky

\vspace{2pt}
\textbf{Ground Truth:} \colorbox{green!20}{\textbf{D) Kentucky}}
}
\end{tcolorbox}

\vspace{0.08cm}

\begin{tcolorbox}[
    width=\linewidth,
    colback=yellow!8,
    colframe=orange!60,
    boxrule=1pt,
    arc=2mm,
    left=3pt,
    right=3pt,
    top=2pt,
    bottom=2pt,
    title={\centering\small\textbf{Expert Reference Steps (13 steps)}},
    fonttitle=\bfseries
]
{\small
\begin{enumerate}[leftmargin=10pt,itemsep=0pt,parsep=0pt,topsep=1pt]
  \item Assess the geographical position of North Dakota, Colorado, and California.
  \item Consider the common knowledge that North Dakota, Colorado, and California are located further west relative to Kentucky.
  \item Recall that in maps, the cardinal directions are used to determine relative positions, with east being to the right when looking at most maps.
  \item Refer to the fact that Kentucky is known to be located in the eastern United States.
  \item Compare the typical eastward location of Kentucky relative to other states mentioned.
  \item Confirm that Kentucky is east of both the Mississippi River and the Rocky Mountains, unlike the other states listed.
  \item Incorporate the general scientific understanding that cardinal directions on a map help determine relative location.
  \item Utilize the information that maps usually have north at the top, implying east would be towards the right side of typical maps.
  \item Eliminate North Dakota as it is in the northern part of central USA.
  \item Eliminate Colorado as it is in the central western region of USA.
\end{enumerate}
\textit{[3 more steps omitted for brevity]}
}
\end{tcolorbox}

\caption{\textbf{Scienceqa training example (ID 26273).} Representative sample from GRPO training dataset showing input image, question with ground truth answer, and expert-generated reference reasoning steps. These step-level annotations enable reinforcement learning with Match F1 rewards without requiring manual step annotation during deployment.}
\label{fig:grpo_train_ex5}
\end{figure}

\begin{figure}[t]
\centering

{{\small\bfseries Textvqa Training Sample (ID 9690)}}\par
\vspace{1pt}
\fcolorbox{black!40}{white}{%
  \includegraphics[width=0.6\linewidth,height=0.35\textheight,keepaspectratio]{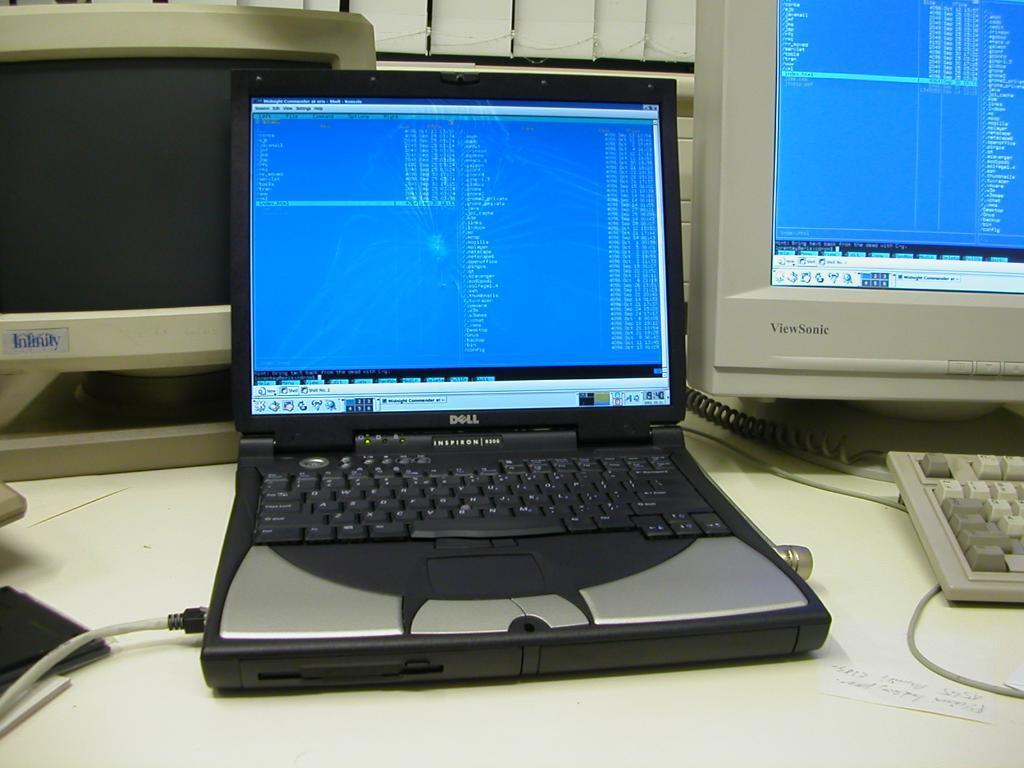}%
}

\vspace{0.08cm}

\begin{tcolorbox}[
    width=\linewidth,
    colback=blue!5,
    colframe=blue!50,
    boxrule=1pt,
    arc=2mm,
    left=3pt,
    right=3pt,
    top=2pt,
    bottom=2pt
]
{\small
\textbf{\textcolor{blue!70!black}{Question:}} what brand is this laptop?

\vspace{2pt}
\textbf{Ground Truth:} \colorbox{green!20}{\textbf{Dell}}
}
\end{tcolorbox}

\vspace{0.08cm}

\begin{tcolorbox}[
    width=\linewidth,
    colback=yellow!8,
    colframe=orange!60,
    boxrule=1pt,
    arc=2mm,
    left=3pt,
    right=3pt,
    top=2pt,
    bottom=2pt,
    title={\centering\small\textbf{Expert Reference Steps (7 steps)}},
    fonttitle=\bfseries
]
{\small
\begin{enumerate}[leftmargin=10pt,itemsep=0pt,parsep=0pt,topsep=1pt]
  \item The image contains two primary objects labeled: ``Laptop'' and ``Computer monitor''.
  \item The visible text on the laptop includes partial strings: ``INSPIRON''.
  \item The text ``ViewSonic'' is seen near the computer monitor.
  \item The text ``INSPIRON'' suggests a partial view of the brand name on the laptop.
  \item Considering the context, the brand name on the laptop appears to be related to the text ``DELL''.
  \item The brand on the laptop is correctly identified based on the visible scene text and object association.
  \item Return the exact answer string
\end{enumerate}
}
\end{tcolorbox}

\caption{\textbf{Textvqa training example (ID 9690).} Representative sample from GRPO training dataset showing input image, question with ground truth answer, and expert-generated reference reasoning steps. These step-level annotations enable reinforcement learning with Match F1 rewards without requiring manual step annotation during deployment.}
\label{fig:grpo_train_ex6}
\end{figure}

\begin{figure}[t]
\centering

{{\small\bfseries Textvqa Training Sample (ID 6036)}}\par
\vspace{1pt}
\fcolorbox{black!40}{white}{%
  \includegraphics[width=0.6\linewidth,height=0.35\textheight,keepaspectratio]{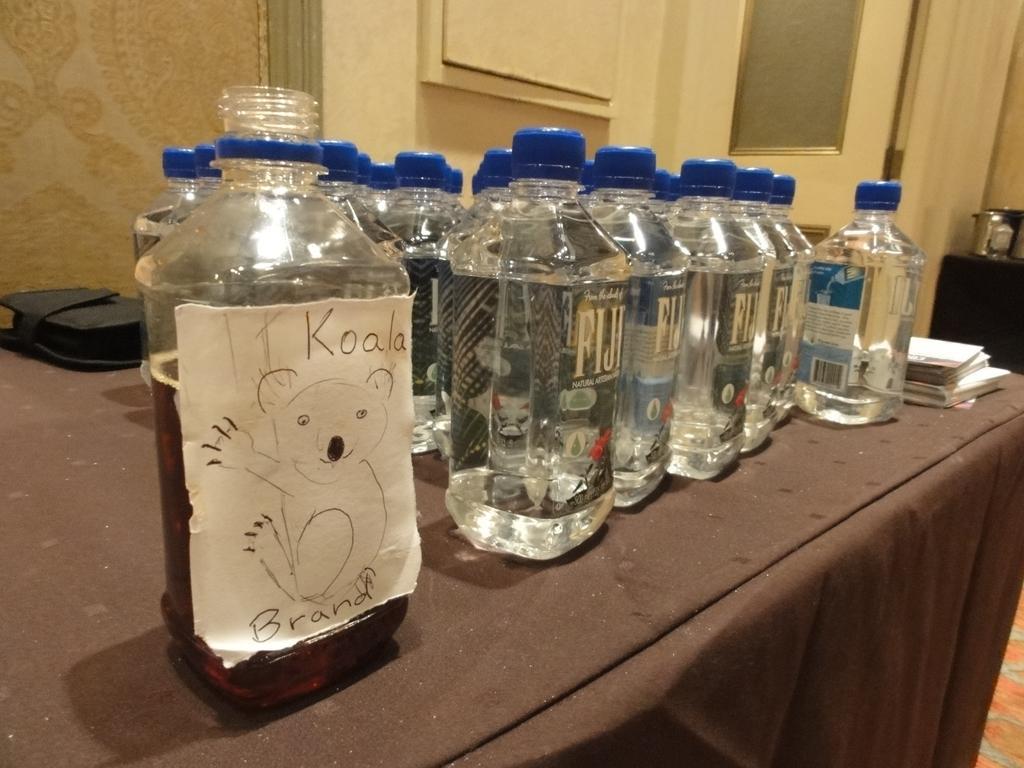}%
}

\vspace{0.08cm}

\begin{tcolorbox}[
    width=\linewidth,
    colback=blue!5,
    colframe=blue!50,
    boxrule=1pt,
    arc=2mm,
    left=3pt,
    right=3pt,
    top=2pt,
    bottom=2pt
]
{\small
\textbf{\textcolor{blue!70!black}{Question:}} what brand of water is shown?

\vspace{2pt}
\textbf{Ground Truth:} \colorbox{green!20}{\textbf{Fiji}}
}
\end{tcolorbox}

\vspace{0.08cm}

\begin{tcolorbox}[
    width=\linewidth,
    colback=yellow!8,
    colframe=orange!60,
    boxrule=1pt,
    arc=2mm,
    left=3pt,
    right=3pt,
    top=2pt,
    bottom=2pt,
    title={\centering\small\textbf{Expert Reference Steps (9 steps)}},
    fonttitle=\bfseries
]
{\small
\begin{enumerate}[leftmargin=10pt,itemsep=0pt,parsep=0pt,topsep=1pt]
  \item Identify visible text in the image that indicates a brand name.
  \item Observe a bottle in the image with text on it.
  \item Read the text on the label of the bottle which says ``Koala''.
  \item Note the presence of another brand name ``Fiji'' on a similar bottle or packaging.
  \item Confirm that both ``Koala'' and ``Fiji'' are displayed prominently on their respective products.
  \item Verify that the text is clearly readable and associated with the respective items.
  \item Cross-check that there are no other brand names visible that could be the answer.
  \item Conclusion: The image shows bottles labeled with the brand ``Fiji''.
  \item Return the exact answer string based on the observed text.
\end{enumerate}
}
\end{tcolorbox}

\caption{\textbf{Textvqa training example (ID 6036).} Representative sample from GRPO training dataset showing input image, question with ground truth answer, and expert-generated reference reasoning steps. These step-level annotations enable reinforcement learning with Match F1 rewards without requiring manual step annotation during deployment.}
\label{fig:grpo_train_ex7}
\end{figure}

\begin{figure}[t]
\centering

{{\small\bfseries Textvqa Training Sample (ID 4450)}}\par
\vspace{1pt}
\fcolorbox{black!40}{white}{%
  \includegraphics[width=0.6\linewidth,height=0.35\textheight,keepaspectratio]{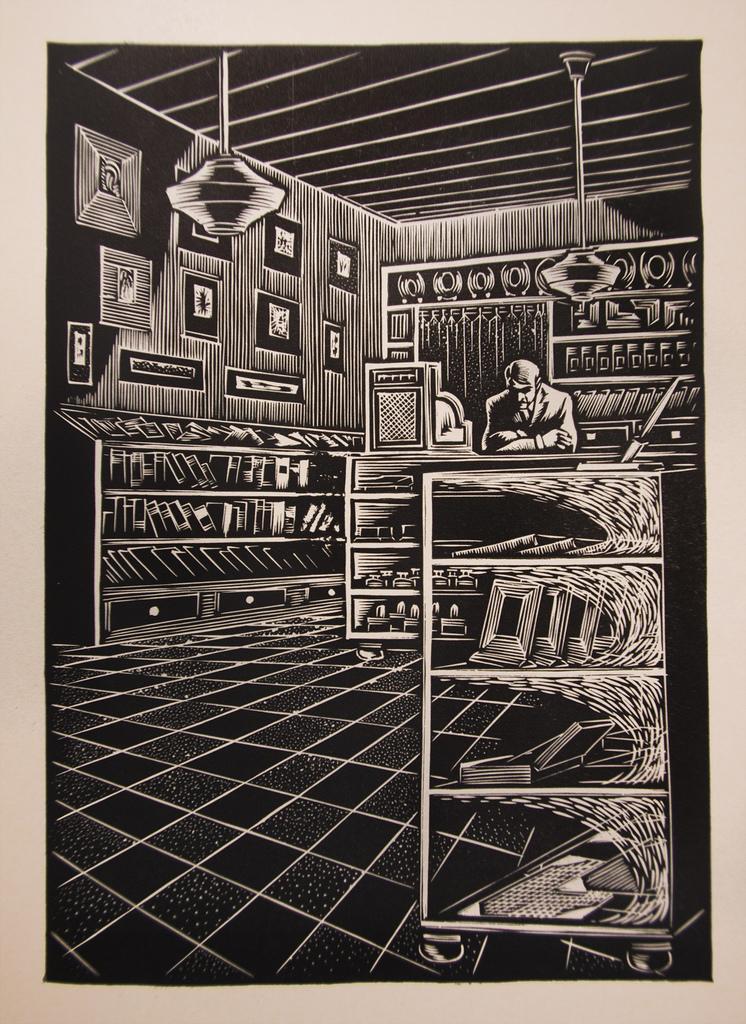}%
}

\vspace{0.08cm}

\begin{tcolorbox}[
    width=\linewidth,
    colback=blue!5,
    colframe=blue!50,
    boxrule=1pt,
    arc=2mm,
    left=3pt,
    right=3pt,
    top=2pt,
    bottom=2pt
]
{\small
\textbf{\textcolor{blue!70!black}{Question:}} how many people are in this photo?

\vspace{2pt}
\textbf{Ground Truth:} \colorbox{green!20}{\textbf{one}}
}
\end{tcolorbox}

\vspace{0.08cm}

\begin{tcolorbox}[
    width=\linewidth,
    colback=yellow!8,
    colframe=orange!60,
    boxrule=1pt,
    arc=2mm,
    left=3pt,
    right=3pt,
    top=2pt,
    bottom=2pt,
    title={\centering\small\textbf{Expert Reference Steps (10 steps)}},
    fonttitle=\bfseries
]
{\small
\begin{enumerate}[leftmargin=10pt,itemsep=0pt,parsep=0pt,topsep=1pt]
  \item Identify visible objects and entities in the image.
  \item Locate the position of a person using detected objects.
  \item Confirm that there is one person in the image.
  \item Note the presence of poster, clothing, and book.
  \item Ensure no additional people are observed near these objects.
  \item Verify that the scene text and surrounding elements do not indicate additional people.
  \item Check for any partial or occluded human forms.
  \item Determine that no text or icon suggests more than one person.
  \item Confirm the observation aligns with the given answer.
  \item Return the exact answer string.
\end{enumerate}
}
\end{tcolorbox}

\caption{\textbf{Textvqa training example (ID 4450).} Representative sample from GRPO training dataset showing input image, question with ground truth answer, and expert-generated reference reasoning steps. These step-level annotations enable reinforcement learning with Match F1 rewards without requiring manual step annotation during deployment.}
\label{fig:grpo_train_ex8}
\end{figure}

\begin{figure}[t]
\centering

{{\small\bfseries Textvqa Training Sample (ID 23353)}}\par
\vspace{1pt}
\fcolorbox{black!40}{white}{%
  \includegraphics[width=0.6\linewidth,height=0.35\textheight,keepaspectratio]{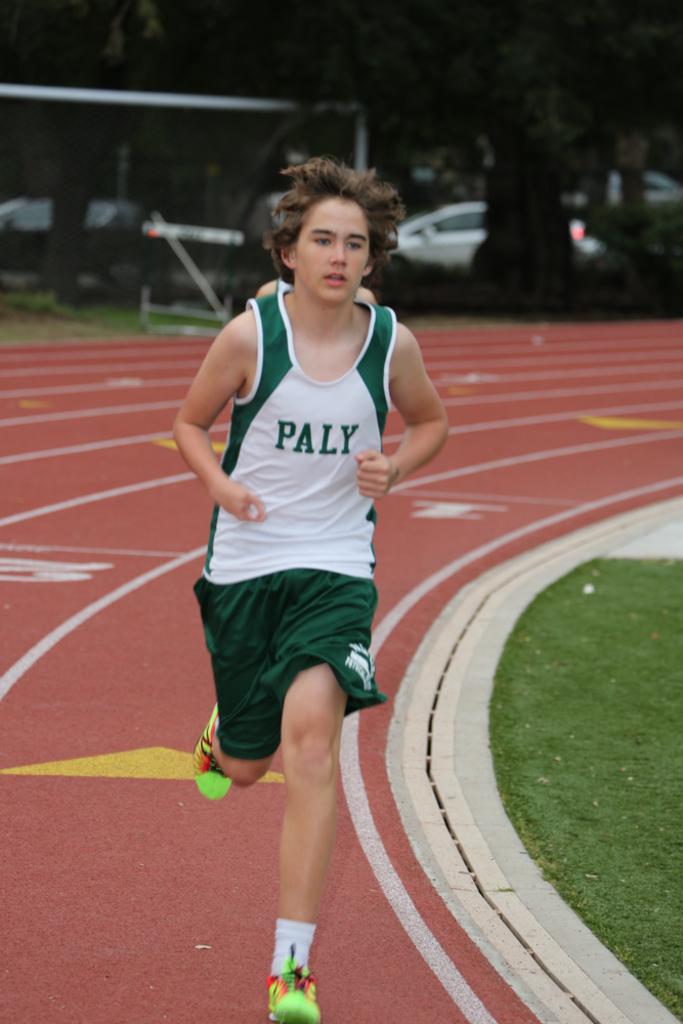}%
}

\vspace{0.08cm}

\begin{tcolorbox}[
    width=\linewidth,
    colback=blue!5,
    colframe=blue!50,
    boxrule=1pt,
    arc=2mm,
    left=3pt,
    right=3pt,
    top=2pt,
    bottom=2pt
]
{\small
\textbf{\textcolor{blue!70!black}{Question:}} what does the runner's jersey read?

\vspace{2pt}
\textbf{Ground Truth:} \colorbox{green!20}{\textbf{Paly}}
}
\end{tcolorbox}

\vspace{0.08cm}

\begin{tcolorbox}[
    width=\linewidth,
    colback=yellow!8,
    colframe=orange!60,
    boxrule=1pt,
    arc=2mm,
    left=3pt,
    right=3pt,
    top=2pt,
    bottom=2pt,
    title={\centering\small\textbf{Expert Reference Steps (15 steps)}},
    fonttitle=\bfseries
]
{\small
\begin{enumerate}[leftmargin=10pt,itemsep=0pt,parsep=0pt,topsep=1pt]
  \item The image shows a runner wearing a sports uniform.
  \item Focus on the runner's jersey in the image.
  \item Locate the text written on the runner's jersey.
  \item The jersey has the text ``PALY'' written on it.
  \item Verify that ``PALY'' is clearly visible on the jersey without ambiguity.
  \item Confirm that there are no other similar texts on the jersey or nearby objects.
  \item Ensure that ``PALY'' is the only readable text on the jersey.
  \item Check that ``PALY'' is written in uppercase letters.
  \item Note that the text is centered on the chest area of the jersey.
  \item Rule out any blur, occlusion, or cropping that might affect readability.
\end{enumerate}
\textit{[5 more steps omitted for brevity]}
}
\end{tcolorbox}

\caption{\textbf{Textvqa training example (ID 23353).} Representative sample from GRPO training dataset showing input image, question with ground truth answer, and expert-generated reference reasoning steps. These step-level annotations enable reinforcement learning with Match F1 rewards without requiring manual step annotation during deployment.}
\label{fig:grpo_train_ex9}
\end{figure}

\begin{figure}[t]
\centering

{{\small\bfseries Textvqa Training Sample (ID 3762)}}\par
\vspace{1pt}
\fcolorbox{black!40}{white}{%
  \includegraphics[width=0.6\linewidth,height=0.35\textheight,keepaspectratio]{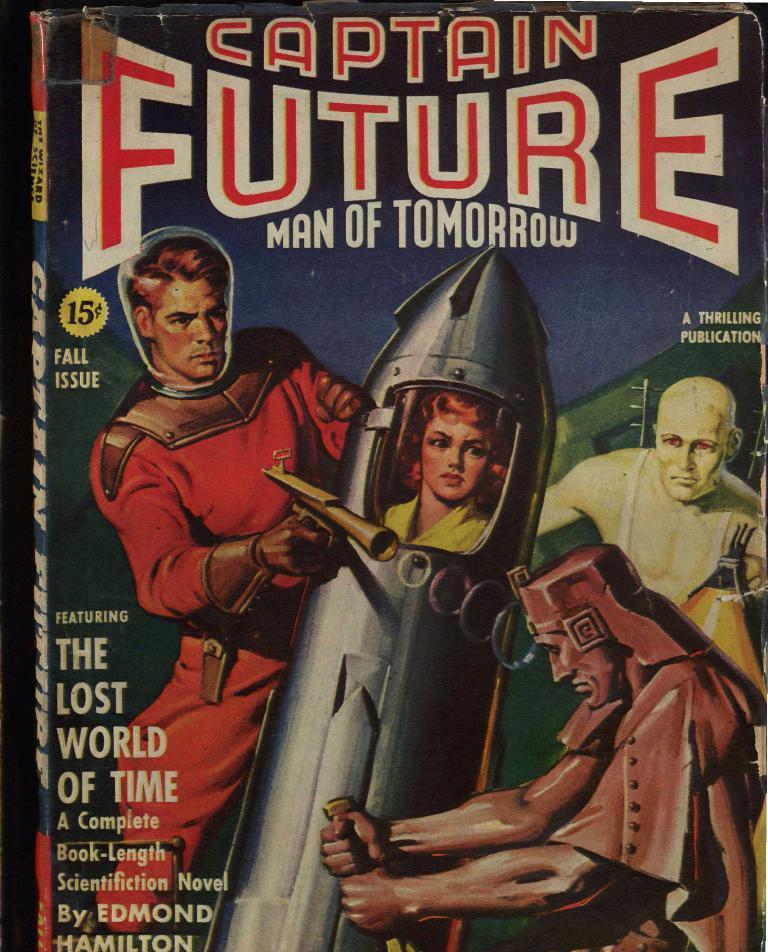}%
}

\vspace{0.08cm}

\begin{tcolorbox}[
    width=\linewidth,
    colback=blue!5,
    colframe=blue!50,
    boxrule=1pt,
    arc=2mm,
    left=3pt,
    right=3pt,
    top=2pt,
    bottom=2pt
]
{\small
\textbf{\textcolor{blue!70!black}{Question:}} what is the title here?

\vspace{2pt}
\textbf{Ground Truth:} \colorbox{green!20}{\textbf{captain future man of tomorrow}}
}
\end{tcolorbox}

\vspace{0.08cm}

\begin{tcolorbox}[
    width=\linewidth,
    colback=yellow!8,
    colframe=orange!60,
    boxrule=1pt,
    arc=2mm,
    left=3pt,
    right=3pt,
    top=2pt,
    bottom=2pt,
    title={\centering\small\textbf{Expert Reference Steps (15 steps)}},
    fonttitle=\bfseries
]
{\small
\begin{enumerate}[leftmargin=10pt,itemsep=0pt,parsep=0pt,topsep=1pt]
  \item The text ``CAPTAIN FUTURE'' is displayed prominently on the book cover.
  \item The word ``TOMORROW'' is positioned below ``CAPTAIN FUTURE'' in large letters.
  \item The text ``man of tomorrow'' is likely derived from the visible words and typical formatting practices.
  \item The book cover features a design that suggests a thematic connection to future-oriented content.
  \item The text ``Book-Length Novel'' is present, indicating the publication type.
  \item The author's name, ``EDMOND HAMILTON'', is visible below the main title.
  \item The publication label ``Scientifiction'' is located at the top of the book cover.
  \item The book shows signs of belonging to a vintage or classic series.
  \item The term ``COMPLETE Book'' suggests the book is part of a larger collection or series.
  \item The  ``15'' may refer to the price.
\end{enumerate}
\textit{[5 more steps omitted for brevity]}
}
\end{tcolorbox}

\caption{\textbf{Textvqa training example (ID 3762).} Representative sample from GRPO training dataset showing input image, question with ground truth answer, and expert-generated reference reasoning steps. These step-level annotations enable reinforcement learning with Match F1 rewards without requiring manual step annotation during deployment.}
\label{fig:grpo_train_ex10}
\end{figure}

\clearpage

\end{document}